\documentclass[11pt, letterpaper, copyright, logo]{googledeepmind}

\usepackage{multirow}
\usepackage[authoryear, sort&compress, round]{natbib}
\usepackage{lipsum}
\usepackage{amsmath}
\usepackage{pstricks, pst-node}
\usepackage{verbatim}
\usepackage{multirow}
\usepackage{multicol}
\usepackage{longtable}
\usepackage{array}
\usepackage{listings, listings-rust}
\usepackage{fontawesome5}
\usepackage{enumitem}
\usepackage{xspace}
\usepackage{bm}
\usepackage{bbm}
\usepackage{mathtools}
\usepackage{soul}
\usepackage{graphicx}
\usepackage[most, breakable, skins]{tcolorbox}
\tcbuselibrary{skins}
\usepackage{subcaption}
\usepackage{amssymb}
\usepackage{colortbl}
\usepackage{csquotes}
\usepackage{etex}
\usepackage{setspace}
\usepackage[inkscapeformat=png]{svg}
\usepackage{tabularx,ragged2e}
\usepackage{makecell}
\usepackage{placeins}
\usepackage[symbol]{footmisc}
\usepackage[dvipsnames]{xcolor}
\usepackage{hyperref}
\usepackage{pifont}
\usepackage{cleveref}

\newcolumntype{L}[1]{>{\raggedright\let\newline\\\arraybackslash\hspace{0pt}}m{#1}}
\newcolumntype{C}[1]{>{\centering}m{#1}}

\newcolumntype{R}[1]{>{\raggedleft\let\newline\\\arraybackslash\hspace{0pt}}m{#1}}

\definecolor{ao}{rgb}{0.0, 0.0, 1.0}

\newcommand{\todo}[1]{}
\newcommand{\todor}[1]{}

\newcommand{\optimam}[1]{}

\newcommand\vcent[1]{\vcenter{\hbox{#1}}}
\newcommand\loudspeaker[1][3]{\ensuremath{\vcent{\rule{.6ex}{.6ex}}\kern-.5ex%
  \vcent{\scalebox{.6}[1]{\rotatebox[origin=center]{90}{$\blacktriangle$}}}%
  \ifnum#1>0\relax\kern.05ex\vcent{\scalebox{.4}{\ttfamily)}}%
  \ifnum#1>1\relax\kern-.4ex\vcent{\scalebox{.56}{\ttfamily)}}%
  \ifnum#1>2\relax\kern-.55ex\vcent{\scalebox{.7}{\ttfamily)}}%
  \fi\fi\fi}%
}

\def\eg{{\em e.g.,}}
\def\ie{{\em i.e.,}}
\def\vs{{\em vs.~}}

\def\basemodel{{Gemini}\xspace}
\def\ourmodel{{Med-Gemini}\xspace}
\def\ourmodeltwod{{\ourmodel-2D}\xspace}
\def\ourmodelthreed{{\ourmodel-3D}\xspace}
\def\ourmodelpolygenic{{\ourmodel-Polygenic}\xspace}

\def\geminixl{{\basemodel Ultra}\xspace}

\def\numberofexample{{7}\xspace}
\def\numberofimage{{3.7}\xspace}

\newcommand{\tick}{\ding{51}}

\let\cite\citep

\title{Advancing Multimodal Medical Capabilities of Gemini}

\author[]{Google Research and Google DeepMind \footnote{\small{ See Contributions and Acknowledgments section for full author list. \\\hspace*{2.7em}\small{Corresponding authors: \{dangolden,\,shekazizi,\,kellych,\,roryp\}@google.com}.}}}

\begin{abstract}

Many clinical tasks require an understanding of specialized data, such as medical images and genomics, which is not typically found in general-purpose large multimodal models. Building upon \basemodel's multimodal models, we develop several models within the new \emph{\ourmodel} family that inherit core capabilities of \basemodel and are optimized for medical use via fine-tuning with 2D and 3D radiology, histopathology, ophthalmology, dermatology and genomic data.
\ourmodeltwod sets a new standard for AI-based chest X-ray (CXR) report generation based on expert evaluation, exceeding previous best results across two separate datasets by an absolute margin of 1\% and 12\%, where 57\% and 96\% of AI reports on normal cases, and 43\% and 65\% on abnormal cases, are evaluated as ``equivalent or better'' than the original radiologists' reports.
We demonstrate the first ever large multimodal model-based report generation for 3D computed tomography (CT) volumes using \ourmodelthreed, with 53\% of AI reports considered clinically acceptable, although additional research is needed to meet expert radiologist reporting quality. Beyond report generation, \ourmodeltwod surpasses the previous best performance in CXR visual question answering (VQA) and performs well in CXR classification and radiology VQA, exceeding SoTA or baselines on 17 of 20 tasks. In histopathology, ophthalmology, and dermatology image classification, \ourmodeltwod surpasses baselines across 18 out of 20 tasks and approaches task-specific model performance. Beyond imaging, \ourmodelpolygenic outperforms the standard linear polygenic risk score-based approach for disease risk prediction and generalizes to genetically correlated diseases for which it has never been trained. Although further development and evaluation are necessary in the safety-critical medical domain, our results highlight the potential of \ourmodel across a wide range of medical tasks.

\end{abstract}

\begin{document}

\maketitle

\clearpage
\section{Introduction}
\label{sec:intro}
\vspace{-6pt}
Medical data from diverse sources like biobanks, electronic health records, medical imaging, wearables, biosensors, and genomic sequencing are enabling the development of multimodal AI solutions that can better capture the complexity of human health and disease~\cite{acosta2022multimodal}. While AI in medicine has primarily focused on narrow tasks with single input and output types~\cite{rajpurkar2022ai}, recent advances in generative AI show promise in addressing multimodal, multi-task challenges in medical settings~\cite{moor2023foundation,moor2023med}. 

The emergence of large language models (LLMs) and large multimodal models (LMMs) such as Flamingo \cite{alayrac2022flamingo}, PaLI ~\cite{chen2022pali}, GPT-4~\cite{achiam2023gpt}, GPT-4v \cite{openai2023gpt4v}, PaLM~\cite{anil2023palm,chowdhery2023palm}, LLaMA~\cite{touvron2023llama}, LLaVa~\cite{liu2023improved,liu2024visual}, and Mistral 7B~\cite{jiang2023mistral} that promise significantly enhanced context length and improved multimodal capabilities suggests that the realization of highly complex multimodal reasoning across various medical data will soon be achievable. These advancements have catalyzed the expansion of LLMs specifically designed for medical domains, such as Med-PaLM and its successor Med-PaLM 2~\cite{singhal2023large,singhal2023towards}, Clinical Camel~\cite{toma2023clinical}, MedAlpaca~\cite{han2023medalpaca}, BioMistral~\cite{labrak2024biomistral}, sc-GPT~\cite{cui2024scgpt}, and others. Going beyond text alone, recent works have extended the capabilities of these base multimodal models by building models that cover various medical imaging modalities like Med-PaLM M~\cite{tu2024towards}, Med-Flamingo~\cite{moor2023med} as well as those that focus on a specific imaging domain, such as radiology~\cite{tanno2024consensus,thawkar2023xraygpt,hyland2023maira,xu2023elixr,hamamci2024ct2rep} and histopathology~\cite{sun2024pathasst,ikezogwo2024quilt,lu2024visual}.

\begin{figure*}[htp]
\small
    \centering
    \includegraphics[width=0.985\textwidth]{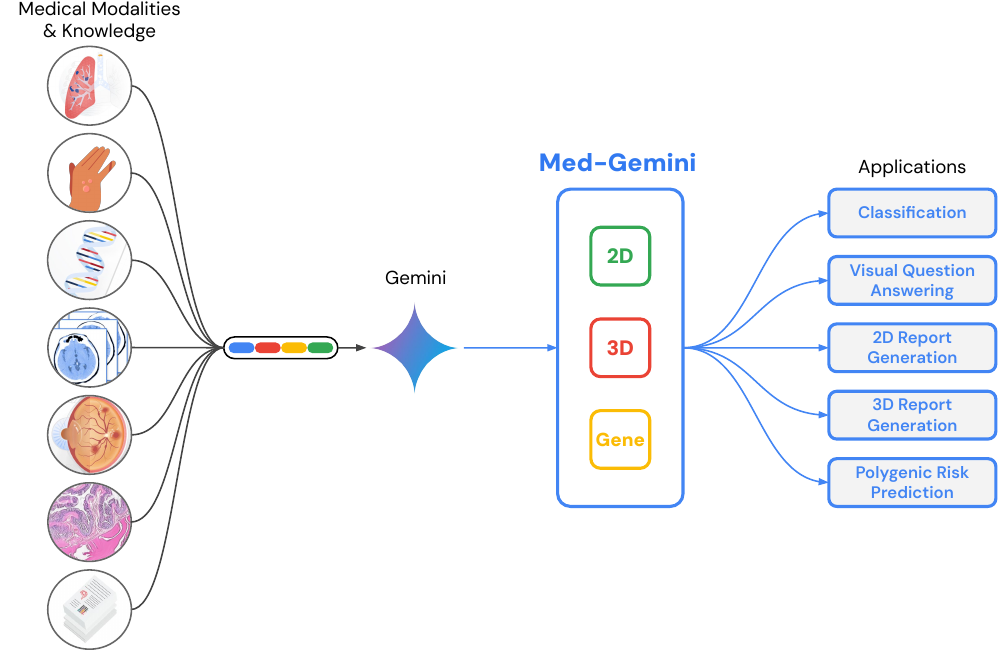} 
    
    \vspace{0.5cm}
    
    \includegraphics[width=0.95\textwidth]{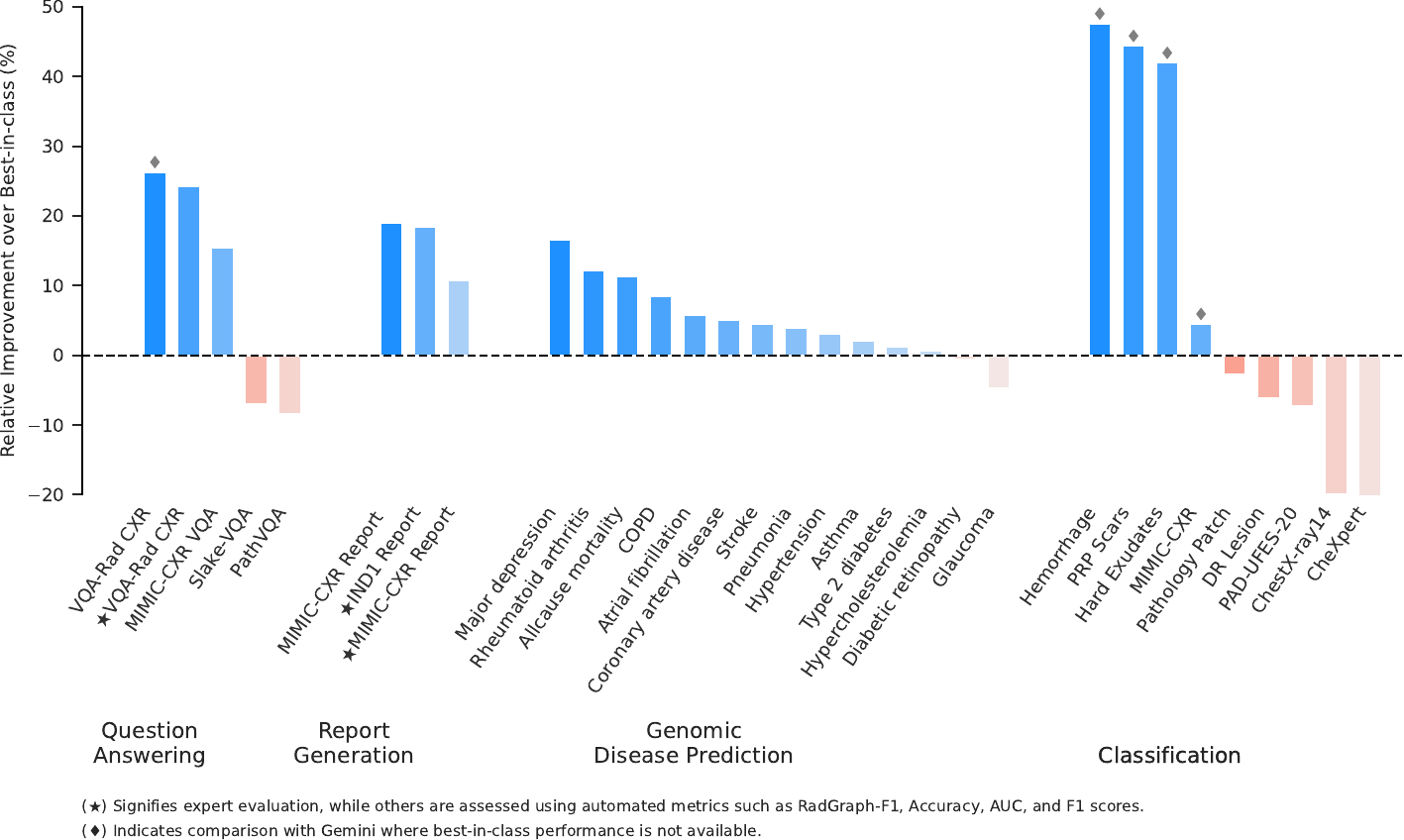}
    \vspace{3pt}
    \caption{\small{\textbf{Overview of our approach to curate and assess our family of medically tuned Gemini models, \ourmodel.} (top) These models build upon Gemini's powerful capabilities in advanced reasoning, multimodal understanding, and long-context processing enriched with patient representation and medical knowledge. (bottom) Relative performance of \ourmodel compared to SoTA or baselines across various tasks as detailed in Table~\ref{tab-appendix:performance-all}. Expert evaluation confirms that \ourmodeltwod sets a new standard for AI-powered chest X-ray report generation, with relative improvements of 10\% and 18\% over the previous leading model across two distinct datasets. In histopathology, ophthalmology, and dermatology image classification, it surpasses baseline on 18 out of 20 tasks and approaches task-specific model performance. \ourmodelpolygenic outperforms the standard approach for disease risk prediction and generalizes to diseases for which it has never been trained. }}
    \label{fig:teaser}
\end{figure*}

The release of the Gemini models~\cite{team2023gemini, team2024gemini}, with their advanced multimodal capabilities and breakthroughs in long-context understanding, marked a significant step forward in multimodal reasoning. Given its inherent human focus, medicine is a field in which advanced multimodal systems like Gemini are expected to be transformative ~\cite{acosta2022multimodal}. Evaluations have already started to evaluate the base performance of these newer multimodal models ~\cite{pal2024gemini}. However, the true potential of multimodal foundation models in the medical field remains largely underexplored due to the complexity of optimizing for problems in this field~\cite{moor2023foundation,rajpurkar2022ai} and a lack of diverse and meaningful evaluations that are grounded in clinical use cases~\cite{royer2024multimedeval,zhang2023biomedgpt,fleming2023medalign}. To better understand the nuances of model capabilities and limitations, it is necessary to optimize multimodal models for a diversity of relevant clinical applications and rigorously evaluate them on appropriate clinical datasets.

This report details our efforts in exploring Gemini's capabilities across a range of challenging multimodal medical tasks. Our evaluation benchmarks include 2D and 3D radiology images, histopathology patches, ophthalmology images, dermatology images, and genetic risk scoring. Our benchmark suite includes both open benchmark datasets and our own curated datasets. Open benchmark datasets have the advantage of being established and enabling direct comparison to others' work, but they are often limited or methodologically flawed, leading to results that can overstate performance. For the custom benchmarks that we introduce, we have prioritized high quality metrics that are closely correlated to clinical utility. In particular, we focused on expert human evaluations for quantifying performance on CXR and CT report generation and on open visual question answering (VQA) questions from VQA-Rad. Additionally, we compared \ourmodel to previous work or to the non-medically tuned version of \basemodel where possible.

Where we believed it was helpful, we have proactively improved the quality of certain open benchmarks. This includes updating and correcting erroneous labels (such as MIMIC-CXR-JPG classification labels), extending the task scope of datasets (such as introducing VQA question/answer pairs for MIMIC-CXR), and refining data splits to remove train-test contamination (such as PAD-UFES-20 and VQA-Rad). We hope to release these improvements publicly soon.

In this report, we expand the fine-tuned family of models, \ourmodel, specifically focusing on medical imaging and genomics. The models described here were tuned on a dataset of \numberofexample  million samples obtained from \numberofimage million medical images and cases, spanning medical image classification, VQA, report generation, and genomic risk prediction, detailed in Section~\ref{sec:data}. Importantly, this dataset includes mostly free text paired with medical data, which eliminates the need for expensive expert labeling of the training data. We intentionally explore both medical image-based tasks and also the non-image-based task of polygenic risk prediction in order to evaluate the potential of \ourmodel beyond imaging and in the crucial medical domain of long term risk prediction. Our findings demonstrate that LMMs have significantly advanced over the past year and are able to perform an increasing range of challenging tasks. Our key contributions are summarized as follows:

\begin{itemize}[leftmargin=1.5em,rightmargin=0em]
\setlength\itemsep{5pt}
\item \textbf{\ourmodel}: A family of generalist medical AI models fine-tuned from \basemodel, capable of performing a diverse set of medical tasks including medical image classification, VQA, report generation, and genomic risk prediction. \ourmodel extends \basemodel's capabilities to include interpretation of diverse medical data, including both genomics and 2D and 3D medical images. Additional capabilities of \ourmodel are described in ``Capabilities of Gemini Models in Medicine'' by \citet{saab2024capabilities}.
\item \textbf{Clinically-relevant benchmarking}: We evaluate \basemodel and \ourmodel on a comprehensive set of clinically relevant benchmarks including 22 datasets across five different tasks and six distinct medical image modalities. Our evaluation suite includes eight out-of-distribution datasets to assess generalization capabilities of this new family of models. Our assessments primarily consist of automated metrics but we rely on expert human evaluation for tasks where expert human judgment is critical, namely chest X-ray and computed tomography (CT) report generation and radiology VQA on open questions in the VQA-Rad dataset.
\item \textbf{Promising or best-in-class performance in several clinically relevant tasks}: \ourmodel demonstrates best in class performance on chest X-ray and CT report generation and chest X-ray classification. \ourmodel can also be used to predict disease and mortality risk more accurately than a standard linear polygenic risk score (PRS) based approach. \ourmodel approaches the performance of models trained using orders of magnitude more training examples on dermatology, histopathology, and ophthalmology image classification and demonstrates competitive performance across several VQA tasks across pathology and radiology.
\end{itemize}

\section{Datasets}
\label{sec:data}
Many different public and private datasets were used in the training and evaluation of \ourmodel. All datasets were de-identified. Open datasets were used in accordance with their existing licenses and private datasets were used with permission and appropriate licenses.

\vspace{-6pt}
\subsection{Datasets for fine-tuning and instruction-tuning}

Datasets were split into train, validation, and test sets by patient identifier when available. When patient identifiers were not available for a dataset, we ensured that there was no case or image overlap between splits.  

\subsubsection{Public datasets}

\paragraph{MIMIC-CXR:} MIMIC-CXR contains 377,110 images from 65,379 patients, with de-identified free-text reports describing the images~\cite{johnson2019mimicdatabase, johnson2019mimic, goldberger2000physiobank}. This dataset is the largest public chest X-ray dataset, acquired in the emergency department of Beth Israel Deaconess Medical Center in the US. For each patient, there are multiple views and a corresponding report labeled for 13 common radiological conditions using the CheXpert labeler~\cite{irvin2019chexpert} or with ``no finding'' if no condition is present.  Available labels include atelectasis, cardiomegaly, consolidation, edema, enlarged cardiomediastinum, fracture, lung lesion, lung opacity, pleural effusion, pleural other, pneumonia, pneumothorax, support devices, and no finding.  We used the MIMIC-CXR training set (237,912 images) to fine-tune Gemini as described in Section~\ref{sec:methods} and detailed in Table~\ref{tab:training-dataset-info}. We further employed the test cases of MIMIC-CXR as a benchmark for multiple evaluation tasks including classification, report generation and VQA. For the report generation task, we used the chest X-ray image corresponding to the frontal view (anterior-posterior or posterior-anterior) to generate the ~\emph{Findings} and ~\emph{Impression} sections, similar to prior works~\cite{tanno2024consensus}. For cases where no frontal view was available, we excluded them from our evaluation. For the VQA task, we utilized the condition-dependent VQA dataset (e.g. pleural effusion presence/location/severity) introduced in~\citet{xu2023elixr}, which we will make publicly available soon. In addition, we have used radiologist-adjudicated updated labels for findings that are also planned to be released soon~\cite{mimic-cxr-gt}.

\paragraph{Mendeley digital knee X-ray images:} This dataset consists of digital X-ray images of knee joints which were collected from hospitals and diagnostic centers. Original images are 8-bit grayscale. Each knee X-ray image was manually annotated by two medical experts following the Kellgren and Lawrence system for classification of osteoarthritis~\cite{gornale2020digital}. There are a total of 1,633 unique images in this dataset, and we utilized 1,469 images in our training.

\paragraph{PAD-UFES-20:} The dataset consists of 1,373 patients, 1,641 skin lesions, and 2,298 images. Skin lesion images were collected from various smartphones and exhibit variations in resolution, size, and lighting conditions. The dataset was acquired in collaboration with the Dermatological and Surgical Assistance Program (PAD) at the Federal University of Espírito Santo (UFES-Brazil)~\cite{pacheco2020pad}. PAD is a non-profit program offering free skin lesion treatment, particularly to those who cannot afford private care. The dataset includes images of six different skin lesion types and diagnostics: three skin diseases and three skin cancers. These include basal cell carcinoma (BCC), squamous cell carcinoma (SCC), actinic keratosis (ACK), seborrheic keratosis (SEK), Bowen’s disease (BOD), melanoma (MEL), and nevus (NEV). We randomly split the dataset into train (corresponding to 90\% of total samples) and test (10\% of total samples). We utilized 2,047 training samples from this dataset in our training corpus when fine-tuning Gemini as described in Section~\ref{sec:methods} and detailed in Table~\ref{tab:training-dataset-info}. We intend to publicly release our dataset split soon.

\paragraph{National Lung Screening Trial (NLST):} We utilized CT volumes from the validation subset of the NLST dataset \cite{nlst2014lung} and processed them through a lung cancer screening system~\cite{kiraly2024} to create a dataset of the most salient 2D slices from CT volumes.  Captions were assigned based on the \textit{scr\char`_group} label in the NLST participant dictionary: values of 1 and 2 were considered as having nodules whereas a value of 3 were considered as not having a nodule. We then selected a total of 2,199 slices consisting of 1,324 studies with nodules and 875 without nodules for further analysis from the previously defined validation set~\cite{ardila2019}. The lung cancer screening system generated captions for each slice based on the first detected region. Slices without nodules received simpler descriptions, for example: \textit{``An axial CT slice of the middle lungs with no nodules.''} For slices containing nodules, the captions included: location of the nodule (left or right lung), suspicion level for malignancy, and estimated size in millimeters based on the screening system's output. We then split the slices into training and validation evenly based on the presence of nodules, allocating 80\% for training, including 1,759 image and caption pairs, and 20\% for validation, including 440 image and caption pairs. All 2D slice images were set to a window of $[-1400 HU, 100 HU]$.

\paragraph{Slake-VQA:} This dataset is a large bilingual (English and Chinese) VQA dataset meticulously annotated by experienced physicians~\cite{liu2021slake}. It offers 642 images with 14,028 question-answer pairs in three imaging modalities (i.e. CXR, CT, MRI). Slake-VQA includes various areas of radiology, covering human body regions like the brain, neck, chest, abdomen, and pelvic cavity. The dataset comprises 9,849 VQA samples for training, 2,109 for validation, and 2,070 for testing. Questions are diverse, including both open-ended (free-form) and closed-ended (yes/no) formats. They probe various image aspects such as plane, quality, position, organ, abnormality, size, color, shape, and related medical knowledge. We used only English-language examples from the official splits, which included 4,919 training, 1,053 validation, and 1,061 test examples. 

\paragraph{PathVQA:} This is a dataset of question-answer pairs on pathology images~\cite{he2020pathvqa}. The dataset includes both open-ended questions and closed-ended (yes/no) questions and is built with automated methods using two publicly-available pathology textbooks and a publicly-available digital library. The dataset includes 32,632 question-answer pairs on 4,289 images. The official training, validation, and test splits contain 19,654, 6,259, and 6,719 QA pairs. We leveraged the official train and test sets for training our model and evaluating its performance, respectively.

\paragraph{VQA-Med:} The VQA-Med-2019 dataset offers a collection of medical images and associated question-answer (QA) pairs for model training and evaluation~\cite{asma2021vqamed}. It includes a training set with 3,200 medical images and 12,792 QA pairs, a validation set with 500 medical images and 2,000 QA pairs, and a test set containing 500 medical images and 500 questions. For the purpose of this report, we removed all images that overlapped with VQA-Rad~\cite{lau2018dataset} to avoid contamination. This resulted in 12,664 QA pairs used in training.

\paragraph{UK Biobank:} Genetic factors play a significant role in an individual's risk of developing various diseases. In this work, we used UK Biobank~\cite{bycroft2018ukbiobank}, a resource of nearly 500,000 de-identified individuals with genetic, lifestyle, and health information, to develop a task that takes as input an embedding of an individual's genomic data and uses it to predict an individual's status for various broad health outcomes. We extracted a set of 432,090 samples of European genetically inferred ancestry with genomic data passing quality control thresholds and split it randomly into train, validation, and test splits containing 60\%, 20\%, and 20\% of the samples, respectively. Following best practices for polygenic risk prediction, we avoided including individuals who were genetically similar in two different data splits~\cite{choi2020tutorial}.

\paragraph{PMC-OA:} PMC-OA is a medical dataset with image-caption pairs collected from PubMedCentral’s OpenAccess subset. Using the method described in~\citet{zhang2023pmc}, we retrieved 3,110,109 scientific papers containing 15,505,259 image-caption pairs. To ensure meaningful analysis, we filtered for image-caption pairs containing at least one photographic image (e.g., excluding images corresponding to data figures), resulting in a final dataset of 2,246,656 image-caption pairs.

\subsubsection{Private datasets}

\paragraph{Histopathology patches:} Pathological examination of tissue samples is crucial for effective diagnosis and treatment planning. Data from nine tasks across six tissue types from prior work~\citep{lai2023domain} were used in our training set (Table~\ref{tab:histopathology-tasks}). Multi-class annotation masks were used for both sampling image patches from whole-slide images as well as generating captions. Patches of size 256~$\times$~256 were sampled from whole slide images in a class-balanced manner. For each of the nine tasks, up to 10,000 image patches were sampled for three different magnification levels (2, 1, and 0.5 microns-per-pixel), resulting in 207,603 unique patches. Patch-level captions were created via prompting of a large language model (Gemini Pro) with inputs including structured slide-level metadata as well as  patch-level annotation labels. Multiple captions per class for each task were generated and then manually reviewed to ensure an appropriate level of detail and accuracy, resulting in 5--7 captions per class across tasks. Combining the sampled patches with our curated captions resulted in 1,550,976 image-text pairs for fine-tuning. For examples of some of our curated captions corresponding to the annotation labels, see Table \ref{tab:histopathology-generated-captions}.

\paragraph{Fundus images (EyePACS):} Diabetic retinopathy is the leading cause of blindness in the working-age population of the developed world. We used the de-identified dataset from EyePACS Inc.~\cite{cuadros2009eyepacs} and converted diabetic lesion-level presence labels to captions. The lesions considered were microaneurysms, hemorrhages, hard exudates, panretinal photocoagulation (PRP) scars, neovascularization of the disc and neovascularization elsewhere. For caption conversion, if a given image has lesion presence, for example, microaneurysm and hemorrhage, the associated generated caption was ``microaneurysm is present, hemorrhage is present.'' For healthy eyes, we use ``no diabetic retinopathy related lesion'' as the caption. 12,976 images with lesions and 3,000 healthy eye images were used to construct the dataset.

\paragraph{Computed tomography images (CT-US1):} A comprehensive dataset comprising 753,247 CT studies with associated radiology reports from 615,384 patients was obtained from three major hospital regions in the United States. These CT studies included  head/neck, chest, heart, abdominal, spine, and extremity regions imaged with and without contrast. To ensure robust evaluation, we employed a patient-level random split for training, validation, and testing.  The data was divided into 70\% for training, 15\% for validation, and 15\% for testing on the patient level. After an ingestion process this resulted in 657,719 training volumes and a total of 23,649 validation volumes.  Due to the reliance on expert evaluation for report generation, a subset of 92 non-contrast head/neck CT volumes from unique patients in the test set was used for model assessment. Volumes were prepared as described in Section~\ref{data_preprocessing}.  Only axial image volumes containing more than 10 slices were included in the prepared data and the volume within the study with the most axial slices was selected for inference.

We also carefully processed the existing dataset to create an extra 2D CT slice dataset specifically tailored for training our 2D model.  This involved filtering radiology reports for specific series and image numbers, selecting the correct images, and windowing them to a window of $[-1000 HU, 100 HU]$. To ensure that the text pertained directly to the CT slice in question and was a comprehensive description of it, captions were generated by combining the sentence of the report referencing the image along with the following sentence. This process resulted in a dataset of 4,009 images consisting of 3,207 training and 802 validation examples, primarily focused on CT studies of the abdomen and pelvis.

\paragraph{Chest X-ray images (CXR-US2):} The CXR-US2 dataset corresponds to the training set of US1 in~\citet{xu2023elixr}. This dataset consists of 132,680 frontal chest X-ray images from 12,988 patients taken at an academic medical center in Illinois, USA. Further descriptive statistics can be found in~\citet{xu2023elixr}. 

\begin{table}[t]
\centering
\caption{\small{\textbf{Overview of the training datasets.} More than \numberofexample  million data samples from \numberofimage million medical images and cases is used for fine-tuning and further instruction-tuning of \basemodel for medical applications in \ourmodel. This includes diverse set of modalities including 2D and 3D radiology images, pathology, ophthalmology, and genomic data.  These datasets includes mostly free text paired with medical data, which eliminates the need for expensive expert labeling of the training data.}}
\footnotesize
\label{tab:training-dataset-info}
\renewcommand{\arraystretch}{1.25}
\begin{tabular}{l|c|c|c|p{2.4in}}
\Xhline{2.5\arrayrulewidth}
\textbf{Modality} & \textbf{Dataset}   & \textbf{No. examples} & \textbf{No. Images} & \textbf{Description} \\
\Xhline{2\arrayrulewidth}
\multirow{6}{*}{Radiology (2D)}  & Slake-VQA          & 4,919      & 450       & Radiology images \& QA pairs \\
                                 & MIMIC-CXR          & 2,142,892  & 231,483   & Radiology images \& free-form reports \\
                                 & Digital Knee X-ray & 1,469      & 1,469     & Knee X-ray images \& labels \\
                                 & CXR-US2            & 132,680    & 132,680   & Radiology images \& free-form reports\\
                                 & NLST               & 2,199      & 2,199     & 2D CT slices \& free-form reports \\
                                 & CT-US1             & 3,207      & 3,207     & 2D CT slices \& free-form reports \\ 
\hline                               
\multirow{1}{*}{Radiology (3D)}  & CT-US1             & 657,719    & 657,719   & 3D CT images \& free-form reports \\  
                                    
\hline 
\multirow{2}{*}{Pathology}   & PathVQA                & 19,654     & 2,599     & Pathology images \& QA pairs \\
                             & Histopathology         & 1,550,976  & 207,603   & Histopathology images, captions, \& QA pairs\\
\hline                              
\multirow{1}{*}{Dermatology} & PAD-UFES-20            & 2,047      & 2,047     & Skin lesion images \& labels \\
\hline 
\multirow{1}{*}{Ophthalmology} & EyePACS                & 14,406    & 14,406    & Fundus images \& labels \\
\hline 
\multirow{2}{*}{Medical VQA} & PMC                    & 2,246,656  & 2,246,656 & PubMed Central images \& caption pairs\\
                             & MedVQA                 & 12,664     & 3,168     & Medical images \& QA pairs \\
\hline 
\multirow{1}{*}{Genomics}     & UK Biobank            & 259,225  & 259,225   & Genomic data \& disease outcomes \\

\Xhline{2.5\arrayrulewidth}
\end{tabular}
\end{table}

\subsection{Held-out datasets for evaluation and benchmarking}
Beyond the test sets associated with our training datasets (described above), we also utilized multiple held-out and out-of-distribution (OOD) datasets.

\vspace{-6pt}
\subsubsection{Public datasets}

\paragraph{CheXpert} The CheXpert dataset is similar to the MIMIC-CXR dataset and consists of 224,316 chest X-ray images (both frontal and lateral views) from 65,240 patients~\cite{irvin2019chexpert}. It labels 14 distinct thoracic conditions, including ``No Finding''. The original CheXpert dataset contains positive, negative, uncertain and unmentioned labels. During evaluation we considered the ``unmentioned'' label as negative and included only chest X-rays depicting frontal views.

\paragraph{VQA-Rad} The VQA-Rad dataset~\cite{lau2018dataset} comprises 315 radiology images sourced from CT, MRI, and X-ray scans, and it encompasses three anatomical regions including the head, abdomen, and chest. This dataset includes a wide array of question types, spanning 11 distinct categories, such as modality, plane, organ system, abnormality, and more, where 58\% of the question-answer pairs are designed to be closed-ended (yes/no or limited choices), while the remaining 42\% are open-ended.

The standard and official splits of the dataset feature 1,797 QA pairs for training and 451 for testing purposes. However, due to contamination of images included in both training/test in the original dataset release, we constructed a new, non-overlapping test and tuning split for the subset of chest X-ray images and associated question-answer pairs, first described in~\citet{xu2023elixr}. In this study, we went one step further and created new image-disjoint splits of train, validation and test sets for all three image types. We ensured that the previous X-ray-only validation and test sets were subsets of the new validation and test sets, respectively, thus enabling comparisons with the ELIXR model~\cite{xu2023elixr} on the new test set, as no former validation examples were included in the new test set. Aside from these constraints, we sampled in a manner that roughly equalizes both the ratio of closed to open question-answer pairs for each anatomical region, see Table~\ref{tab:vqarad-open-closed-ratio}, as well as the distribution across the 11 different question types across the three splits, see Table~\ref{tab:vqarad-question-type-distrib} in Section~\ref{sub:new_rad_vqa_splits}. Henceforth, we refer to this new three-way split as the ``balanced split''. In total, the balanced VQA-RAD dataset split, which we will make publicly available soon, comprises 2,248 pairs of questions and answers, encompassing 1,299 closed-ended questions and 949 open-ended questions.

\paragraph{ChestX-ray14}  
The ChestX-ray14 dataset~\cite{summers2019nih} is a comprehensive medical imaging dataset containing 112,120 frontal-view chest X-ray images from 30,805 unique patients.  ChestX-ray14 builds upon the ChestX-ray8 dataset~\cite{wang2017chestx}, expanding the number of labeled diseases to fourteen common thoracic pathologies, including Atelectasis, Consolidation, Infiltration, Pneumothorax, Edema, Emphysema, Fibrosis, Effusion, Pneumonia, Pleural thickening, Cardiomegaly, Nodule, Mass, and Hernia. Because these labels are automatically derived using NLP techniques and therefore contain inherent uncertainty, we restricted our evaluation to a subset of 1,962 cases focusing on three radiologist-adjudicated conditions~\cite{majkowska2020chest}, namely lung opacity, pneumothorax, and fracture.

\paragraph{TCGA study type} This dataset utilizes histopathology images from The Cancer Genome Atlas (TCGA), for which different study types correspond to different cancer types with additional information via portal.gdc.cancer.gov. The patches from this dataset are sampled from 2,952 training slides, 1,466 validation slides, and 1,489 test slides across ten (10) distinct TCGA study types: BLCA, BRCA, COAD, HNSC, KIRC, LIHC, LUAD, LUSC, OV, and STAD~\cite{lai2023domain}. We used the test set as an out-of-distribution dataset to evaluate our model's ability to generalize to different histopathology-related tasks.

\subsubsection{Private datasets}

\paragraph{IND1} This is a private research dataset of a similar scale as MIMIC-CXR, which we refer to as IND1~\cite{nabulsi2021deep}. This dataset comprises 263,021 de-identified frontal chest X-rays (digital and scanned) along with their corresponding reports. The X-rays were collected from five regional centers (Bangalore, Bhubaneswar, Chennai, Hyderabad, and New Delhi) across a large hospital group in India between November 2010 and January 2018~\cite{ahn2022association}. We used the same test set as~\cite{tanno2024consensus}, and following their framework, 300 of those cases are used for human evaluation.

\paragraph{TTH tissue type} The TTH tissue type dataset, introduced by~\citet{weng2019multimodal} and~\citet{lai2023domain}, represents a patch-level tissue type classification task. This internal dataset comprises 17,319 training slides, 6,488 validation slides, and 6,719 test slides, encompassing a total of 16 distinct tissue types. These tissue types include Appendix, Breast, Cervix, Colon and Rectum, Fallopian Tube, Gallbladder, Liver, Lymph Node, Ovary, Placenta, Prostate, Skin, Thyroid, Upper GI, Uterus, and Vas Deferens. We used the test set to evaluate generalization of our model to different histopathology tasks.   

\begin{table}[]
\centering
\footnotesize
\caption{\small{\textbf{Overview of the datasets used for evaluating our fine-tuned Gemini models, \ourmodel.} 
Our evaluation leveraged a robust dataset suite encompassing \textcolor{black}{22} datasets across \textcolor{black}{5} different types of clinically relevant tasks. This included \textcolor{black}{8} out-of-distribution datasets to assess generalization and spanned \textcolor{black}{7} distinct medical image modalities. We explicitly explored medical image classification, VQA, 2D report generation, 3D report generation, and disease prediction from genetic risk embeddings. The total number of evaluation samples across these datasets exceeds 40,000.
}}
\label{tab:validation-dataset-info}
\renewcommand{\arraystretch}{1.15}
\begin{tabular}{@{}l|l|c|c|c}
\Xhline{2.5\arrayrulewidth}
\textbf{Task}  & \textbf{Dataset/Setup}& \textbf{Modality}  & \textbf{No. Samples} & \textbf{OOD}                 \\
\Xhline{2\arrayrulewidth}
\multirow{4}{*}{\begin{tabular}[c]{@{}l@{}}Radiology report generation (2D)\end{tabular}}    & MIMIC-CXR                     & Chest X-ray & 912          & -                   \\
                                                                                             & MIMIC-CXR (Expert Evaluation) & Chest X-ray & 206          & -                   \\ 
                                                                                             & IND1 (Expert Evaluation)      & Chest X-ray & 300          & \tick               \\
\hline
\multirow{1}{*}{\begin{tabular}[c]{@{}l@{}}Radiology report generation  (3D)\end{tabular}}  & CT-US1 (Expert Evaluation)                 & CT          & 92          & -                   \\ 
\hline
\multirow{3}{*}{VQA}                                                   & MIMIC-CXR VQA          & Radiology   & 226          & -                   \\
                                                                                             & Slake-VQA (English-only) & Radiology   & 1,061        & -                   \\
                                                                                             & VQA-Rad                & Radiology   & 2,248        & \tick               \\
                                                                                             & PathVQA                & Pathology   & 6,719        & -                   \\
\hline
\multirow{16}{*}{Classification}                                                             & PAD-UFES-20                     & Dermatology & 251          & -                   \\
                                                                                             & MIMIC-CXR (Abnormal/Normal)     & Chest X-ray & 2,242        & -                   \\
                                                                                             & MIMIC-CXR (5 classes)           & Chest X-ray & 2,242        & -                   \\
                                                                                             & CheXpert (Abnormal/Normal)      & Chest X-ray & 1,962        & \tick               \\
                                                                                             & CheXpert (5 classes)            & Chest X-ray & 518          & \tick               \\
                                                                                             & ChestX-ray14 (3 classes)        & Chest X-ray & 1,962        & \tick               \\

                                                                                             & CAMELYON16 (2 classes)          & Histopathology   & 258          & -              \\
                                                                                             & Gleason NCB (4 classes)         & Histopathology   & 88           & -              \\      
                                                                                             & Gleason RP (4 classes)          & Histopathology   & 202          & -              \\
                                                                                             & Lung AD (9 classes)             & Histopathology   & 202          & -              \\ 
                                                                                             & Breast IC (3 classes)           & Histopathology   & 669          & -              \\ 
                                                                                             & Breast NP (3 classes)           & Histopathology   & 945          & -              \\ 
                                                                                             & Breast TF (3 classes)           & Histopathology   & 945          & -              \\ 
                                                                                             & CIN (3 classes)                 & Histopathology   & 229          & -              \\       
                                                                                             & CRC (2 classes)                 & Histopathology   & 44           & -              \\

                                                                                             & TCGA Study Type (10 classes)  & Histopathology   & 1,489    & \tick               \\
                                                                                             & Tissue Type (16 classes)      & Histopathology   & 6,719    & \tick               \\    
                                                                                             & EyePACS Hard Exudates ( 2 classes) & Ophthalmology & 498      & -       \\
                                                                                             & EyePACS Hemorrhage (2 classes)     & Ophthalmology & 498      & -       \\
                                                                                             & EyePACS DR Lesion(Abnormal/Normal)            & Ophthalmology & 490      & -       \\
\hline                                                                                            
\multirow{14}{*}{\begin{tabular}[c]{@{}l@{}}Risk Prediction (UK Biobank) \end{tabular}}       & Coronary artery disease   & Genomic     & 400        & -   \\
                                                                                              & Stroke   & Genomic     & 400        & -  \\
                                                                                              & Type 2 diabetes   & Genomic     & 400        & -  \\
                                                                                              & Glaucoma   & Genomic     & 400        & -  \\
                                                                                              & Chronic obstructive pulmonary disease   & Genomic     & 400        & -  \\
                                                                                              & Rheumatoid arthritis   & Genomic     & 400        & -  \\
                                                                                              & Major depression   & Genomic     & 400        & -  \\
                                                                                              & All-cause mortality   & Genomic     & 400        & -  \\
                                                                                              & Hypertension   & Genomic     & 400        & \tick  \\ 
                                                                                              & Hypercholesterolemia    & Genomic     & 400        & \tick  \\ 
                                                                                              & Atrial fibrillation   & Genomic     & 400        & \tick  \\ 
                                                                                              & Diabetic retinopathy   & Genomic     & 400        & \tick  \\ 
                                                                                              & Pneumonia   & Genomic     & 400        & \tick  \\
                                                                                              & Asthma   & Genomic     & 400        & \tick  \\
\Xhline{2.5\arrayrulewidth}            
\end{tabular}
\end{table}

\subsection{Data preprocessing} 
\label{data_preprocessing}
\paragraph{Radiology 2D images} When available, images acquired in DICOM format were used to directly create examples for training and inference. In the case of X-rays, raw pixel data were extracted from the DICOM image pixel data, and the look up table (LUT, part of the DICOM metadata) was subsequently applied. If a DICOM file contained multiple LUTs, we used the first LUT entry. If the window width and window center were defined, these were also used for preprocessing. The final pixel data were re-scaled to the full range of $[0, 65535]$ for the 16-bit PNG format. X-ray images in a preprocessed format were taken as is.

\paragraph{CT volumes} All 3D CT volumes were derived from DICOM images. Only axial slices were used to establish a standardized anatomical perspective. Slices were sorted based on the Image Position (Patient) attribute and used to compute slice spacing and reconstruct volumes. Subsequently, images were clipped with a Hounsfield Unit (HU) range of $[-1024 HU, 1024 HU]$ to cover a full spectrum of densities (e.g. the typical window/level values of brain, soft tissues) and then scaled to $[0.0, 1.0]$. Finally, tricubic interpolation was used to resample all images to a voxel spacing of 0.7mm~$\times$~0.7mm~$\times$~1.4mm, ensuring consistent uniform resolution for accurate comparative analysis.

\paragraph{Genomics:} A genomic featurization for an individual consists of polygenic risk scores (PRSs) for 7,415 traits. Each PRS estimates the genetic risk of the individual for a particular disease or trait, calculated by aggregating the estimated effects of many common variants associated with the condition. Each PRS was computed using genome-wide association study summary statistics computed by the Pan-UKB Consortium \cite{panukb2020}. These genomic features were then converted to images by projecting the PRSs into patch-aligned squares of 8~$\times$~8 pixels with values between $[0, 255]$. The 3 RGB channels of the images were used to stack 3 different p-value thresholds of the projections. The PRS features were obtained from the genetic information of 314,540 individuals of European genetically inferred ancestry from the UK Biobank~\cite{sudlow2015mr,bycroft2018ukbiobank}.

To create training and evaluation labels, we selected eight in-distribution health outcomes which have strong heritability (i.e. genetic information plays an important role in influencing susceptibility~\cite{Visscher2008heritability}), span multiple organ systems, and are challenging to predict from polygenic risk scores alone: coronary artery disease, stroke, type 2 diabetes, glaucoma, chronic obstructive pulmonary disease (COPD), rheumatoid arthritis, major depression, and all-cause mortality. Additionally, to assess model generalization, we selected six out-of-distribution (OOD) health outcomes that share genetic correlation with one or more of the in-distribution health outcomes: hypertension, hypercholesterolemia, atrial fibrillation, diabetic retinopathy, asthma, and pneumonia (Table \ref{tab-app:genomics_data_overview}).

\paragraph{Pathology patches}
Patches with initial size of 256~$\times$~256 pixels were sampled from whole slide images using multi-class annotation masks in a class-balanced manner across three different magnification levels (2, 1, and 0.5 microns-per-pixel).

\paragraph{Input preprocessing and tokenization} Images from all 2D datasets were uniformly resized to 768~$\times$~768 pixels, preserving aspect ratio with padding, with pixel intensities scaled to $[0,1]$. This ensured image resolution would be high enough for the fine-grained detail of medical images. For text, we used the native Gemini SentencePiece tokenizer~\cite{kudo2018sentencepiece,team2023gemini} without modification.

\section{Modeling Methodology}
\label{sec:methods}
\subsection{Model architecture}

Gemini builds upon the robust foundation of Transformer decoders~\cite{vaswani2017attention, parmar2018image}, offering significant architectural and optimization enhancements for efficient, stable large-scale training~\cite{team2023gemini,barham2022pathways}. This equips Gemini with exceptional natural language understanding and text generation capabilities. Of particular interest for medical data processing, Gemini's multimodal design draws inspiration from foundational Google research on Flamingo ~\cite{alayrac2022flamingo}, CoCa~\cite{yu2022coca,yan2022videococa} and PaLI~\cite{chen2022pali}, enabling enhanced multimodal understanding and reasoning.

Gemini handles video understanding by encoding frames as a sequence within its large context window~\cite{team2023gemini,team2024gemini}. This allows seamless integration of video frames, multi-slice images, text, or audio inputs.  The model even supports variable input resolutions, enabling it to prioritize computational resources for tasks requiring high-resolution analysis. Gemini 1.5 specifically is a mid-size model with a context window of up to 1 million tokens and performance on par with the largest Gemini model, 1.0 Ultra. Given this exceptional efficiency, we chose to finetune \ourmodel from Gemini 1.5.

\subsection{Multimodal fine-tuning}

Three custom versions of the Gemini 1.5 Pro vision encoder were trained for 2D modalities, 3D modalities, and genomics. In our initial experiments, we found that custom vision encoders for each type of data format performed better than a single vision encoder for all data formats. Furthermore, fine-tuning the vision encoder as well as the language component in Gemini led to significantly better visual understanding in comparison to a model that used the native vision encoder of Gemini 1.5 Pro models. From these three custom vision encoders, we trained three specific variants of \ourmodel which we refer to as \ourmodeltwod, \ourmodelthreed, and \ourmodelpolygenic. Notably, \ourmodeltwod includes all conventional medical images that are encoded in 2D (e.g. chest X-ray, CT slices, pathology patches), \ourmodelthreed was built on top of \ourmodeltwod and handles 3D medical data (e.g. CT), and \ourmodelpolygenic was trained for a novel image encoding derived from non-image features (e.g. genomics).  For all three model variants, fine-tuning was framed as a captioning or VQA task.

\paragraph{Fine-tuning for 2D modalities - \ourmodeltwod}
All 2D modalities were fine-tuned together using the training mix described in Section~\ref{sec:data} and Table~\ref{tab:training-dataset-info} to create \ourmodeltwod.  The 2D modalities used for fine-tuning included the described radiology, pathology, dermatology, and ophthalmology images.

\paragraph{Fine-tuning for 3D modalities - \ourmodelthreed}

To interpret 3D medical data, we leveraged the video understanding capabilities of \basemodel~\cite{team2024gemini}. Use of the \basemodel video encoder allows \ourmodelthreed to process multiple 2D slices, replacing the time axis with the depth dimension, with computed tomography (CT) as our example modality. This 3D fine-tuned model can then synthesize information across a series of 2D slices to generate radiology reports. Use of this video encoding capability will permit analysis of other volumetric and time-series medical data (e.g. MRI, ultrasound) in the future.

\paragraph{Fine-tuning for genomics - \ourmodelpolygenic} Genomics ``images'' (polygenic risk scores (PRS) projected into 2D, see Section~\ref{data_preprocessing}) were included in the mixture of datasets used to fine-tune the \ourmodelpolygenic vision encoder, and were trained to predict eight broad health outcomes (coronary artery disease, stroke, type 2 diabetes, glaucoma, chronic obstructive pulmonary disease, rheumatoid arthritis, major depression, and all-cause mortality) in a captioning task.

\paragraph{Instruction fine-tuning}

To optimize the instruction-following capabilities of the fine-tuned \ourmodel even further, we subsequently employed an instruction-tuning phase. In this phase, we fine-tuned Gemini 1.5 Pro on a curated collection of multimodal data consisting of carefully crafted instruction and response pairs. By exposing the model to these examples, we refined its ability to not only understand the content of medical images and signals, but also to follow nuanced instructions and generate tailored outputs.

\subsection{Model training and inference infrastructure}
Like its predecessor Gemini 1.5 Pro and all other Gemini models, \ourmodel was trained on large-scale Google TPUv4 accelerator pods spread across multiple data-centers. This training setup significantly scales up from our previous flagship PaLM family~\cite{chowdhery2023palm}. The Gemini architecture ensures efficient serving on TPU accelerators at scale.  For detailed information on training and serving Gemini models, see~\cite{team2023gemini,team2024gemini}.

\section{Evaluation and Results}
\label{sec:eval}
The following sections explore in detail how \basemodel and \ourmodel perform across various modalities and tasks in the medical field. Due to restrictions in our data licenses, our evaluation was limited to internal models. Our evaluation leveraged a robust dataset suite encompassing 22 datasets across four different clinically relevant tasks (report generation, VQA, classification, risk prediction). This dataset includes eight out-of-distribution datasets to assess generalization and spanned seven distinct medical image modalities. An overview of the evaluation datasets is provided in Table~\ref{tab:validation-dataset-info}. The total number of evaluation samples across these datasets exceeded 40,000.

\subsection{Medical image classification}\label{sub:med-img-classification}
To rigorously evaluate \ourmodel's in-distribution and out-of-distribution performance, we employed a comprehensive medical image classification benchmark. This benchmark encompassed diverse modalities: skin lesion classification, chest X-ray classification, histopathology patch classification, and fundus image classification.  We approached classification as a generative multi-choice task for zero-shot classification (no supporting example in prompt) and linear probing for label-efficient setups. This design allowed for a thorough assessment of \ourmodel's robustness and adaptability across various medical imaging domains. 

\paragraph{Chest X-ray image classification} 
Our chest X-ray image classification evaluation focused on two key classification scenarios. First, we considered multi-label classification for the presence of each of five types of frequently occurring conditions: atelectasis, cardiomegaly, consolidation, pulmonary edema, and pleural effusion. This follows the suggestions from~\citet{tanno2024consensus,irvin2019chexpert,azizi2021big,azizi2023robust}.  Second, we performed binary classification for all images as either normal or abnormal, based on the CheXpert ``no finding'' label for frontal chest X-rays~\cite{irvin2019chexpert}. These two scenarios are used consistently across MIMIC-CXR and our out-of-distribution dataset CheXpert~\cite{irvin2019chexpert}. In addition, for ChestX-ray14~\cite{wang2017chestx,summers2019nih} we focused the evaluation of our model on three specific conditions including lung opacity, pneumothorax, and fracture. For all test tests, we only included images that were frontal view (i.e. view position ``AP'' or ``PA''). In addition, for MIMIC-CXR it is required the original report to contain a ``Findings'' section that could be extracted via regular expression matching.

We manually explored the validation set, prompting each model either with a multi-select prompt for all labels (\ie 5 conditions plus normal/abnormal) at once, or multiple binary Yes/No prompts for each label separately, and found binary prompts to yield better Macro F1 results for \ourmodel for all labels, and slightly better results for Gemini Ultra except for predicting normal/abnormal. The prompts used for evaluation are listed in Section~\ref{sub:vqa_and_class_eval_prompts}. Answers were generated using nucleus sampling with a temperature of 0.0, a top\_p of 0.75 and an output token limit of 200. Generated answers were normalized and matched against ``yes''/``no'' ground truth strings, which directly corresponded to 1.0 and 0.0 label values for all evaluation data sets. For multi-label, multi-class scenarios, we evaluated the average accuracy using the class-weighted F1 score. Details of the metrics used can be found in Section~\ref{sec-app-evaluation-metrics}. The MIMIC-CXR labels were revised based on a selective review of flagged reports by board-certified radiologists. See Section~\ref{sub:mimic_labels} for more details about the revised MIMIC-CXR labels. For the MIMIC CXR evaluations, we excluded case/condition combinations with an ``uncertain'' (-1.0) or no label (blank), except for the ``No Findings'' condition, where all cases with a non-positive (1.0) label were considered negative. Classification results using data-efficient learning are described separately in Section~\ref{sec:data-efficient-classification}.

Table~\ref{tab:results-cxr-classification} shows the comparison of the performance on the chest X-ray classification task between \ourmodel and \geminixl for in- and out-of-distribution datasets. Our medically tuned model outperformed \geminixl across most labels on the in-distribution MIMIC-CXR dataset. Notably, we demonstrated significantly stronger performance on the normal/abnormal classification despite using a multi-select prompt for Gemini, which specified all the ``abnormal'' conditions (and yielded better results on the validation set than a dedicated binary prompt), versus a very short normal/abnormal binary prompt for \ourmodel.  However, on the more challenging out-of-distribution datasets (CheXpert and ChestX-ray14), performance is varied.  \ourmodel excels in some tasks such as cardiomegaly or pleural effusion detection on CheXpert, while lagging in others like fracture detection in ChestX-ray14, which is a strong minority class there. These results suggest room for improvement in handling significant domain shifts.

\begin{table}[t]
\small
\centering
\caption{\small{\textbf{\ourmodeltwod Performance on the chest X-ray classification task.} Comparison of the performance on the chest X-ray classification task between \ourmodel, \geminixl and also SoTA, if comparable or available. For MIMIC-CXR datasets we utilized the revised labels (see Section~\ref{sub:mimic_labels}) denoted by~$\dagger$.  Our medically fine-tuned model demonstrated superior performance on the in-distribution MIMIC-CXR dataset. However, for out-of-distribution datasets, \ourmodel excelled in specific tasks such as Cardiomegaly detection, but both \geminixl and \ourmodel fell short of models exclusively trained for chest X-ray classification, and for strong minority class with distinctly different visual features such as Fracture. }}
\renewcommand{\arraystretch}{1.15}
\label{tab:results-cxr-classification}
\begin{tabular}{l|l|ccc|c}
\Xhline{2.5\arrayrulewidth}
\multirow{2}{*}{\textbf{Dataset}}          &\multirow{2}{*}{\textbf{Condition}}& \multicolumn{3}{c|}{\textbf{F1 score(\%)}}   &   \multirow{2}{*}{\textbf{OOD}} \\ \cline{3-5}

                           &                  & \textbf{\ourmodel} & \textbf{\geminixl}  &  \textbf{SoTA} & \\
\Xhline{2\arrayrulewidth}
\multirow{7}{*}{CheXpert}  & {Atelectasis}          & 49.7              & 51.1          & \textbf{64.6}$\ddagger$     & \multirow{7}{*}{\tick}\\
                           & {Cardiomegaly}         &         72.0      & 51.5          & \textbf{74.3}$\ddagger$    & \\
                           & {Consolidation}        &         23.0      & 17.9          & \textbf{33.3}$\ddagger$     & \\
                           & {Edema}                & 32.7              & 46.2          & \textbf{60.2}$\ddagger$     & \\
                           & {Pleural Effusion}     &         64.4      & 46.6          & \textbf{70.4}$\ddagger$     & \\ \cline{2-5}
                           & Macro-F1               &         48.3      & 42.6          & \textbf{60.6}$\ddagger$   & \\ \cline{2-5}
                           & {Normal/Abnormal}      & \textbf{54.3}     & 52.7          & -        & \\ \hline
\multirow{4}{*}{ChestX-ray14} & {Lung Opacity}      & 79.9     & 73.3 & \textbf{88.8}\textasteriskcentered & \multirow{4}{*}{\tick}\\
                           & {Pneumothorax}         & 55.3     & 29.3 & \textbf{58.4}\textasteriskcentered & \\
                           & {Fracture}             & 5.5      & 0    & \textbf{27.8}\textasteriskcentered & \\ \cline{2-5}
                           & Macro-F1 & \textbf 46.7           & 34.2 & \textbf{58.3}\textasteriskcentered & \\ \hline
\multirow{7}{*}{MIMIC-CXR} 
                           & \multirow{1}{*}{Atelectasis~$\dagger$}      & \textbf{99.8}     & 88.1             &   \multirow{6}{*}{\S} & \multirow{7}{*}{-}\\ 
                           & \multirow{1}{*}{Cardiomegaly~$\dagger$}     & 94.1              & \textbf{94.6}    &          & \\ 
                           & \multirow{1}{*}{Consolidation~$\dagger$}    & \textbf{82.0}     & 77.0             &           & \\ 
                           & \multirow{1}{*}{Edema~$\dagger$}            & \textbf{86.8}     & 86.4             &           & \\ 
                           & \multirow{1}{*}{Pleural Effusion~$\dagger$} & \textbf{90.8}     & 88.1             &           & \\ \cline{2-5}
                           & \multirow{1}{*}{Macro-F1~$\dagger$}         & \textbf90.7       & 86.8             &           & \\ \cline{2-5}
                           & \multirow{1}{*}{Normal/Abnormal}            & \textbf{42.0}     & 29.7             &           & \\
\Xhline{2.5\arrayrulewidth}
\end{tabular}
{
\\
\raggedright
\vspace{0.02in}
\footnotesize{\hspace{.3in}$\dagger$ Revised labels~\cite{mimic-cxr-gt}. \\}
\footnotesize{\hspace{.3in}$\ddagger$ Results from CheXzero model~\cite{tiu2022expert}.\\}
\footnotesize{\hspace{.3in}$\S$~SoTA not compatible with test set nor labels.\\}
\footnotesize{\hspace{.3in}$*$ Labels and prediction from~\citet{majkowska2020chest}. F1 scores computed from their PPV and sensitivity.\\}
}
\end{table}

\begin{figure}[t]
    \centering

    \subfloat[\centering {\footnotesize In-distribution}]{{\includegraphics[width=0.8\linewidth,trim=0 4 0 0]{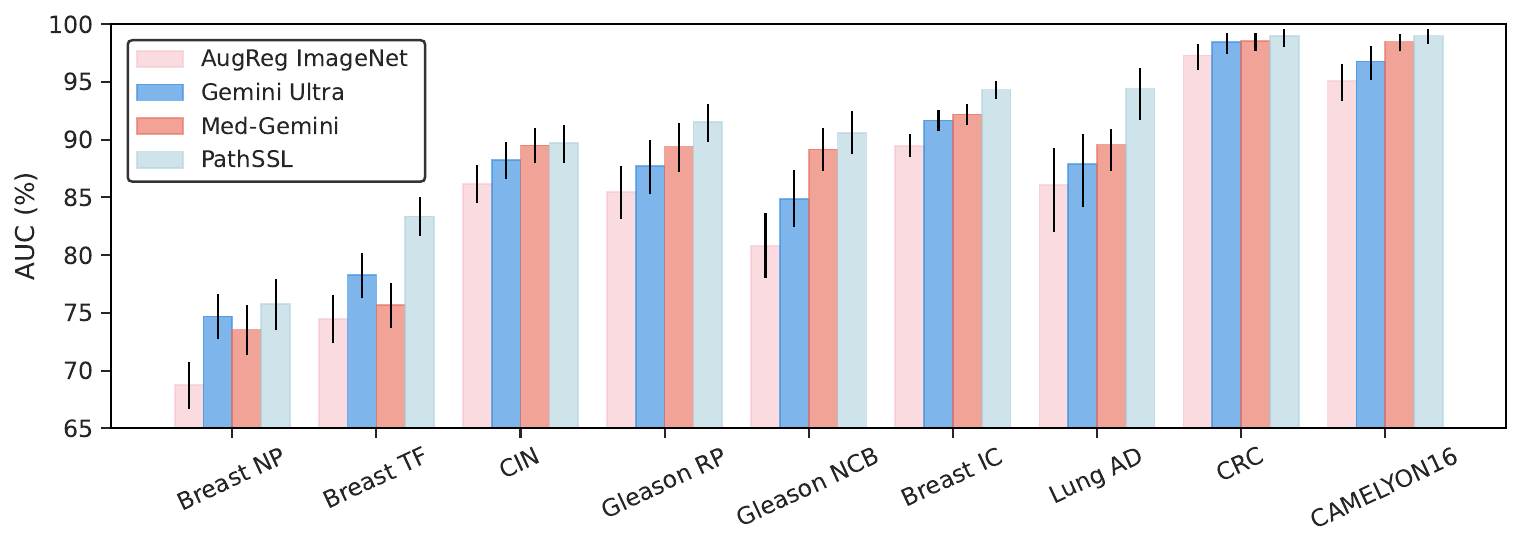}}}
    \subfloat[\centering {\footnotesize Out-of-distribution}]{{ \includegraphics[width=0.1845\linewidth]{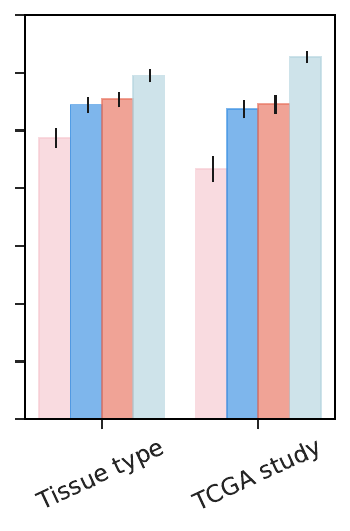}}}
    \vspace{-6pt}
    \caption{\small{\textbf{\ourmodeltwod histopathology image classification performance.} Linear-probing on histopathology patch-classification tasks (macro-averaged AUC percentages with 95\% confidence intervals) on our in-distribution and held-out out-of-distribution tasks. Overall, while \ourmodeltwod outperforms \basemodel Ultra on 7 out of 9 in-distribution tasks and on both out-of-distribution tasks, it does not improve over a histopathology-specific foundation model (PathSSL) on any of our tasks.}}
    \label{fig:classification-path}
\end{figure}

\vspace{-6pt}
\paragraph{Histopathology image classification}
We evaluated the image embeddings of \ourmodeltwod via linear probing on the 11 tasks from~\citet{lai2023domain} and summarized in Table~\ref{tab:histopathology-tasks}. \ourmodel was fine-tuned on data corresponding to 9 of these tasks (in-distribution), while 2 tasks were held-out (out-of-distribution). Together, these evaluation tasks cover a total of 17 tissue types, 12 cancer types, and several different types of classification tasks (\eg tumor identification, grading, subtyping) across 3 magnifications. Linear probing was done as in~\citet{lai2023domain}: briefly, a logistic regression model with L2-regularization was fit for each task, and task-specific regularization weights and magnifications were selected using the validation sets. Linear probe metrics on the test sets were calculated using 5,000 patches with logistic regression models trained on embeddings from the 10,000 train set plus 5,000 validation set patches. Confidence intervals for macro-averaged AUCs were computed via blocked bootstrap (blocking on slides) with 10,000 replicates. For comparison, we also report performance with embeddings from an ImageNet21k-based ViT-S/16 model trained using the AugReg method~\citep{steiner2021train}, embeddings produced by the vision encoder in Gemini Ultra, and embeddings from a histopathology-specialized model trained via self-supervision~\citep{lai2023domain} (PathSSL). Results are reported in Figure~\ref{fig:classification-path}. While the PathSSL embedding model is specialized to the histopathology domain, the image embeddings in \ourmodeltwod achieved comparable performance while also demonstrating strong results across multiple other clinical domains.

\paragraph{Skin lesion classification}

\begin{table}[t]
\small
\centering
\vspace{-6pt}
\caption{\small{\textbf{Performance on PAD-UFES-20 classification task.} AUC linear probing demonstrates that Gemini Ultra and Med-Gemini-2D generate robust skin lesion classification embeddings, with \ourmodeltwod further approaching the performance of the specialized Derm Foundation model.}}
\renewcommand{\arraystretch}{1.2}
\label{tab:results-skin-classification}
\begin{tabular}{p{6cm}|ccc}
\Xhline{2.5\arrayrulewidth}
\textbf{Metric}                       & \textbf{Weighted-AUC (\%)} & \textbf{Weighted-F1 (\%)} & \textbf{Accuracy (\%)} \\ 
\Xhline{2\arrayrulewidth}
\geminixl                             & \makecell{92.6 \\ \scriptsize{(89.8, 95.1)}}  & \makecell{60.3 \\ \scriptsize{(54.6, 66.9)}}           & \makecell{60.6 \\ \scriptsize{(54.1, 66.8)}}           \\
\hline
\ourmodeltwod                             & \makecell{92.1 \\ \scriptsize{(89.2, 94.7)}}           & \makecell{71.4 \\ \scriptsize{(65.4, 77.4)}}  & \makecell{73.3 \\ \scriptsize{(67.7, 78.9)}}  \\
\hline
Derm Foundation~\cite{derm2024google} & \makecell{\textbf{94.0} \\ \scriptsize{(91.6, 96.1)}}          & \makecell{\textbf{77.0} \\ \scriptsize{(71.4, 82.4)}}          & \makecell{\textbf{76.9} \\ \scriptsize{(71.3, 82.5)}}           \\
\Xhline{2.5\arrayrulewidth}
\end{tabular}
\end{table}

\ourmodeltwod achieves competitive classification accuracy using just dermatological images alone as input, and does not rely on metadata (e.g. patient demographics, lesion symptoms, living conditions). Such metadata, while provided in PAD-UFES-20, are not always readily available in clinical settings. We note that our evaluation is not directly comparable to Med-PaLM M \cite{tu2024towards} since (a) Med-PaLM M inputs an additional 14 clinical attributes and (b) we created different train and test splits to remove patient overlap between splits in the original Med-PaLM M work (we hope to publicly release these updated splits soon).
To establish context for model performance, we compared Gemini performance with Derm Foundation~\cite{derm2024google} which is a specialized dermatology model developed by Google. We trained linear probing classifiers on top of the Derm Foundation embeddings in this comparison.

We utilized the following three metrics for evaluation. (1) Weighted-AUC (by class prevalence): We first extracted the embedding outputs from \ourmodeltwod's image encoder, \geminixl's image encoder and Derm Foundation. Then we individually trained linear probes on top of the embeddings on the entire PAD-UFES-20 train split to classify 6 skin lesion types, and computes their weighted-AUC on the test split. (2) Weighted-F1 (by class prevalence): For \ourmodeltwod and \geminixl, we extracted classification prediction based on string matching from the model output. For Derm Foundation, we took the argmax of the linear probe prediction as the predicted class. (3) Accuracy: We computed the classification prediction follows the same method as in weighted-F1.

Table~\ref{tab:results-skin-classification} shows the performance comparison. From the AUC linear probing results, we can see that both \geminixl and \ourmodeltwod produced robust embeddings for skin lesion classification and had on-par performance when compared with the specialized Derm Foundation model. From F1 and accuracy, we can see that the fine-tuning in \ourmodeltwod improved the LLM's understanding of the embedding space for skin lesions and achieved performance close to the Derm Foundation model.

\paragraph{Fundus image classification}
We evaluated the performance of \ourmodeltwod on four ophthalmology classification tasks. First, we approached the identification of three common Diabetic Retinopathy (DR) lesions, more specifically hard exudates, hemorrhage, and panretinal photocoagulation (PRP) scars, as a multi-label classification challenge. Then, we treated anomaly detection as a binary classification problem, where the model was tasked with determining the presence or absence of DR lesions in a fundus image. EyePACS dataset \cite{cuadros2009eyepacs} was used for all tasks, with fundus images balanced for equal distribution of positive and negative labels.

We benchmarked \ourmodeltwod against \geminixl, which used both the fundus image and a multiple-choice-like prompt for prediction. To extract the prediction labels, we searched for specific markers that the LLM was constrained to output (e.g. (G) if no DR lesion is present). In contrast, \ourmodeltwod relied only on the image, with prediction labels extracted by searching for keywords, like ``hemorrhage.'' For the anomaly detection task, we compare to a third model similar to \cite{krause2018grader} that has been trained using supervised learning to detect different grades of DR (none, mild, moderate, severe, proliferative). If the predicted DR grade is ``none,'' the fundus is considered normal (no DR lesion), otherwise, the image is predicted to have DR lesions present. We note that this supervised model was carefully trained on a much larger dataset containing more than 3 million fundus images from diverse manufactures/data sources/geography, and could be considered as an ``upper bound'' of this task.

\begin{table}[t]
\centering
\small
\caption{\small{\textbf{Performance comparison of \ourmodeltwod, \geminixl, and a supervised model that is trained with additional data for fundus image classification.} For all tasks, \ourmodeltwod demonstrates significant improvement over \geminixl. For lesion presence classification, \ourmodeltwod is on-par with the supervised model for sensitivity, while specificity still have room for improvement (likely due to orders of magnitude less data used during the training process).}}
\label{tab:ophto-classification}
\renewcommand{\arraystretch}{1.2}
\begin{tabular}{l|l|c|c|c|c}
\Xhline{2.5\arrayrulewidth}
\textbf{Condition} & \textbf{Model} & \makecell{\textbf{Accuracy}\\\textbf{(\%)}} & \makecell{\textbf{Sensitivity}\\\textbf{(\%)}} & \makecell{\textbf{Specificity}\\\textbf{(\%)}} & \makecell{\textbf{F1 score}\\\textbf{(\%)}} \\ 
\Xhline{2\arrayrulewidth}

\multirow{2}{*}{Hard exudates}      & \geminixl  & 55.2 & 71.5 & 39.0 & 61.5 \\ \cline{2-6}
                                    & \ourmodeltwod         & \textbf{88.4} & \textbf{80.3} & \textbf{96.4} & \textbf{87.3} \\ \hline
\multirow{2}{*}{Hemorrhage}         & \geminixl         & 49.0 & 69.9 & 28.1 & 57.8 \\ \cline{2-6}
                                    & \ourmodeltwod         & \textbf{84.7} & \textbf{88.4} & \textbf{81.1} & \textbf{85.3} \\ \hline
\multirow{2}{*}{PRP Scars}          & \geminixl         & 57.2 & 56.8 & 57.6 & 57.0 \\ \cline{2-6}
                                    & \ourmodeltwod         & \textbf{84.6} & \textbf{71.6} & \textbf{97.6} & \textbf{82.3} \\ \hline
\multirow{3}{*}{DR lesions present} & \geminixl         & 58.6 & 73.6 & 43.8 & 63.9 \\ \cline{2-6}
                                    & \ourmodeltwod         & 84.9 & 96.3 & 73.5 & 86.4 \\ \cline{2-6}
                                    & \makecell[l]{Supervised model trained \\ with additional data} & \textbf{94.7} & \textbf{96.8} & \textbf{92.6} & \textbf{94.9} \\
\Xhline{2.5\arrayrulewidth}
\end{tabular}
\end{table}

Table~\ref{tab:ophto-classification} displays the performance of \ourmodeltwod and \geminixl on the classification of hard exudates, hemorrhages and PRP Scars, utilizing accuracy, sensitivity, specificity, and F1 score as evaluation metrics. It also compares the performance of DR lesions detection between, \ourmodeltwod, \geminixl and the strong supervised model. Results demonstrate that \ourmodeltwod consistently outperformed \geminixl on both multi-label and binary classification tasks. Notably, \ourmodeltwod achieved significantly higher specificity in hard exudate classification (96.4\% vs. \geminixl’s 39.0\%) and hemorrhage classification (81.1\% vs. \geminixl’s 28.1\%). These results highlight the benefits of task-specific fine-tuning for this highly-specialized medical domain. \ourmodeltwod underperformed on the anomaly detection task compared to the strong supervised model, but it is important to acknowledge the strong supervised model's significant advantage in the $\sim200\times$~volume of labeled data used during its training process.

For three other attempted classification tasks, namely detection of microaneurysms, neovascularization of the optic disc and neovascularization elsewhere, \ourmodeltwod appeared to be miscalibrated, predicting most cases as negative in the LLM text output. While this is likely related to the training dataset distribution and overall data mixing ratio, further work is needed to improve question-answering based classification and calibration.

\subsection{Visual question answering (VQA)}
We assessed \ourmodeltwod's performance on VQA tasks across a range of diverse medical specialties, including radiology, dermatology, and pathology, and spanning a wide range of open-ended and closed-ended questions. Table~\ref{tab:results-vqa} summarizes the overall VQA results. Prompt templates were manually optimized for each model and VQA dataset on the validation splits, and are listed in Table~\ref{tab:vqa_eval_prompts}. Model answers were generated using the same method and parameters as for CXR classification, see Section~\ref{sub:med-img-classification}. That is, the generative sampling was not constrained by a given test vocabulary in any manner, as it was in related work \cite{li2023self,li2023masked,zhang2023biomedgpt}, typically only to the test set's ground truth answers, for the reasons described e.g. in~\citet{tu2024towards,van2023open}. In other words, answers were generated in a truly generative, open-ended, and zero-shot manner. 

\begin{table}[t!]
\centering
\caption{\small{\textbf{Evaluation details for VQA tasks.}  Our \ourmodeltwod's VQA performance was evaluated across diverse medical specialties using various datasets. Notably, in radiology tasks, \ourmodeltwod outperformed \basemodel and previous best-in-class performance across different subsets and metrics, improving best-in-class performance by over 11\%. Performance is reported using Mean Tokenized F1-score, Mean Expert Score, and Accuracy. Based on radiologist assessments, we excluded several cases from the VQA-Rad dataset due to questions being deemed unanswerable from the provided images. For VQA-Rad we report the accuracy for the balanced test split and the test split suggested by~\citet{xu2023elixr}. In pathology, it showed reasonable performance on this useful, albeit noisy dataset, as measured by accuracy for Yes/No questions and average tokenized F1-score for zero-shot responses.}}
\label{tab:results-vqa}
\renewcommand{\arraystretch}{1.18}
\small
\begin{tabular}{@{\hspace{0.1em}}l@{\hspace{0.5em}}|c@{\hspace{0.5em}}c|c|c|c@{\hspace{0.1em}}}
\Xhline{2.5\arrayrulewidth}
\textbf{Dataset} & \textbf{Subset} & \textbf{Metric(\%)} & \textbf{\ourmodel}              &  \textbf{\geminixl} &  \textbf{SoTA} \\
\Xhline{2\arrayrulewidth}
\multirow{11}{*}{VQA-Rad} 

&\multirow{3}{*}{\makecell{Test set from\\\citet{xu2023elixr}\\(CXR only)}} & \begin{tabular}[c]{@{}c@{}}Expert score\\(Excluding 12 cases)\end{tabular}          & \textbf{71.9} &  -    &\begin{tabular}[c]{@{}c@{}}57.9\\\cite{xu2023elixr}\end{tabular} \\ 
&                                  & \begin{tabular}[c]{@{}c@{}}Accuracy\\(Closed only)\end{tabular}                & \textbf{78.8} & -  &\begin{tabular}[c]{@{}c@{}}67.1\\\cite{xu2023elixr}\end{tabular} \\ \cline{2-6} 

&\multirow{3}{*}{\makecell{Balanced Test\\(CXR only)}}  
                                    & \begin{tabular}[c]{@{}c@{}}Expert score\\(Excluding 4 cases)\end{tabular}     & \textbf{71.8}   &  -   &  \begin{tabular}[c]{@{}c@{}}55.7$^\ddagger$\\\cite{xu2023elixr}\end{tabular}   \\
&                                  & \begin{tabular}[c]{@{}c@{}}Tokenized F1\end{tabular}                           & \textbf{62.6}   &   49.0  & - \\
&                                  & \begin{tabular}[c]{@{}c@{}}Accuracy\\(Closed only)\end{tabular}                & \textbf{78.1}   &   62.4  &   - \\ \cline{2-6}

&\multirow{3}{*}{\makecell{Balanced Test\\(CXR, CT, MR)}}   
                                   & \begin{tabular}[c]{@{}c@{}}Expert score\\(Excluding 44 cases)\end{tabular}     & \textbf{61.9}   &  -    &  -  \\
&                                  & \begin{tabular}[c]{@{}c@{}}Tokenized F1\end{tabular}                           & \textbf{50.1}   &   46.4  &  - \\
&                                  & \begin{tabular}[c]{@{}c@{}}Accuracy\\(Closed only)\end{tabular}                & \textbf{69.7}   &   64.9  &  - \\ 
\Xhline{2\arrayrulewidth}

\multirow{3}{*}{Slake-VQA}   
&\multirow{3}{*}{\makecell{Official Test\\(English only)}}   
                                   & \begin{tabular}[c]{@{}c@{}}Tokenized F1\end{tabular}                           & 75.8  & 44.6 & \begin{tabular}[c]{@{}c@{}}\textbf{89.3}$^\dagger$\\\cite{tu2024towards}\end{tabular}\\
&                                  & \begin{tabular}[c]{@{}c@{}}Accuracy\\(Closed only)\end{tabular}                & 84.8  & 70.4 & \begin{tabular}[c]{@{}c@{}}\textbf{91.1}\\\cite{li2023self}\end{tabular} \\                                  

\Xhline{2\arrayrulewidth}   

\multirow{3}{*}{PathVQA}   
&\multirow{3}{*}{\makecell{Official Test}}
                                  & \begin{tabular}[c]{@{}l@{}}Tokenized F1\end{tabular}                            & 58.7  & 34.9 &\begin{tabular}[c]{@{}c@{}}\textbf{62.7}$^\dagger$\\\cite{tu2024towards}\end{tabular} \\
&                                 & \begin{tabular}[c]{@{}c@{}}Accuracy\\(Closed only)\end{tabular}                 & 83.3  & 62.8   & \begin{tabular}[c]{@{}c@{}}\textbf{90.9}\\\cite{sun2024pathasst}\end{tabular} \\ 
\Xhline{2\arrayrulewidth}

\multirow{3}{*}{\makecell{MIMIC-CXR \\VQA}}
& \multirow{2}{*}{\makecell{Test split from\\ \citet{xu2023elixr}}} 
                                  & Tokenized F1                                                                    & \textbf{52.5}  & 44.8  & - \\  
&                                 & \begin{tabular}[c]{@{}c@{}}Accuracy\\(Closed only)\end{tabular}                 & \textbf{78.6}  & 70.9  & \begin{tabular}[c]{@{}c@{}}68.1\\\cite{xu2023elixr}\end{tabular}  \\
\Xhline{2.5\arrayrulewidth}
\end{tabular}

{\raggedright
\vspace{0.04in}
\scriptsize{
$\dagger$ Using one-shot prompt with a text-only exemplar~\cite{tu2024towards}. \\
$\ddagger$ Average of original scores over new, smaller test set. \\
} }

\end{table}

For close-ended questions, we measured accuracy based on exact matches of normalized model vs ground truth answers, and compare against SOTA based on vocabulary-constrained answer generations, since the freely generated answers mostly matched the overall set of ground truth answers. For open-ended questions, we report the average token-wise F1 score~\cite{tu2024towards} between the normalized answers of the model and the ground truth, and only compare against SOTA results where answers were generated in the same zero-shot manner. In addition, for \ourmodeltwod results on VQA-Rad, one board-certified radiologist scored the answers using the 3-point scoring rubric introduced in~\citet{xu2023elixr}, in order to compare against results of the prior ELIXR model.

We assessed \ourmodeltwod's VQA capabilities in radiology using three datasets from distinct domains. First, we evaluated on the MIMIC-CXR VQA test set, an in-distribution benchmark containing 226 question-answer pairs for 48 chest X-ray images suggested by~\citet{xu2023elixr}.  Second, we used the English-only 1,061 question and answer pairs in the test split of Slake VQA, a large bilingual (English and Chinese) VQA dataset. Finally, we employed the VQA-Rad dataset, leveraging the new three-way balanced split detailed in Section~\ref{sec:data}.  As discussed previously, to evaluate out-of-distribution performance and facilitate a head-to-head comparison with the previous best-in-class model (ELIXR), we did not fine-tune our model on either the VQA-Rad training images or questions.  This approach ensures both ELIXR and our model are tested on the same, larger VQA-Rad test set~\cite{xu2023elixr}. Moreover we evaluated the performance of our model on both chest X-ray only and all modality (CT, MRI, and X-ray). 

As Table~\ref{tab:results-vqa} demonstrates, \ourmodeltwod outperformed many previous results and \basemodel across different subsets and metrics. Specifically, in the chest X-ray only (ELIXR split) subset, our model achieved a remarkable expert-evaluated accuracy score of 71.9 and an accuracy of 78.8 in closed-ended questions, improving the best-in-class number by 14 and 11.7, respectively. In the chest X-ray-only balanced split subset, our model maintained strong performance with an expert-evaluated accuracy of 71.8 improving best-in-class number by 16 and an accuracy of 78.1 in closed-ended questions. Moreover, across all modalities in the balanced split, our model achieved competitive results and improved over \basemodel, demonstrating its versatility. In the MIMIC-CXR VQA dataset, it achieved an accuracy of 78.6\% on Yes/No questions and a tokenized F1 score of 52.5 overall. In the Slake VQA dataset, our model significantly outperformed \basemodel and achieves performance close to state-of-the-art with 84.8 accuracy on close-ended questions, showcasing its capability across different domains. Its mean tokenized F1-score across all English questions at 75.8 is lower than for the MedPaLM-M model~\cite{tu2024towards} at 89.3, which might be partially attributable to it being prompted in a zero-shot manner, versus a one-shot text-only prompt for the latter. Contrary to MedPaLM-M, \ourmodeltwod was not fine-tuned with one-shot examples, hence this prompting technique would yield worse results during inference.

To evaluate \ourmodeltwod's VQA capabilities in  pathology, we utilized the PathVQA dataset~\cite{he2020pathvqa}. For this dataset, our model achieved an accuracy of 83.3 at Yes/No questions, and a tokenized F1-score of 58.7 over all questions which improves over \basemodel. Its overall zero-shot results are slightly below those of the MedPaLM-M model~\cite{tu2024towards}, which reports an overall tokenized F1-score of 62.7, albeit employing a text-only one-shot prompting technique here as well. This technique involved an additional exemplar question-and-answer pair along with an image placeholder string <img> provided as a one-shot example during evaluation. However, while these results can provide a general sense of VQA capabilities for images comprising both histopathology and general anatomic pathology photographs and diagrams, given the known issues with QA pairs and image quality in this auto-generated dataset~\cite{lu2024visual}, we suggest cautious interpretation.

We also conducted a qualitative review of model behavior for histopathology and radiology VQA tasks. Examples are shown in Figure~\ref{fig:example-1-2}, and Figure~\ref{fig:example-pathology} .

\subsection{Report generation for chest X-rays} \label{report_gen_2d}
In clinical practice, the role of the radiologist extends far beyond narrow interpretation of radiology images. Radiologists are tasked with conveying nuanced findings within a broader clinical context, synthesizing information, and providing recommendations for patient care. Expert radiologists use natural language to articulate this synthesis of imaging findings, overall impressions, and recommendations in written reports. Unlike some prior work, our model was tuned for the difficult task of generating both the `FINDINGS' and `IMPRESSION' sections of chest X-ray reports for frontal view chest radiographs (anterior-posterior or posterior-anterior), covering comprehensively both the observations and inferences typically made by radiologists during a study.

\begin{table}[t!]
\centering
\small
\caption{\small{\textbf{Evaluation details report generation in chest X-rays.} \ourmodeltwod sets a new standard for AI generated chest X-ray (CXR) report generation based on expert evaluation, exceeding previous best results across two separate datasets.}}
\label{tab:report-eval-2d}
\renewcommand{\arraystretch}{1.3}
\begin{tabular}{@{\hspace{0.0em}}ll@{\hspace{0.2em}}c@{\hspace{0.2em}}@{\hspace{0.1em}}c@{\hspace{0.0em}}}
\Xhline{2.5\arrayrulewidth}
\textbf{Dataset}   & \textbf{Metric} & \textbf{\ourmodel} & \textbf{SoTA} \\
\Xhline{2\arrayrulewidth}
\multirow{3}{*}{MIMIC-CXR} 
&         AI superior or similar to original report (Normal)      &     \textbf{57\%}     &  45\% \cite{tanno2024consensus}              \\
&         AI superior or similar to original report (Abnormal)    &     \textbf{43\%}     &  42\% \cite{tanno2024consensus}              \\
&         Clinically acceptable AI generated report (All)        &     72\%              &  -                 \\
\Xhline{2\arrayrulewidth}
\multirow{3}{*}{IND1} 

&         AI superior or similar to original report (Normal)      &     \textbf{96\%}    &  85\% \cite{tanno2024consensus}              \\
&         AI superior or similar to original report (Abnormal)    &     \textbf{65\%}    &  53\% \cite{tanno2024consensus}              \\
&         Clinically acceptable AI generated report (All)         &     88\%             &  -                 \\
\Xhline{2.5\arrayrulewidth}
\end{tabular}
\end{table}

Table~\ref{tab:mimic-report-all-metrics} presents the performance comparison of various models in generating radiology reports for chest X-rays using the publicly available MIMIC-CXR dataset. The ``Sections'' column indicates whether the model generates the `FINDINGS' (`F') or `IMPRESSIONS' (`I') section of the report, with metrics drawn from published research. Higher values in all metrics indicate superior performance. Notably, our model undertakes the more challenging task of generating both sections (F + I) for frontal chest X-rays, aiming to capture the radiologist's holistic interpretation of the study. 

Following common practice, we leveraged the established n-gram based methods such as ROUGE-L, BLEU-4 to evaluate the generated reports quality against the ground-truth. Additionally, we measured the RadGraph F1-score, which is the F1 score between the entities extracted from the reference report and generated one using RadGraph~\cite{jain2021radgraph}. RadGraph accounts for not only the absence or presence of findings in the report, but also their relationships to image features. \ourmodel achieved a RadGraph F1-score of 24.4\%, marking a notable improvement of 3.9\% compared to the previous top-performing model.

For the IND1 dataset we did not evaluate automated metrics, as automated metrics such as RadGraph F1-score are specifically trained on MIMIC-CXR to measure performance of US-style chest X-ray report and are not capable of handling the out-of-distribution format of IND-1 dataset reports obtained in an India-based clinical setting.

\begin{table}[t!]
\centering
\small
\caption{\small\textbf{Automated report generation metrics on the MIMIC-CXR dataset.} This table presents the performance of various models on generating radiology reports for chest X-rays using the publicly available MIMIC-CXR dataset. The \emph{Sections} column indicates whether the model generates the FINDINGS (F) or IMPRESSION (I) section of the report, with metrics sourced from published research. For all of the metrics higher is better. Bold values highlight the best results in each category for (F + I) methods. Notably, our model tackles the more challenging task of creating both sections (F + I) for frontal chest X-rays (anterior-posterior or posterior-anterior views), aiming to capture the radiologist's comprehensive interpretation of the study. \ourmodel achieved a RadGraph F1-score of 24.36\% on chest X-ray report generation, demonstrating a 4.0\%+ improvement over the previous best-in-class score.}
\label{tab:mimic-report-all-metrics}
\renewcommand{\arraystretch}{1.15}
\begin{tabular}{l|c|ccc|c}
\Xhline{2.5\arrayrulewidth}
\multirow{2}{*}{\textbf{Model}}       & \multirow{2}{*}{\textbf{Sections}} & \multicolumn{3}{c|}{\textbf{NLG Metrics(\%)}} & \multicolumn{1}{c}{\textbf{Clinical Metrics(\%)}} \\ [0.2em] \cline{3-6} 
                             &                         & CIDEr    & BLEU4    & Rouge-L   & RadGraph F1-score     \\ [0.2em]\Xhline{2\arrayrulewidth}
CXR-RePaiR \citep{endo2021retrieval}                   & F & -     & 2.1  & 14.3 & 9.1 \\
$\mathcal{M}^2$ Transformer \citep{miura2020improving} & F & 50.9  & 11.4 & -    & 22.0 \\
RGRG \citep{tanida2023interactive}                     & F & 49.5  & 12.6 & 26.4 & - \\
METransformer \citep{wang2023metransformer}            & F & 36.2  & 12.4 & 29.1 & - \\
Med-PaLM M, 12B \citep{tu2024towards}                  & F & 23.4  & 10.4 & 26.2 & 25.2 \\
Med-PaLM M, 84B \citep{tu2024towards}                  & F & 26.2  & 11.3 & 27.3 & 26.7 \\ 
MAIRA-1~\cite{hyland2023maira}                         & F & -     & 14.2 & 28.9 & 24.3 \\\Xhline{2\arrayrulewidth}
R2Gen \citep{chen2020generating}                       & F + I & - & 10.3 & 27.7 & 13.4 \\
WCT \citep{yan2021weakly}                              & F + I & - & 14.4 & 0.274 & 14.3 \\
CvT-21DistillGPT2 \citep{nicolson2023improving}        & F + I & \textbf{36.1} & 12.4 & 28.5 & 15.4 \\
BioVil-T \citep{bannur2023learning}                    & F + I & - & 9.2 & 29.6 & - \\
R2GenGPT \citep{wang2023r2gengpt}                      & F + I & 26.9 & 13.4 & \textbf{29.7} & - \\

Flamingo-CXR \citep{tanno2024consensus}                & F + I & 13.8 & 10.1 & \textbf{29.7} & 20.5 \\\Xhline{2\arrayrulewidth}
\ourmodeltwod                                              & F + I & 17.5  & \textbf{20.5} & 28.3           & \textbf{24.4}  \\
\Xhline{2.5\arrayrulewidth}
\end{tabular}
\end{table}

\paragraph{Human evaluation rubric for report generation} For report generation we devised a novel evaluation rubric, expanding on those used in Flamingo-CXR~\cite{tanno2024consensus} and Med-PaLM M~\cite{tu2024towards}, to understand potential impact on clinical management of patients. The evaluation rubric consists of six categories that compare two reports. It provides an improved granularity around patient impact and was used for both the CXR and CT generated reports. Table~\ref{tab:human_report_rubric} defines the labels for comparing the AI and original radiologist reports for the same study. This rubric along with training materials and examples were provided to radiologist labelers as training material, prior to any labeling. In each example seen by labelers, the origin of the reports (AI vs. original) was masked and the reports were shown in random order to avoid bias.

\begin{table}[htp]
\small
\centering
\caption{\small{\textbf{Human evaluation rubric comparing AI generated radiology reports to original reports.}}}
\label{tab:human_report_rubric}
\definecolor{A2}{RGB}{87,172,255}
\definecolor{A1}{RGB}{194,226,237}
\definecolor{Csame}{RGB}{222,222,222}
\definecolor{B1}{RGB}{255,221,227}
\definecolor{B2}{RGB}{255,138,117}
\definecolor{Xneither}{RGB}{135,135,135}

\begin{tabular}{p{2.5cm}p{13cm}} 
\toprule
\textbf{Rubric Score} & \textbf{Rubric Definition}  \\
\hline

\hline \rowcolor{A2}
\multirow{2}{*}{{\textbf{A2}}}  &   \small{Report A captures key clinically relevant findings that are not found in B. Report A would result in correct patient management and report B would not.}        \\ \hline \rowcolor{A1}
\multirow{2}{*}{\textbf{A1}}  & Report A captures more relevant findings, but both would result in the same correct patient management.   \\ \hline \rowcolor{Csame}
\multirow{2}{*}{\textbf{C}}   & \RaggedRight{Both reports capture similar findings in the image and would result in correct patient management.}  \\ \hline \rowcolor{B1}
\multirow{2}{*}{\textbf{B1}}  & Report B captures more relevant findings, but both would result in the same correct patient management. \\ \hline \rowcolor{B2}
\multirow{2}{*}{\textbf{B2}}  & Report B captures key clinically relevant findings that are not found in A. Report B would result in correct patient management and report A would not.  \\ \hline  \rowcolor{Xneither}
\multirow{1}{*}{\textbf{X}}   & Neither report would result in correct patient management.  \\
\bottomrule
\end{tabular}
\end{table}

\begin{figure}[t]%
    \centering
    \subfloat[\centering MIMIC-CXR subset of 306 cases]{{\includegraphics[width=\linewidth,trim=0 10 0 0]{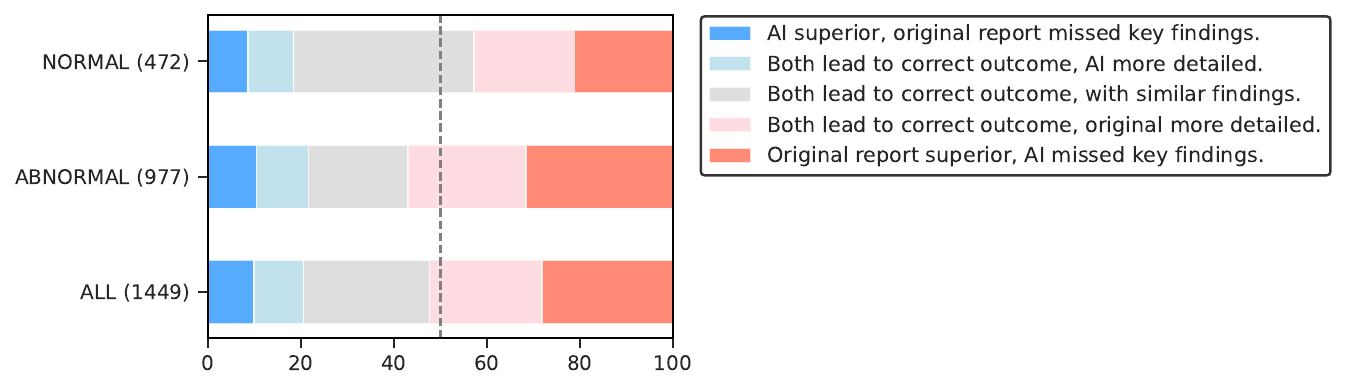}}} \\ \hfill%
    \subfloat[\centering IND1 subset of 300 cases]{{\includegraphics[width=\linewidth,trim=0 10 0 0]{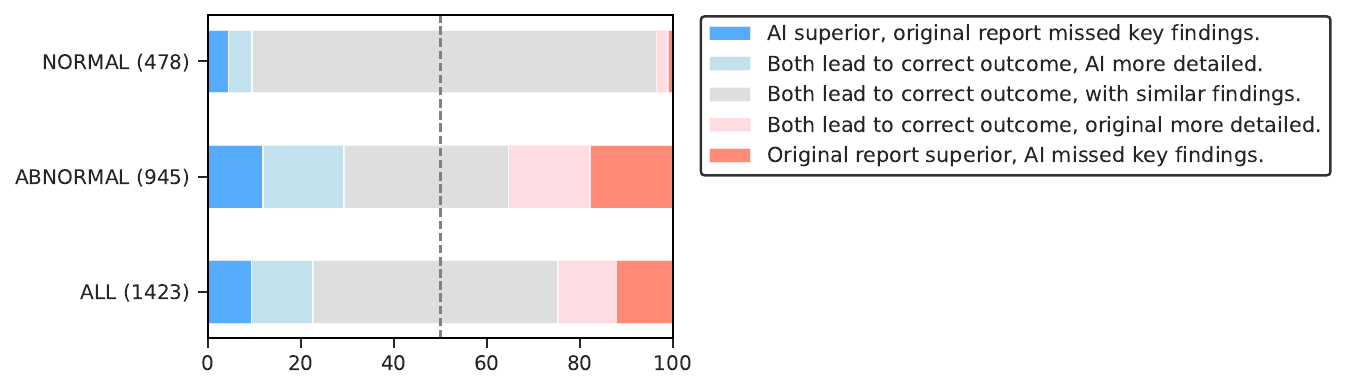}}}%
    \caption{\small{\ourmodeltwod CXR report generation results based on 4 India-based board-certified radiologists and one US-based academic board-certified radiologist performing report comparisons. \textbf{(a)} a subset of 306 MIMIC-CXR cases. Overall 48\% of the cases were equal or superior to the original reports and 72\% lead to the same clinical outcome. \textbf{(b)} a subset of 300 cases from IND1. Overall 75\% of the reports were equal or superior to the original reports. In both studies, the AI performance was better on the normal cases.}}%
    \label{fig:apollo_mimic_cxr}%
\end{figure}

Five India-based board-certified radiologists, one India-based thoracic specialist, and one US-based academic thoracic radiologist evaluated a total of 606 cases: 306 from the MIMIC test dataset and 300 from IND1. After the study completion, readers were compared using their mean Quadratic Kappa~\cite{sim2005kappa} to the two thoracic specialists. Two readers falling below 0.2, i.e. ``none to slight agreement'' were eliminated from the final results. The results were computed based on the total sum of categories for the selected reports after elimination of scores of the \textbf{X} category. The percentage of cases within each category were then plotted sequentially along a horizontal plot for all, abnormal, and normal cases as shown in Figure~\ref{fig:apollo_mimic_cxr} and summaries are shown in Table~\ref{tab:report-eval-2d}

In examining cases that fell into the \textbf{A1} and \textbf{B1} categories, i.e. where one report captures clinical findings but both would result in the same patient management, similar reasoning was given in both categories. These included missing less critical findings and descriptiveness of findings. Examples of missed findings include: mild cardiomegaly, calcified granulomas, and old fractures. In terms of descriptiveness, examples include: better descriptions of bulla, proper identification of devices, and clearly discerning mass versus pneumonia and other less explicit diagnoses. Reports falling into categories \textbf{A2} and \textbf{B2} missed key findings, including: failures in assessing tube positions, missed nodules, and missed pneumothraces. 



\subsection{Report generation for head/neck CT volumes} \label{report_gen_3d}

3D imaging modalities often involve more complex data preparation and longer radiologist interpretation time in comparison to 2D images such as X-rays, making the paired image-text data required for generative AI modeling scarcer and more expensive. Additionally, radiology reports tend to be much longer and imaging features much sparser for 3D images than for 2D images. Given this relative data scarcity and information complexity (with correspondingly increased memory requirements), end-to-end modeling to convert 3D radiology images to text reports has previously been infeasible. With its increased computational capacity and extensive domain-specific pretraining, \ourmodelthreed, building on other recent generative AI work such as~\citet{hamamci2024ct2rep}, is the first LLM-based generative AI model able to interpret a 3D medical imaging modality end to end from the CT volume to text.

Using the same human evaluation rubrics introduced in Section~\ref{report_gen_2d}, we evaluated a total of 92 non-contrast head/neck CT studies consisting of 27 Normal-labeled cases without findings and 65 Abnormal-labeled cases that contained findings, including both acute findings such as cerebrovascular accidents as well as findings that are common consequences of aging, such as atrophy. Studies were initially divided into normal and abnormal candidates based on the length of the impression sections of the reports. A random subset within each was selected and then manually classified by a board-certified radiologist into the normal or abnormal category based on the full radiology report. Studies classified as normal contained no findings.

In reviewing the reports, a single academic board-certified examined the study and all series using a web-based Picture Archiving and Communication System (PACS) viewer. The radiologist graded the two reports using the same rubric presented for evaluating CXR reports. For each rating, the radiologist also recorded a comment describing why the rating was given. The model generated the report based on a single series with the most slices and did not have access to any of the other series. The model was given the patient history in the form of text during inference.

\begin{figure}[t]
    \centering
    \includegraphics[width=\linewidth]{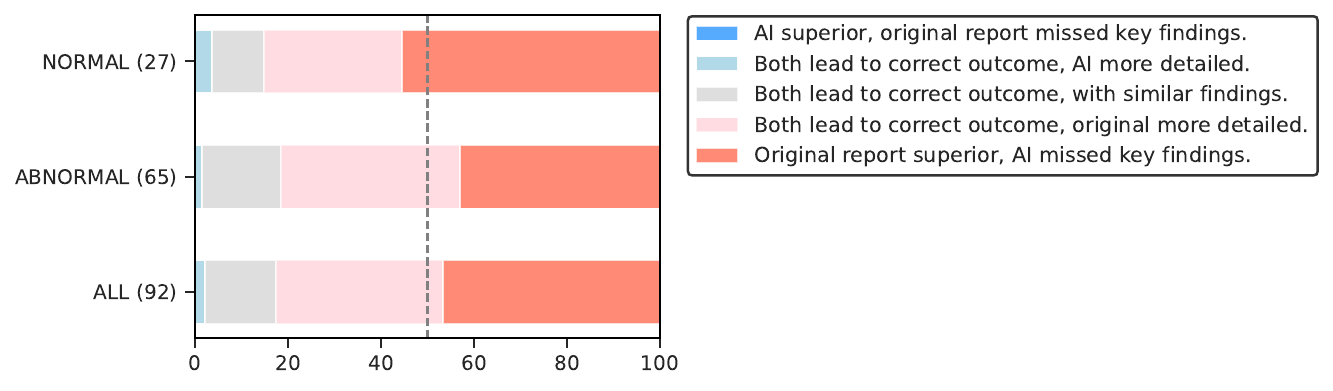}
    \caption{\small{\ourmodelthreed Head CT report generation scores on a subset of 92 cases from the CT-US1 test set scored by a US-based board certified radiologist.  Across all cases, 17\% of the AI generated reports were graded equivalent or superior to that of radiologists' reports, while 53\% were judged as resulting in equivalent patient care. Performance was better overall on abnormal versus normal cases.}}
    \label{fig:ct_results}
\end{figure}

Results are shown in Figure \ref{fig:ct_results} and Table \ref{tab:report-eval-ct-us-comparison}. We found that 45\% of AI reports on normal studies and 57\% of AI reports on abnormal studies would have resulted in the correct clinical management of the patient, though some of those AI reports included errors that would not directly affect management. We did find, however, that only 17\% of AI reports were considered to be of equivalent or better quality than the original radiologist reports. In examining the notes on errors from the generated reports, \ie those that were scored \textbf{B2}, roughly half involved missed findings while the other half involved hallucinations such as identified subdural hematomas or cysts. In terms of \textbf{B1} category reports, comments about the generated report mention it either incorrectly estimates or under-characterizes white matter changes.

While our early results presented here leave significant room for future improvement, the potential opportunity for AI in volumetric imaging is vast.  This difficulty in reporting on volumetric data can result in concerning diagnostic delays~\cite{nhsctscans}. The ability to safely triage, expedite, and quality check existing reports could be highly beneficial in health systems around the world.

\begin{table}[t!]
\centering
\small
\caption{\small\textbf{Human evaluation results for Head CT Volume report generation} Note there is no existing best-in-class performance for this task as report generation from Head/Neck 3D CT volumes is a new capability. Additionally, the model had access to a single series in the study for report generation.}
\label{tab:report-eval-ct-us-comparison}
\renewcommand{\arraystretch}{1.5}
\begin{tabular}{ll|c}
\Xhline{2.5\arrayrulewidth}
\textbf{Dataset}   & \textbf{Metric} & \textbf{\ourmodel} \\
\Xhline{2\arrayrulewidth}
\multirow{3}{*}{CT-US1}
&         AI superior or similar to original report (Normal) &     15\%             \\
&         AI superior or similar to original report (Abnormal) &     18\%             \\
&         Clinically acceptable AI generated report (All)       &     53\%                       \\
\Xhline{2.5\arrayrulewidth}
\end{tabular}
\end{table}

\subsection{Disease prediction from genetic information}
Personalized medicine can benefit greatly from genetics, as disease risks depend heavily on an individual's genetic makeup. To leverage this powerful information, we expanded our model's ability to process genetic information in the form of an RGB image by featurizing the genome into polygenic risk scores (PRS) as explained in Section~\ref{sec:data}.

\begin{figure}[t]
    \centering
    \includegraphics[width=1\linewidth]{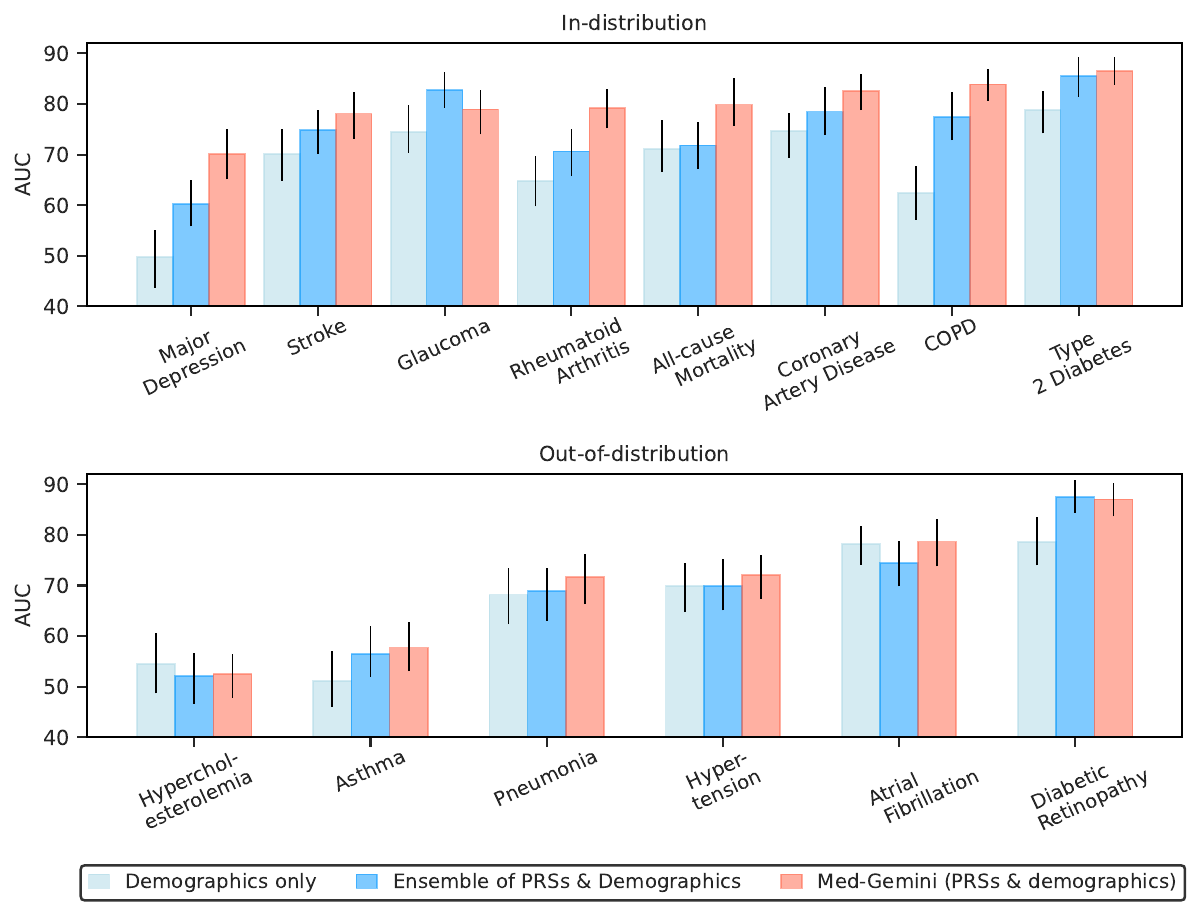}
    \caption{\small{Health outcome prediction using \ourmodelpolygenic compared to two baselines for both in-distribution and out-of-distribution outcomes. ``Demographics only'' used a linear probe of \texttt{age}, \texttt{sex}, and \texttt{BMI} to predict each health outcome, and ``Ensemble of PRSs and demographics'' combined demographics with all 7,145 PRSs in a linear probe. \ourmodelpolygenic was prompted with both an individual's PRS image and demographics. For out-of-distribution health outcomes, the linear probes (``Ensemble of PRSs and demographics'' and ``Demographics only'') were trained to predict the most-correlated in-distribution outcome (Table \ref{tab-app:genomics_ood_correlations}), and those predictions were then evaluated on the out-of-distribution outcome.}} 
    \label{fig:genomics_mosaic_vs_baselines}
\end{figure}

To assess the disease risk prediction capability of \ourmodelpolygenic, we created benchmarks by training linear models on all PRS featurizations plus demographics (``Ensemble of PRSs and demographics'') which is the current best practice for using PRSs for disease prediction \cite{Alb2023multiprs,Truong2024integrativeprs}. For in-distribution health outcomes (see Section \ref{sec:data}) used in \ourmodelpolygenic training, we directly applied the trained ``Ensemble of PRSs and demographics'' models as benchmarks. For health outcomes that were never used in the training process (out-of-distribution or OOD, see Section \ref{sec:data}) but share some genetic correlation with the in-distribution outcomes, we first calculated their phenotypic correlations with all the in-distribution outcomes, and used the model trained to predict the most correlated in-distribution outcome to generate a maximally strong performance benchmark.

We evaluated \ourmodelpolygenic performance on case/control balanced datasets sampled from the test split (200 cases, 200 controls per outcome) for computational efficiency (Section \ref{sec-app:PRS}). We obtained a disease probability score from \ourmodelpolygenic by prompting it to predict the status of a given health outcome using a text prompt and the genetic risk ``image'' (Table \ref{tab-app:genomics_prompt_example}), and computed the probability as the ratio of the likelihoods of the model generating a positive and negative prediction. \ourmodelpolygenic achieved higher AUCs than the PRS linear model benchmarks for all in-distribution health outcomes except glaucoma (Figure \ref{fig:genomics_mosaic_vs_baselines}). To evaluate zero-shot generalization ability, we prompted \ourmodelpolygenic to predict disease status for the six out-of-distribution health outcomes. \ourmodelpolygenic achieved similar performance to benchmarks trained on the most correlated in-distribution outcome (Table \ref{tab-app:genomics_ood_correlations}) despite never being instructed about the associations between in-distribution and out-of-distribution outcomes (Figure \ref{fig:genomics_mosaic_vs_baselines}).

 Additionally, we compared the performance of linear probes of the \ourmodelpolygenic embeddings and directly prompting \ourmodelpolygenic in the evaluation sets of 400 individuals. Comparisons of the AUCs show that while \ourmodelpolygenic performs similarly to the linear probe when each is only given demographic information, it often outperforms the linear probe when incorporating both PRSs and demographics (Figure \ref{fig-app:genomics_coca_embeddings_vs_mosaic}). This performance increase is largely attributable to \ourmodelpolygenic modeling non-linear interactions between genomic information and demographics (Table \ref{tab-app:genomics_nonlinear}).

We caution that the AUC values reported here represent an upper bound on model performance since the GWASs used to create the PRS features were performed within the UK Biobank. However, the relative performance of different models that all operate on this in-sample data is the measure of interest for these analyses.

\section{Qualitative Results}
\begin{figure}[htp]
    \centering
    \includegraphics[width=1.0\linewidth]{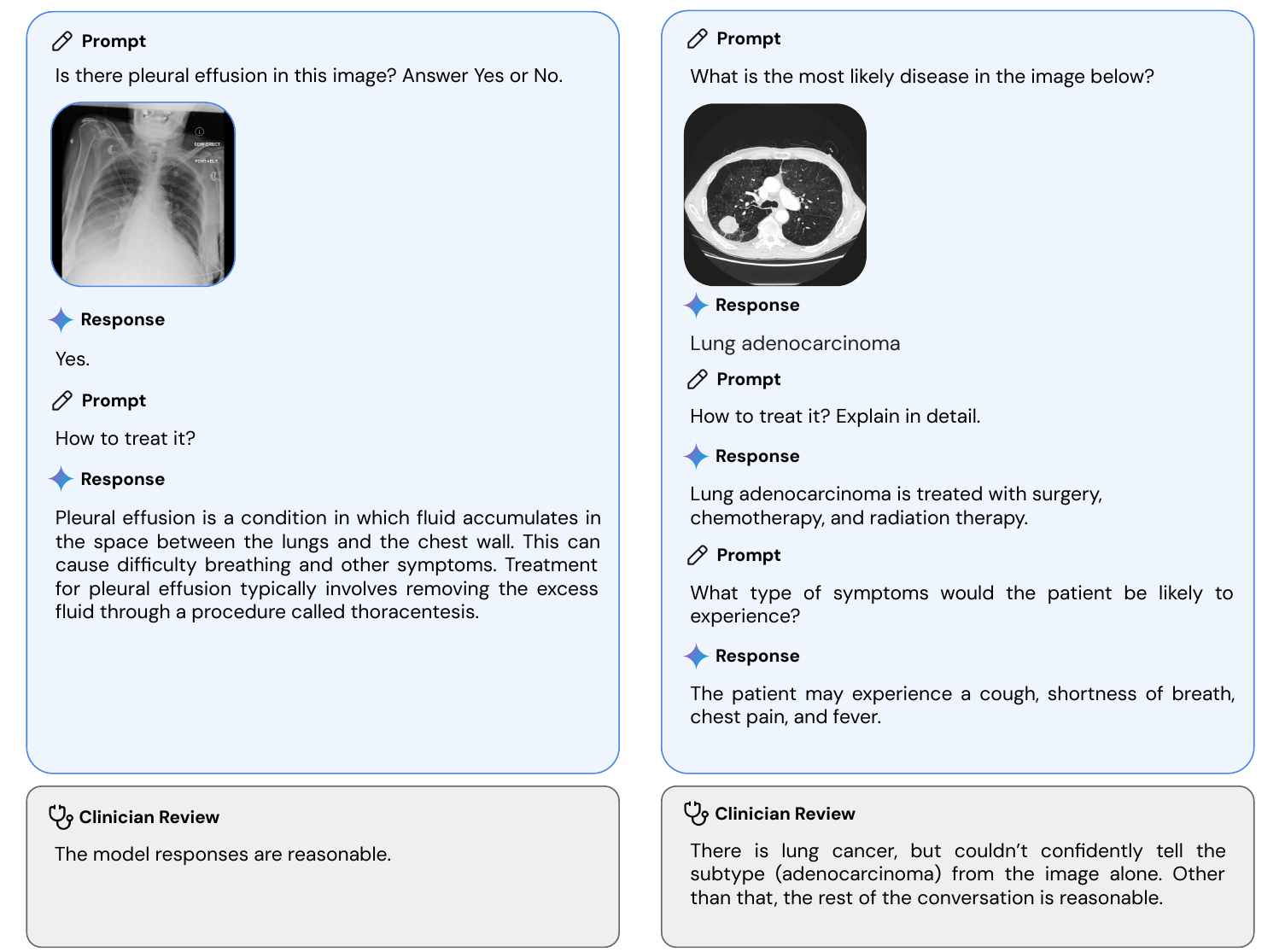}
    \includegraphics[width=1.0\linewidth]{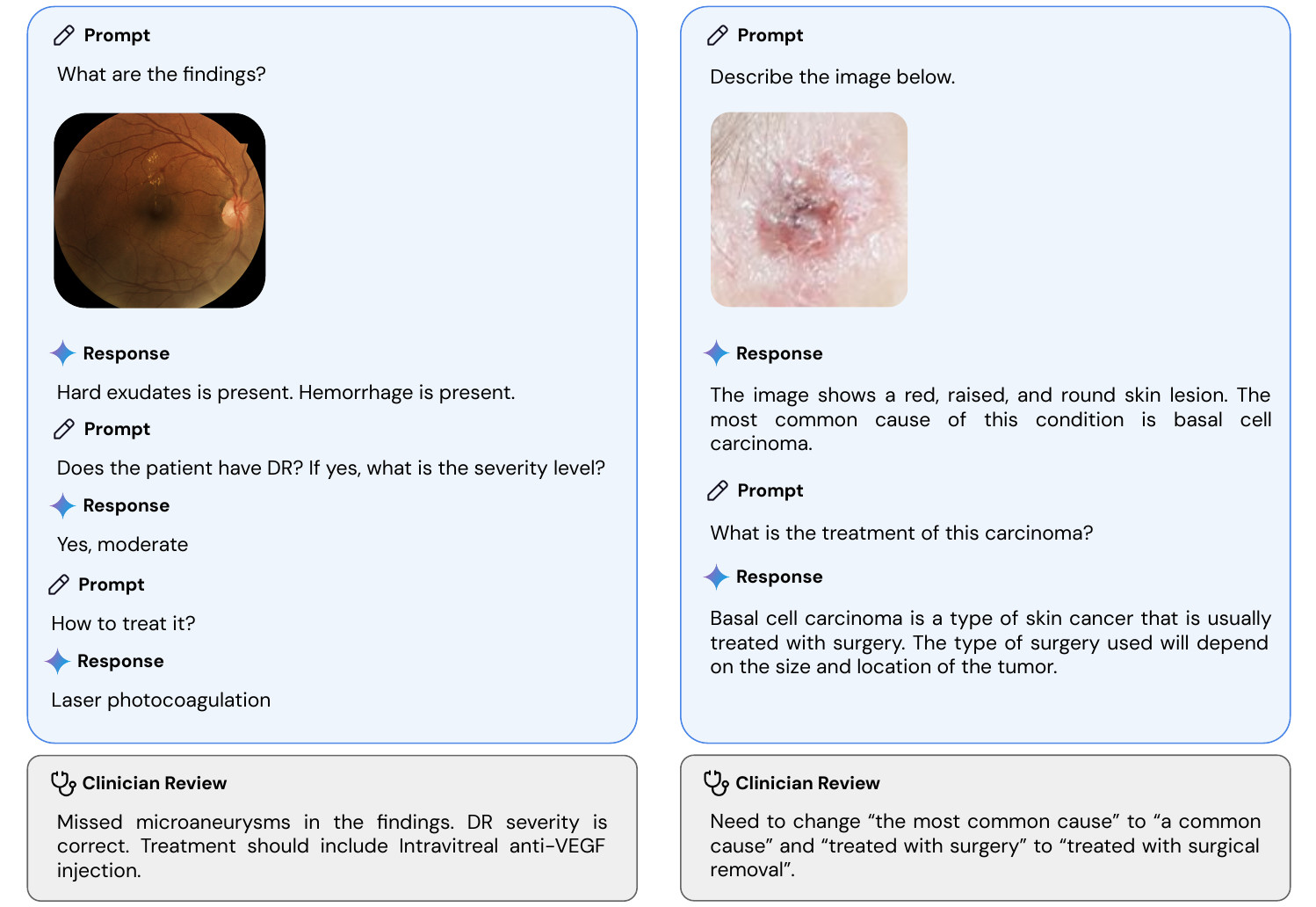}
    \caption{\small{\textbf{Example of 2D medical image dialogue via open ended question answering.} For chest X-ray~\cite{johnson2019mimicdatabase}, lung CT~\cite{liu2021slake}, fundus images \cite{cuadros2009eyepacs}, and skin lesion images~\cite{pacheco2020pad}.}}
    \label{fig:example-1-2}
\end{figure}

\begin{figure}[htp]
    \centering
    \includegraphics[width=\linewidth]{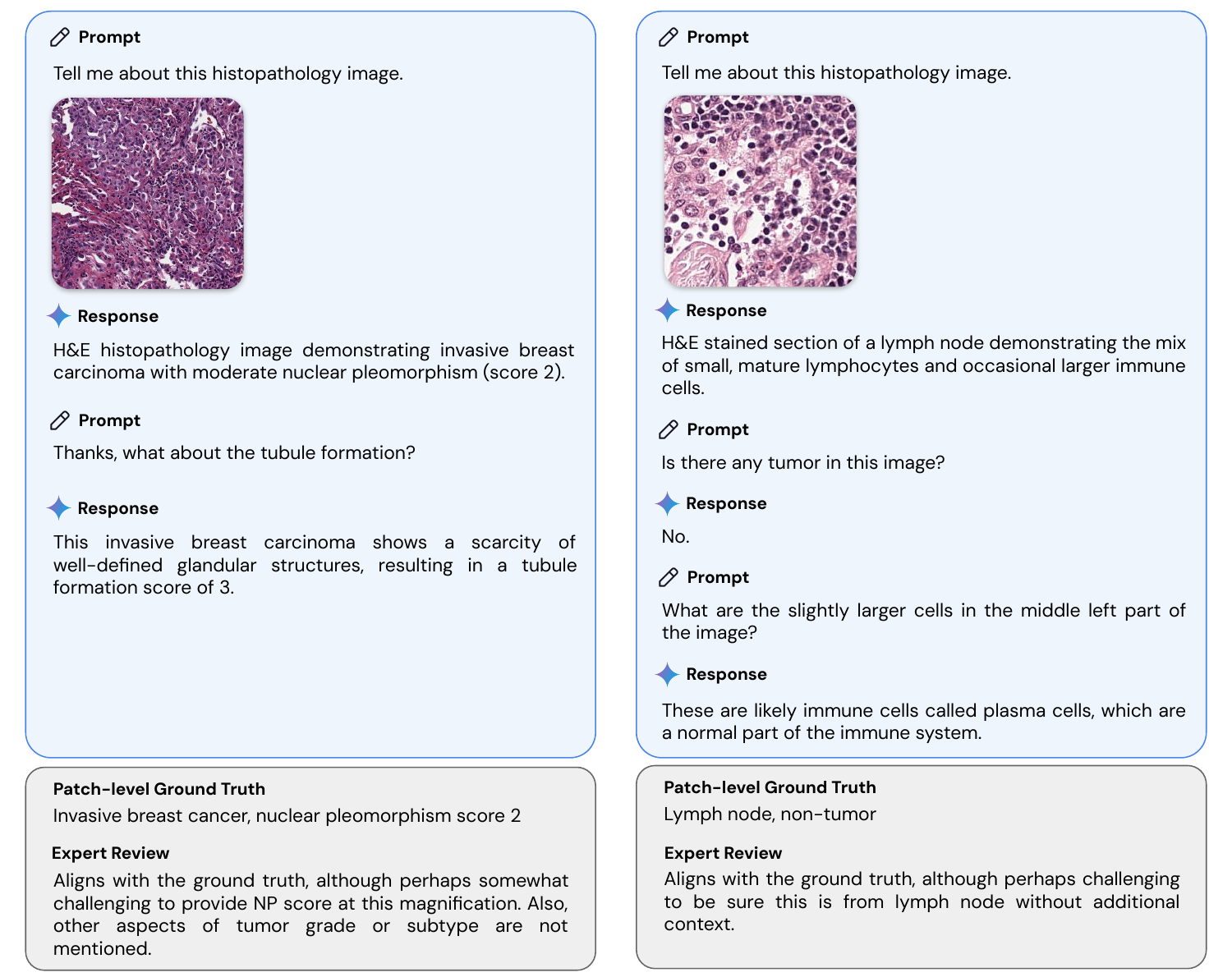}
    \includegraphics[width=\linewidth]{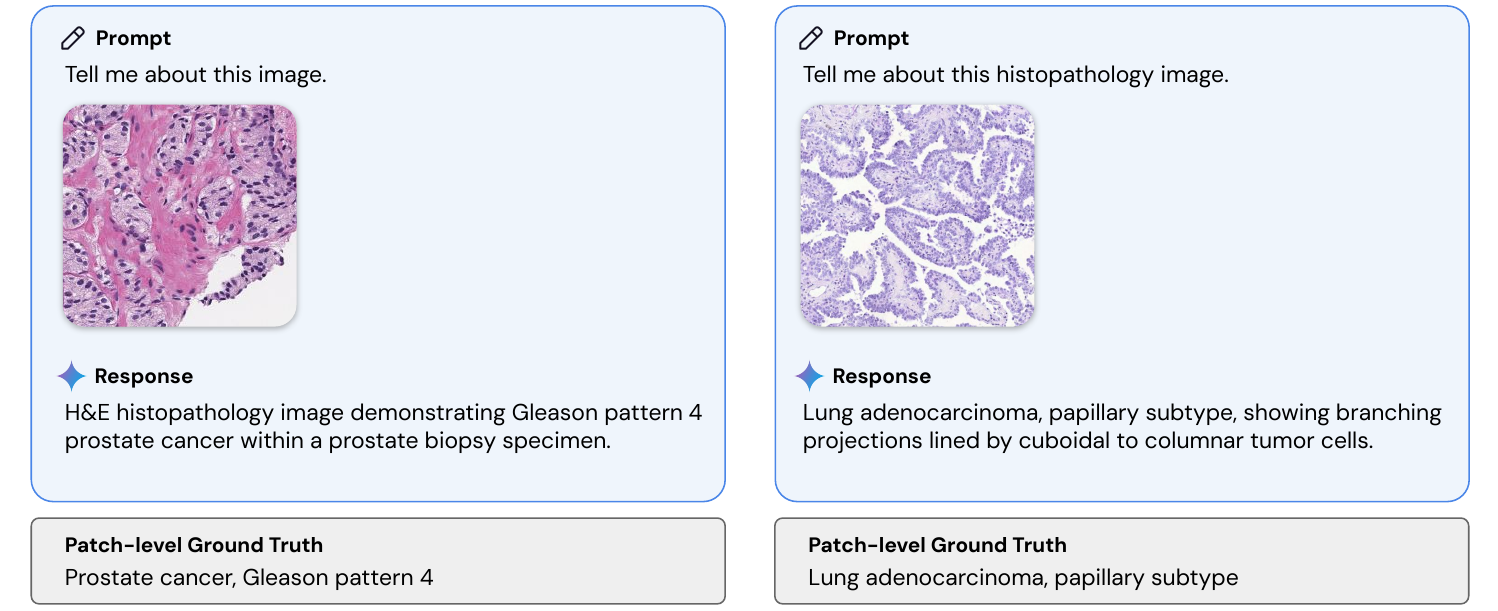}
    \vspace{3pt}
    \caption{\small{\textbf{Examples of histopathology image-based dialogue.} These examples highlight accurate histopathology interpretation and communication of information across a range of tissues and findings with only a small amount of visual context. While there is some promising initial evidence for ability to interact and reason further about images (top right), current capabilities for follow up interaction and exploration of input images are limited and this remains an active area for improvement. This demonstration utilizes histopathology patches from TCGA, CAMELYON16, and other data sources as described in~\citet{lai2023domain}.}}
    \label{fig:example-pathology}
\end{figure}

\begin{figure}[htp]
    \centering
    \includegraphics[width=\linewidth]{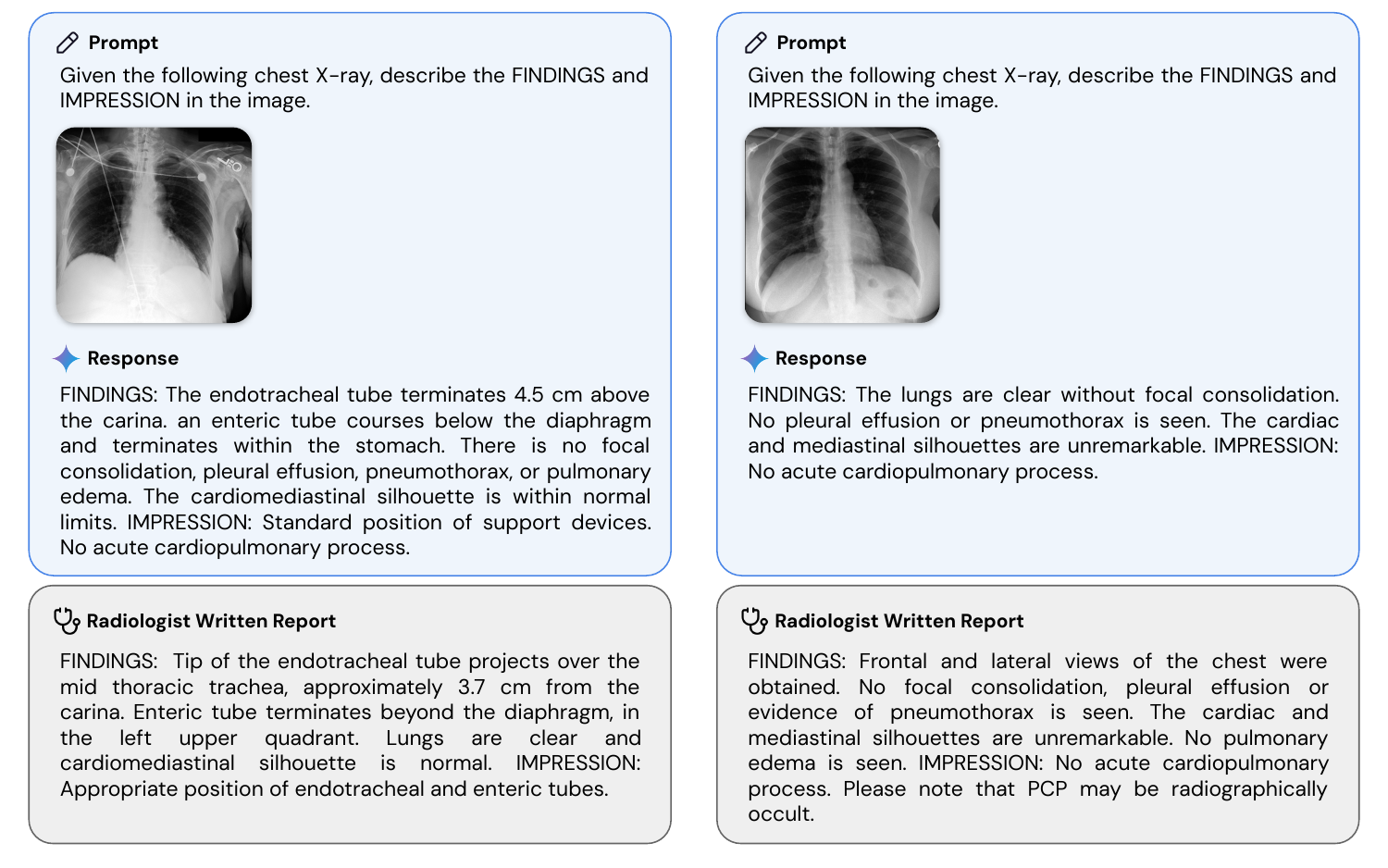} \vspace{6pt}
    \includegraphics[width=\linewidth]{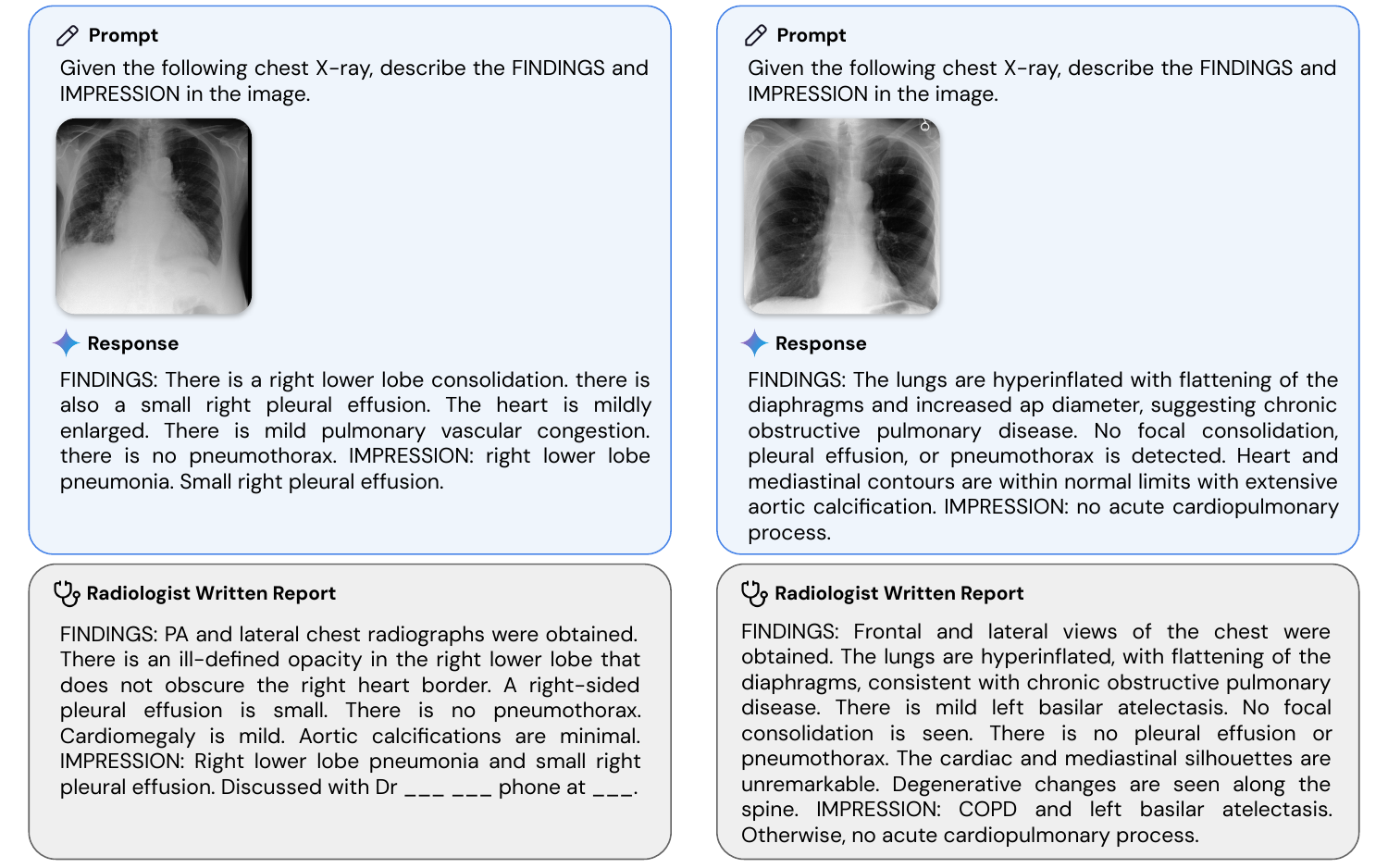}
    \caption{\small{\textbf{Examples of chest X-ray report generation.} These examples demonstrate the capability of \ourmodel for CXR report generation on various conditions. Top-left, support devices; Top-right, normal case; Bottom-left, acute abnormality; Bottom-right, chronic abnormality. A radiologist reviewed all these examples and confirmed model generated reports are reasonable with one note for the bottom-right case where the ``increased ap diameter'' is usually detected from a lateral image, and thus a hallucination from the model.}}
    \label{fig:example-cxr-report}

\end{figure}

\begin{figure}[htp]
    \centering
    \includegraphics[width=\linewidth]{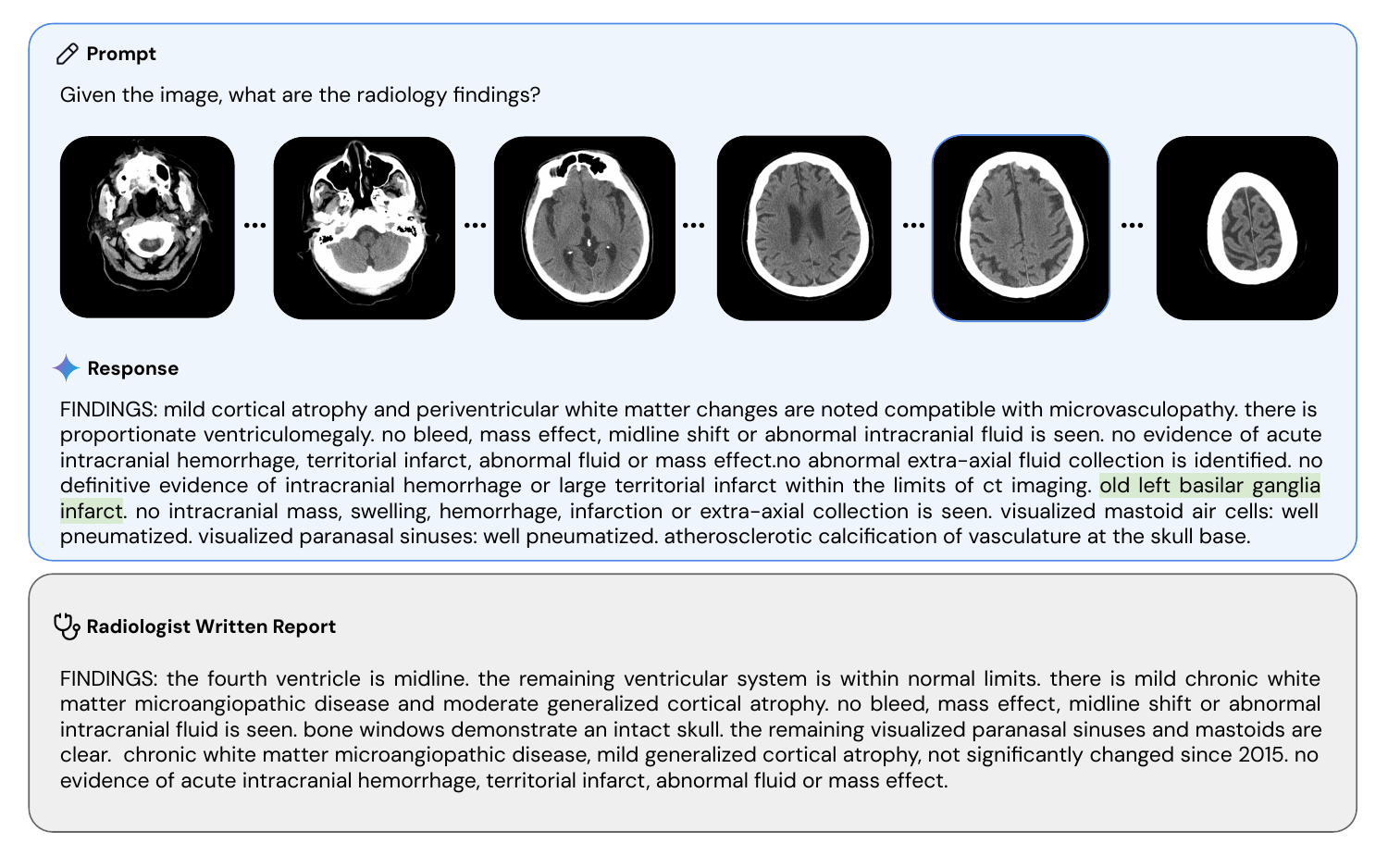}
    \includegraphics[width=\linewidth]{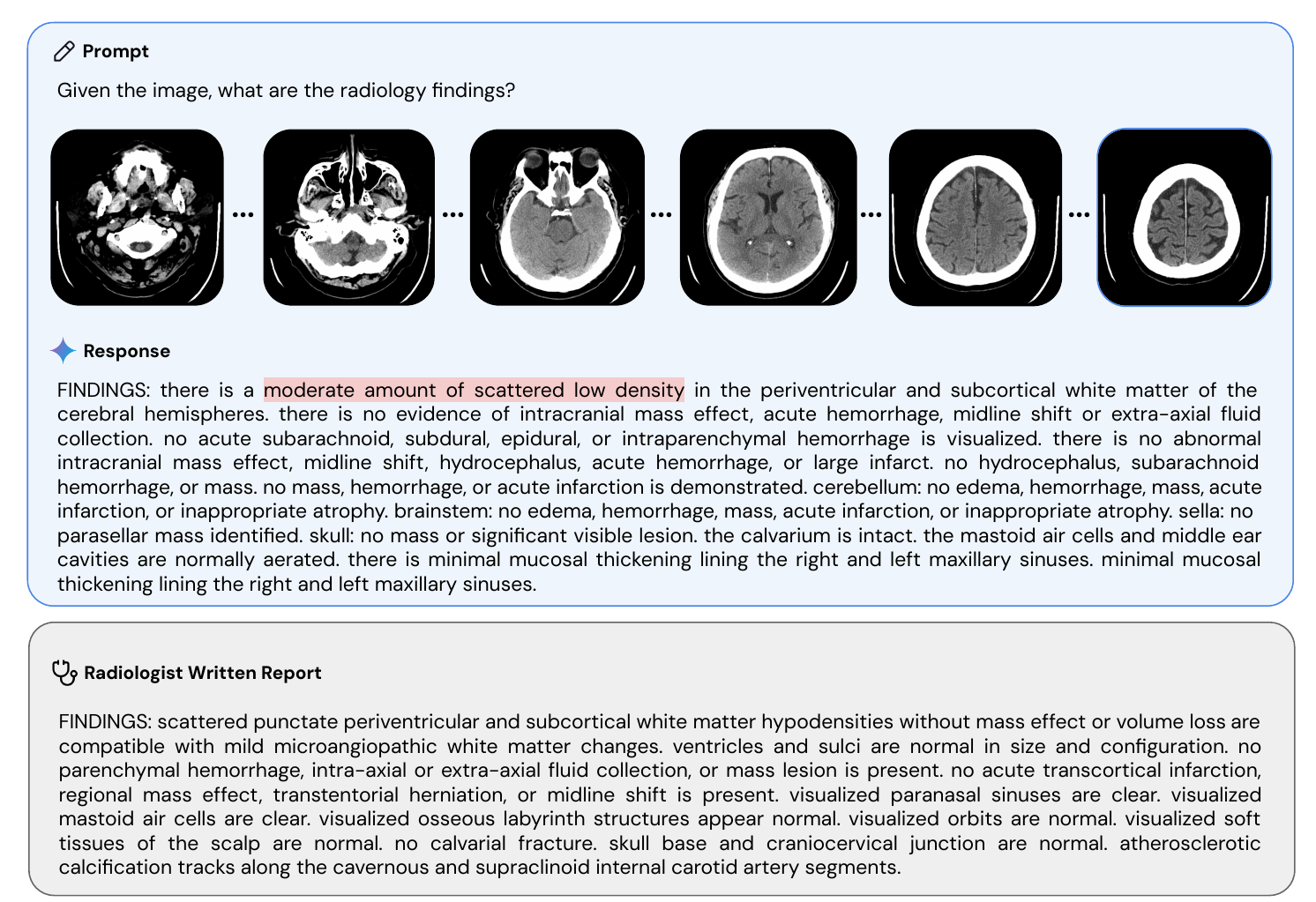}
    \caption{\small{\textbf{Examples of 3D Head CT report generations.} These examples showcase 3D medical image dialogue for Head CT report generation: (top) correct abnormal case, (bottom) incorrect abnormal case.  While \ourmodel can identify some abnormalities missed by radiologist generated reports (highlighted in green), it can also mischaracterize findings that are present (highlighted in red) or hallucinate findings that are absent from the image.}}
    \label{fig:example-3d-scan}
\end{figure}

In this section we provide a few examples showcasing our model's capability in medical dialogue for diverse set of medical modalities including chest X-ray, CT, fundus, dermatology, pathology, depicted in Figures~\ref{fig:example-1-2},~\ref{fig:example-pathology} as well as 2D (Figure~\ref{fig:example-cxr-report}) and 3D (Figure~\ref{fig:example-3d-scan}) radiology report generation. As highlighted in these examples, \ourmodel is able to provide accurate and reasonable multimodal dialogue and interpretation capabilities across a variety of medical imaging domains. At the same time, expert review of these examples highlights areas for improvement regarding the phrasing, accuracy, appropriate level of detail, and completeness of generated responses. 

\begin{figure}[htp]
    \centering
    \includegraphics[width=\linewidth]{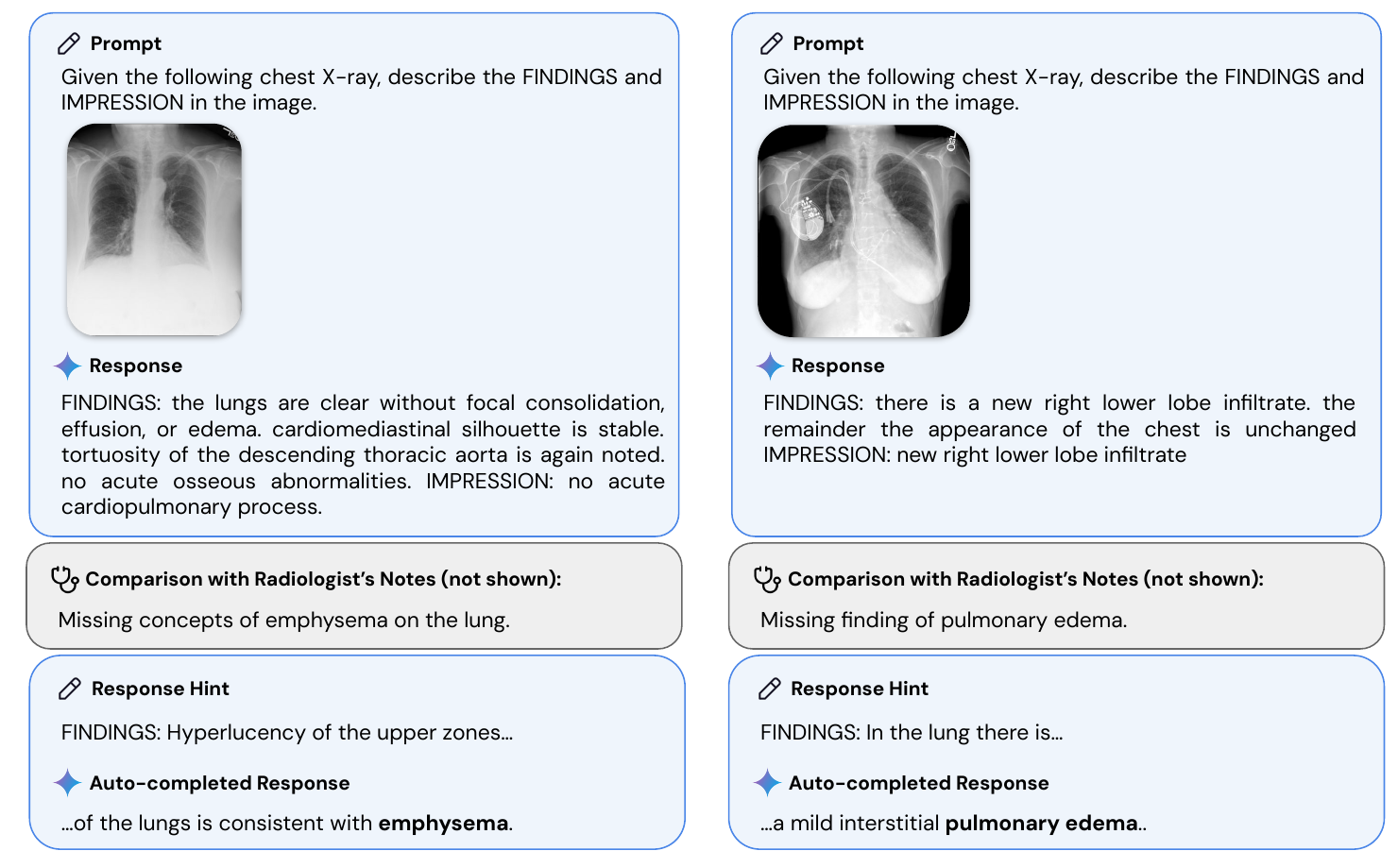} \caption{\small{\textbf{Examples of chest X-ray report autocompletion.} In these examples, particular concepts were missing from the report generated without any hint, and were recovered with the autocomplete prefix hint. A) Emphysema, B) Pulmonary Edema. 
    }}
    \label{fig:example-cxr-report-autocomplete}
\end{figure}

In addition to understanding automated report generation capabilities, it is important to consider plausible real world assistive use cases. As a proof of concept, we experimented with directing the model’s attention to a specific region/organ within the CXR (Figure \ref{fig:example-cxr-report-autocomplete}).

Lastly, as shown in the above examples, even though \ourmodel was only fine-tuned with data directly related to image interpretation (e.g. there were no question-answer pairs related to treatments or symptoms in the fine-tuning set), \ourmodel can still leverage the medical knowledge from Gemini pretraining to give simple but reasonable answers to those questions. While we emphasize that real-world medical diagnosis, prognosis, and treatment information is much more complicated and nuanced than the examples provided here, these examples serve as a proof of concept for combining large model pretraining with domain specialization, an active area for further improvements.

\section{Related Work}
\label{sec:related}
\paragraph{The evolution of medical language models}
Large language models (LLMs) built on Transformer architectures~\cite{parmar2018image,vaswani2017attention} have seen rapid advancement, driving significant progress in natural language processing and multimodal modeling.  Pathway scaling methods~\cite{barham2022pathways} have been crucial in enabling the development of ever-larger models like the PaLM family including PaLM, PaLM 2, and PaLM-E~\cite{anil2023palm,chowdhery2023palm,driess2023palm}. Other significant LLMs include BERT~\cite{devlin2018bert}, GPT family~\cite{radford2019language, brown2020language,achiam2023gpt}, T5~\cite{raffel2020exploring}, and LLaMA~\cite{touvron2023llama}, Hyena~\cite{poli2023hyena}, Mistral 7B~\cite{jiang2023mistral}. LLMs are often refined through techniques like Chain of Thought (CoT) prompting~\cite{wei2022chain} or fine-tuning (FLAN)~\cite{wei2022chain}.

These advancements have catalyzed an expansion of LLMs specifically designed for medical domains, such as PubMedGPT~\cite{bolton2022pubmedgpt}, BioGPT~\cite{luo2022biogpt}, Med-PaLM~\cite{singhal2023large} and its successor Med-PaLM 2~\cite{singhal2023towards}, Clinical Camel~\cite{toma2023clinical}, MedAlpaca~\cite{han2023medalpaca}, BioMistral~\cite{labrak2024biomistral}, LLMs for clinical trial recruitment ~\cite{wornow2024zero}, and others. Language models can handle omic information, as demonstrated by models such as HyenaDNA~\cite{nguyen2024hyenadna}, BioT5~\cite{pei2023biot5}, sc-GPT~\cite{cui2024scgpt}, and ProtLLM~\cite{zhuo2024protllm}.

\paragraph{Multimodal models in medicine}
Beyond language and text alone, multimodal models like Flamingo \cite{alayrac2022flamingo}, PaLI ~\cite{chen2022pali}, GPT-4~\cite{achiam2023gpt}, GPT-4v~\cite{openai2023gpt4v}, and LLaVa~\cite{liu2023improved,liu2024visual} have demonstrated remarkable capability in processing both text and images. Gemini~\cite{team2023gemini,team2024gemini} introduced further advancement in multimodal capabilities, exhibiting a distinct ability to reason across text, images, and other modalities such as video and audio.

Building upon these capable generic multimodal models, for medical applications specifically, recent works include vision-language models that span multiple medical imaging modalities as well as those that focus on a specific imaging domain, such as radiology or histopathology.  Efforts such as Med-Flamingo \cite{moor2023med}, BiomedCLIP~\cite{zhang2023biomedgpt}, Med-PaLM M~\cite{tu2024towards}, BiomedGPT~\cite{zhang2023biomedgpt}, Flamingo-CXR~\cite{tanno2024consensus}, LLaVa-Med~\cite{li2024llava}, PMC-VQA~\cite{zhang2023pmc}, RadFM~\cite{wu2023towards}, ELIXR~\cite{xu2023elixr}, XrayGPT~\cite{thawkar2023xraygpt}, MAIRA-1~\cite{hyland2023maira},  HeLM~\cite{belyaeva2023multimodal}, M-REGLE~\cite{zhou2024utilizing}, CONCH~\cite{lu2024visual}, PLIP~\cite{huang2023visual}, PathAsst~\cite{sun2024pathasst}, QuiltNet-B-32~\cite{ikezogwo2024quilt} and many others specifically explore the potential of multimodal models for medical applications, signaling a growing interest in this area.  These methods cover a range from generalist to specialist approaches. Models such as MAIRA-1~\cite{hyland2023maira}, XrayGPT~\cite{thawkar2023xraygpt}, Radiology-GPT \cite{liu2024radiologygpt}, and CT2Rep~\cite{hamamci2024ct2rep} focus on radiology report generation, and among modalities choose only chest X-ray or chest CT report generation. Some of these approaches broaden their capabilities to cover multiple types of modalities but focus on only one task, such as methods that aim for VQA capabilities like LLaVA-Med \cite{li2024llava}, and PMC-VQA~\cite{zhang2023pmc}, aiming to build assistants for medical question answering. 

While specialized VLMs demonstrate particular strengths, generalist models capable of handling a wide range of tasks and modalities, such as Med-PaLM M, are gaining prominence.  The field of medical AI is witnessing the emergence of comprehensive `Generalist Medical AI' models~\cite{moor2023med,moor2023foundation,tu2024towards,zhang2023biomedgpt} and the orchestration of AI tools for medical tasks using LLMs~\cite{ferber2024autonomous}. These models aspire to provide robust interaction with medical information in a manner similar to what general-purpose LLMs have done for broader domains. Pioneering efforts like these offer important initial insights into the potential for large multimodal models to provide assistance across various medical tasks using a unified platform.  This inconsistency underscores the urgent need for a unified benchmark to enable meaningful evaluation in this rapidly evolving field.

\paragraph{Multimodal evaluation benchmark and metrics}  Evaluation of medical VLMs suffers from a lack of consistency and standardization, creating a new landscape for works proposing new benchmarks to fill this gap. Multiple recent works demonstrate this inconsistency with varying tasks, datasets, and completely distinct sets of metrics, hindering direct comparison even for a same dataset. Along these lines, multiple recent works ~\cite{royer2024multimedeval,tu2024towards,wu2023towards,moor2023med,fleming2023medalign} suggest multimodal benchmarks such MultiMedEval, MultiMedBench, RadBench, RadMD, and MedMD to evaluate these generalist and multimodal models in a more systematic fashion.

\section{Discussion}
\label{sec:discussion}
In this study, we present three new models within the \ourmodel family, based upon \basemodel 1.5, across various medical modalities. We show promising performance across a number of tasks, including classification, VQA, and report generation. Our \ourmodel models are able to process complex medical data types, including 2D and 3D radiology images, histopathology patches, ophthalmology images, dermatology images, and genetic risk scores. Importantly, our models were fine-tuned using predominantly medical data and paired free text descriptive reports. These reports are ubiquitous in healthcare and our ability to use them as a training objective reduces the need for further expensive expert labelling.

The results in this study show early potential across a number of different tasks and individual modalities. We believe that the combination of tasks and multiple modalities in future work will enable AI models to address a far wider range of applications than has been previously possible. Longer context windows and improved reasoning abilities will enable decision-making that incorporates historical context, more closely reflecting how human specialists operate.

The opportunity for LMMs to analyze complex medical types including 3D radiology and large pathology images presents an exciting range of potential downstream applications. This work showcases our early explorations in CT, a three-dimensional modality that has been challenging to integrate with LMMs to date. This is due to a combination of vast data size, architectural limitations, and the jump in clinical task complexity of interpreting 3D imaging modalities (\vs 2D). While our results are currently a proof of concept, and do not yet reach performance required for clinical use, we expect architectures to rapidly improve. We look forward to exploring other similar complex modalities in future work.

While our findings in this study are promising and provide a glimpse into the potential of LMMs in medicine, it is important to thoroughly test them beyond traditional academic benchmarks. This is necessary to ensure they are safe and reliable before considering use in real-world situations, especially in safety critical areas like healthcare. In this work, we have tried to go deeper into the nuance of medical evaluation through the use of panels of specialists to assess and rate the performance of models on tasks such as report generation and question answering. We believe that an increasingly diverse range of healthcare professionals need to be deeply involved in future iterations of this technology, helping to guide the models towards capabilities that have valuable real world utility. There are a number of areas on which future evaluations should focus before models like these are considered safe and effective for clinical use:

\paragraph{Closing the gap between benchmark and bedside}
Despite the potential of machine learning in healthcare, there is growing concern about the reliability of algorithm validation methods. In medical image analysis, improvement on simple benchmark performance metrics may not translate to improved outcomes in clinical settings, leading to a disconnect between expectations and real-world usefulness. Benchmark datasets are an important step towards developing clinically useful models, but given their limitations in size, scope, and reflection of real world distributions, they are not themselves a proxy for real-world performance. The potential for generative AI lies foremost in assisting, rather than replacing human specialists in the diagnosis and management of disease; evaluations should shift from static benchmarks to realistic clinical scenarios that assess AI-human collaboration and its impact on patient outcomes.

\paragraph{Identifying and mitigating data bias and safety risks}
LLMs and LMMs trained on vast datasets risk inheriting biases and errors from their source data. This can lead to misdiagnoses and amplification of systemic bias. Before models like these are used in real world settings, careful evaluations that address safety and bias risks should be performed and any discovered risks should be mitigated~\cite{weng2024intentional}. End users should also carefully validate model performance for their specific use cases and patient populations.

\paragraph{Minimizing data contamination when evaluating zero-shot generalization in large models}
While LLMs exhibit impressive zero-shot generalization, it's important to note that their massive training datasets increase the potential for data contamination, which may result in overestimation of their true generalization abilities. Large models like \basemodel might have inadvertently ``seen'' examples related to the task during training, even if those examples were not explicitly labeled. This hidden exposure compromises our understanding of models' true ability to generalize to completely novel concepts when evaluating on open datasets. Researchers are actively investigating the impact of data contamination to ensure we accurately gauge capabilities of such large models~\cite{vogel2022vl,udandarao2024no}. Prospective studies, while typically more expensive and time-consuming to execute than retrospective studies, are another option for mitigating this risk.

\section{Conclusion}
\label{sec:conclusion}

Multimodal generative AI, exemplified by powerful models like \basemodel, holds great potential to revolutionize healthcare. While medicine is a rapidly growing use case for these new models, general purpose models may not naturally perform well in the medical domain due to its highly specialized data.

To explore the potential for models like \basemodel in medicine, we developed several models within the new \ourmodel family, a series of models built upon the multimodal foundation of \basemodel and fine-tuned on a diverse range of medical data including radiology, histopathology, ophthalmology, dermatology and genomics. We assessed our \ourmodel models' performance using a comprehensive medical benchmarking suite, including both established benchmarks and custom benchmarks designed to reflect clinical relevance. Notably, some benchmarks involved evaluations by medical experts for tasks such as generating CXR and CT reports and radiology VQA.

\ourmodeltwod sets a new standard for expert-evaluated chest X-ray report generation, outperforming previous models, and \ourmodelthreed showcases the first LMM-based report generation for 3D CT. Beyond report generation, \ourmodeltwod demonstrates exceptional performance in VQA and classification across various medical imaging modalities. Beyond imaging, \ourmodelpolygenic outperforms conventional polygenic risk score methods in predicting disease risk. These results demonstrate the potential of the \basemodel foundation and the fine-tuned \ourmodel family in the medical domain. Nonetheless, the results also underscore the need for further rigorous research to ensure safe and effective implementation in real-world clinical settings.

While advanced capabilities on individual medical tasks are useful in their own right, we envision a future in which all of these capabilities are integrated together into comprehensive systems to perform a range of complex multidisciplinary clinical tasks, working alongside humans to maximize clinical efficacy and improve patient outcomes. The results presented in this study represent a step towards realizing this vision.

\newpage
\section{Contributions and Acknowledgments}
\subsubsection*{Contributions}
Authors are listed here associated with their primary workstreams. Many authors contributed to additional workstreams beyond the one under which they are listed.

\begin{multicols*}{2}
\setlength{\columnsep}{5pt}
\small

\vspace{-1.0\baselineskip}
\begin{itemize}[leftmargin=1em,rightmargin=0em]
\setlength\itemsep{0pt}
    \item[] \textbf{Technical Leads} \\\vspace{-6pt}
    \item[] Lin Yang\textsuperscript{$*$},\textsuperscript{1}
    \item[] Shawn Xu\textsuperscript{$*$},\textsuperscript{1}
    \item[] Andrew Sellergren\textsuperscript{*},\textsuperscript{1} \\\vspace{-6pt}
    \item[] \footnotesize{$*$ Equal contribution}
\end{itemize}

\vspace{-1.0\baselineskip}
\begin{itemize}[leftmargin=1em,rightmargin=0em]
\setlength\itemsep{0pt}
    \item[] \textbf{Chest X-Ray} \\\vspace{-6pt}
    \item[] Timo Kohlberger\textsuperscript{1}
    \item[] Ira Ktena\textsuperscript{2}
    \item[] Kendall Park\textsuperscript{1}
    \item[] Ryutaro Tanno\textsuperscript{2}
    \item[] David G. T. Barrett\textsuperscript{2}
    \item[] Wei-Hung Weng\textsuperscript{1}
    \item[] Khaled Saab\textsuperscript{1}
    \item[] Tao Tu\textsuperscript{1}
\end{itemize}

\vspace{-1.0\baselineskip}
\begin{itemize}[leftmargin=1em,rightmargin=0em]
\setlength\itemsep{0pt}
    \item[] \textbf{Computed Tomography} \\\vspace{-6pt}
    \item[] Atilla Kiraly\textsuperscript{1}
    \item[] Akshay Goel\textsuperscript{1}
    \item[] Arnav Agharwal\textsuperscript{1}
    \item[] Nick George\textsuperscript{1}
\end{itemize}

\vspace{-1.0\baselineskip}
\begin{itemize}[leftmargin=1em,rightmargin=0em]
\setlength\itemsep{0pt}
    \item[] \textbf{Genomics} \\\vspace{-6pt}
    \item[] Cory Y. McLean\textsuperscript{1}
    \item[] Yuchen Zhou\textsuperscript{1}
    \item[] Farhad Hormozdiari\textsuperscript{1}
    \item[] Eric Wang\textsuperscript{2}
\end{itemize}

\vspace{-1.0\baselineskip}
\begin{itemize}[leftmargin=1em,rightmargin=0em]
\setlength\itemsep{0pt}
    \item[] \textbf{Pathology} \\\vspace{-6pt}
    \item[] Dave Steiner\textsuperscript{1}
    \item[] Faruk Ahmed\textsuperscript{1}
    \item[] Tiam Jaroensri\textsuperscript{1}
    \item[] Ellery Wulczyn\textsuperscript{1}
\end{itemize}

\vspace{-1.0\baselineskip}
\begin{itemize}[leftmargin=1em,rightmargin=0em]
\setlength\itemsep{0pt}
    \item[] \textbf{Additional Contributions} \\\vspace{-6pt}
    \item[] Fayaz Jamil\textsuperscript{1}
    \item[] Theo Guidroz\textsuperscript{1}
    \item[] Yang Wang\textsuperscript{3}
    \item[] Siyuan Qiao\textsuperscript{2}
    \item[] Yun Liu\textsuperscript{1}
    \item[] S. Sara Mahdavi\textsuperscript{2}
\end{itemize}

\vspace{-1.0\baselineskip}
\begin{itemize}[leftmargin=1em,rightmargin=0em]
\setlength\itemsep{0pt}
    \item[] \textbf{Clinical} \\\vspace{-6pt}
    \item[] Charles Lau\textsuperscript{4}
    \item[] Sreenivasa Raju Kalidindi\textsuperscript{5}
    \item[] Mozziyar Etemadi\textsuperscript{6}
    \item[] Jorge Cuadros\textsuperscript{7}
    \item[] Gregory Sorensen\textsuperscript{8}
\end{itemize}

\columnbreak

\vspace{-1.0\baselineskip}
\begin{itemize}[leftmargin=1em,rightmargin=0em]
\setlength\itemsep{0pt}
    \item[] \textbf{Google Research and Google DeepMind\\ Leadership} \\\vspace{-6pt}
    \item[] Shruthi Prabhakara\textsuperscript{1}
    \item[] Daniel Tse\textsuperscript{1}
    \item[] Shravya Shetty\textsuperscript{1}
    \item[] Greg Corrado\textsuperscript{1}
    \item[] Katherine Chou\textsuperscript{1}
    \item[] Yossi Matias\textsuperscript{1}
    \item[] S. M. Ali Eslami\textsuperscript{2}
    \item[] David Fleet\textsuperscript{2}
    \item[] Joelle Barral\textsuperscript{2}
\end{itemize}

\vspace{-1.0\baselineskip}
\begin{itemize}[leftmargin=1em,rightmargin=0em]
\setlength\itemsep{0pt}
    \item[] \textbf{Research Leads} \\\vspace{-6pt}
    \item[] Daniel Golden\textsuperscript{$\dagger$},\textsuperscript{1}
    \item[] Shekoofeh Azizi\textsuperscript{$\dagger$},\textsuperscript{2}
    \item[] Rory Pilgrim\textsuperscript{1}
    \item[] Christopher Kelly\textsuperscript{1} \\\vspace{-6pt}
    \item[] \footnotesize{$\dagger$ Equal contribution} 
\end{itemize}

\footnotesize
\noindent\rule{4cm}{0.4pt}
\begin{itemize}[leftmargin=1em,rightmargin=0em]
\setlength\itemsep{2pt}
    \item[] \textsuperscript{1}Google Research
    \item[] \textsuperscript{2}Google DeepMind
    \item[] \textsuperscript{3}Verily Life Sciences
    \item[] \textsuperscript{4}Google Research via Advanced Clinical
    \item[] \textsuperscript{5}Apollo Radiology International
    \item[] \textsuperscript{6}Northwestern Medicine
    \item[] \textsuperscript{7}EyePACS, Inc and Meredith Morgan University\\ Eye Center, University of California at Berkeley
    \item[] \textsuperscript{8}DeepHealth / RadNet
\end{itemize}

\end{multicols*}

\subsubsection*{Acknowledgements}
This project was an extensive collaboration between many teams at Google Research and Google DeepMind. We thank Kevin Swersky and Mike Schaekermannn for their feedback and insight, which significantly contributed to the enhancement of this report. We also thank Sami Lachgar, Lauren Winer, Maggie Shiels, Jessica Valdez, Jon Small, Aaron Abood, Rishad Patel, Christian Wright, Annisah Um’rani, Jean-baptiste Alayrac, Aishwarya Kamath, Viorica Patraucean, Rory Sayres, Abbi Ward, Louis Blankemeier, Olga Kanzheleva, Taedong Yun, Ksenia Konyushkova, Christos Kaplanis, Juanma Zambrano Chaves, Alan Karthikesalingam, Vivek Natarajan, and Can Kirmizi for their valuable insights, technical support and feedback during our research. We thank Kimberly Kanada and Ilana Traynis for their review of the qualitative examples shown in this manuscript. We are grateful to Jonathon Shlens, Dale Webster and Oriol Vinyals for their support during the course of this project. We also thank Michael Colligan and Brittany Stein from DeepHealth/RadNet for their support with data curation. 

This research was conducted using the UK Biobank Resource under application number 65275. The results shown here are in part based upon data generated by the \href{https://www.cancer.gov/tcga}{TCGA Research Network}. 
The authors thank the National Cancer Institute for access to
NCI's data collected by the National Lung Screening Trial (NLST). The statements contained herein are solely those of the authors and do not represent or imply concurrence or endorsement by NCI.

\subsubsection*{Data Availability}
Except IND1, CXR-US2, and CT-US1, Eyepacs, and TTH, which are private datasets, the rest of the datasets utilized for developing, benchmarking, and evaluation of \basemodel and \ourmodel in this report are publicly accessible with appropriate permissions.  We intend to publicly release our updated classification labels and custom VQA question and answer pairs for the MIMIC-CXR dataset, our splits for the PAD-UFES-20 and VQA-Rad datasets, and several suggested replacement question and answer pairs for the VQA-Rad dataset which were recommended by our reading radiologist. This text will be updated when that data is available.

\subsubsection*{Code Availability}
We will not open-source the model code and weights because of the safety concerns associated with unmonitored use in medical settings. To ensure responsible innovation, we will collaborate with our research partners and healthcare providers to validate and explore safe applications of the Gemini and \ourmodel through Google Cloud APIs.

\subsubsection*{Competing Interests}
This study was funded by Alphabet Inc and/or a subsidiary thereof (‘Alphabet’). Authors who are affiliated with Google Research, Google DeepMind, and Verily Life Sciences are employees of Alphabet and may own stock as part of the standard compensation package. 

\subsubsection*{Use of AI in Manuscript Preparation}
This manuscript was written manually, with a small number of copy edits performed using Gemini. The authors take all responsibility for the contents. 

\clearpage
\bibliography{main}

\newpage
\appendix
\label{sec:appendix}

\renewcommand{\thesection}{A.\arabic{section}}
\renewcommand{\thefigure}{A.\arabic{figure}}
\renewcommand{\thetable}{A.\arabic{table}} 
\renewcommand{\theequation}{A.\arabic{equation}} 
\renewcommand{\theHsection}{A\arabic{section}}

\setcounter{section}{0}
\setcounter{figure}{0}
\setcounter{table}{0}
\setcounter{equation}{0}

\noindent \textbf{\LARGE{Appendix}}\\
\normalfont

\section{Additional data details}

\subsection{Revised MIMIC-CXR classification labels}\label{sub:mimic_labels}

One of the limitation of the MIMIC-CXR dataset is the lack of ground-truth labels. MIMIC-CXR~JPG \cite{MIMIC-CXR-JPG2019} extracted structured labels from 277,827 radiology reports using CheXpert~\cite{irvin2019chexpert}, a natural language processing (NLP) tool to extract observations from radiology reports. In order to improve upon these labels on the test subset, we utilized Med-PaLM 2~\cite{singhal2023towards} coupled with US-based board certified radiologists to refine those labels. This work is further adjudication of the labels used in~\citet{xu2023elixr}. We first used a keyword search to identify reports containing text associated with the a given finding (\eg ``Cardiomegaly''). Next, Med-PaLM 2 was applied to the flagged radiology reports on a per-label basis using two queries shown in Table~\ref{tab:medplam2queries} for a total of 23,824 queries. All identified positive and negative labels that disagreed with the original labels were flagged for human verification.

Three US-based board certified radiologists reviewed the 1,378 flagged labels and a fourth academic US-based board certified thoracic radiologist adjudicated the responses of reviewer disagreements. For each finding and report, radiologists selected one of four possible labels defined by MIMIC-CXR~JPG: positive, negative, uncertain, and not mentioned. Zero-round adjudication was performed on the reviewers’ annotations. There was strong inter-rater agreement (Fleiss’ $\kappa$ = 0.71); reviewers were unanimous for 77\% of the labels. In cases of disagreement between reviewers, majority vote was used (21\%), and when all three reviewers disagreed (2\%) a senior academic thoracic radiologist provided the final determination.

In the final analysis of the flagged reports and findings, Med-PaLM 2's label matched the ground truth 66\% of the time while the original labels were correct in 19\% of the cases. The labels are in preparation to be released~\cite{mimic-cxr-gt}.

\begin{table}[h]
\centering
\small
\caption{\small{\textbf{Prompt styles used for Med-Palm2 in extracting structured labels.} Two prompt styles were used to extract labels from the given MIMIC-CXR radiology reports. The following shows the template used for identifying fractures.}}
\label{tab:medplam2queries}
\renewcommand{\arraystretch}{1.3}
\begin{tabular}{l|l}
\Xhline{2.5\arrayrulewidth}
\textbf{Prompt Style} & \textbf{Prompt Example} \\
\Xhline{2\arrayrulewidth}
Bot     & \begin{tabular}[c]{@{}l@{}} {\small [Bot] I'm a helpful radiology assistant, who provides concise answers} \\
{\small to questions about information in a chest x-ray report.} \\
{\small [User] Determine the answer to the following question: } \\
{\small [Does the patient have a fracture?], }\\
{\small given the context of the follow chest x-ray report: <REPORT TEXT>}\\
{\small Do not mention conditions or parts of the report not relevant to the question.}\\
{\small Make sure to only answer: [Does the patient have a fracture?]}\\
{\small [Bot]}\end{tabular}
\\ \hline                                                                                                               
Question          & \begin{tabular}[c]{@{}l@{}}{\small You are a helpful medical knowledge assistant.} \\ 
{\small Provide useful, complete, concise, and scientifically-grounded} \\ 
{\small queries to radiology reports.}\\ \\  
{\small Does this report mention that the patient has a fracture?} \\ 
Report:\textless{}REPORT TEXT\textgreater{}\end{tabular} \\
\Xhline{2.5\arrayrulewidth}
\end{tabular}
\end{table}

\subsection{Prompts for VQA and CXR classification evaluations}\label{sub:vqa_and_class_eval_prompts}

We explored both binary question prompting for each of the evaluated top 5 conditions for MIMIC-CXR including the abnormal/normal class, as well as multi-select prompts for all of them at once on the validation set. Since binary prompts overall yielded better macro F1 scores for \ourmodel, these were used at evaluation. For Gemini Ultra binary question were used for the top-5 conditions and a multi-select prompt for the normal/abnormal condition, see Tables~\ref{tab:cxr_classification_eval_prompts} and~\ref{tab:cxr_classifcation_question_arguments}. Each prompt template was crafted for each model in aiming to optimize its performance, which yielded very short prompt templates for both classification and VQA for \ourmodel, since it was fine-tuned with clinical questions.

\begin{table}[h]
\centering
\small
\caption{\small{\textbf{Zero-shot prompt templates used for VQA evaluations (excluding formatting tokens).} The following prompt templates were used for each model and VQA evaluation data set. The <~question~> placeholder was replaced with the individual question of each VQA triplet.}}
\renewcommand{\arraystretch}{1.5}
\label{tab:vqa_eval_prompts}
\begin{tabular}{l|c|p{11cm}}
\Xhline{2.5\arrayrulewidth}
\textbf{VQA Dataset} & \textbf{Model} & \textbf{Prompt Template} \\
\Xhline{2\arrayrulewidth}
\multirow{5}{*}{VQA-Rad} & \multirow{4}{*}{Gemini} & \small < image > You are a helpful radiology assistant.
  Given this radiology image, which can be frontal chest X-ray provide a very short, concise answer, like: ``pleural effusion,'' ``yes,'' ``right,'' to the following question: < question >\\ \cline{2-3}
  & \ourmodel    & \small < image >< question > \\ \hline
  \multirow{5}{*}{Slake-VQA} & \multirow{4}{*}{Gemini}  & \small < image > You are a helpful radiology assistant.
Given this radiology image, which can be a frontal chest X-ray, a single slice head or abdominal CT or MR image, provide a very short, concise answer, like: ``pleural effusion,'' ``yes,'' ``right,'' to the following question: < question >\\ \cline{2-3}
  & \ourmodel    & \small < image >< question > \\ \hline
\multirow{4}{*}{PathVQA} & \multirow{3}{*}{Gemini} & \small < image > You are a helpful pathology assistant.
Given the pathology image, provide a concise answer, like: ``pleural effusion,'' ``yes,'' ``right,'' to the following question: < question >\\ \cline{2-3}
  & \ourmodel    & \small < image >< question > \\
 \Xhline{2.5\arrayrulewidth}
\end{tabular}
\end{table}

\begin{table}[h]
\centering
\small
\caption{\small{\textbf{Zero-shot prompts used evaluating classification performance on chest X-ray.}} 
Binary questions were employed for \ourmodel on all CXR classification conditions, and for Gemini Ultra on all except for the normal/abnormal condition (label ``No Finding''). The <~question~> placeholder was replaced with the label-dependent text listed in Table~\ref{tab:cxr_classifcation_question_arguments}, while <~view position~> with the respective MIMIC-CXR meta information (e.g. ``AP'').}
\label{tab:cxr_classification_eval_prompts}
\renewcommand{\arraystretch}{1.5}
\begin{tabular}{l|c|l}
\Xhline{2.5\arrayrulewidth}
\textbf{Model} & \textbf{Used for} & \textbf{Prompt template} \\
\Xhline{2\arrayrulewidth}\small
\multirow{6}{*}{Gemini} & \multirow{6}{*}{Top-5 Conditions} & \begin{tabular}[t]{@{}p{10.5cm}@{}}\small < image > You are a helpful radiology assistant.\\ The following is a question about findings in chest X-ray in the frontal view. Answer only with Yes or No.\\
Q: Given the X-ray < view position > image, < question > indicated in this chest X-ray image?\\
A: \end{tabular} \\ \midrule
\multirow{1}{*}{\ourmodel} & \begin{tabular}[c]{@{}l@{}}{\makecell{Top-5 Conditions\\and\\Normal/Abnormal}}\end{tabular} & \multirow{1}{*}{< image >< question >} \\   \midrule
\multirow{10}{*}{Gemini} & \multirow{10}{*}{\begin{tabular}[c]{@{}l@{}}{\makecell{Normal/Abnormal}}\end{tabular}} & \begin{tabular}[t]{@{}p{10.5cm}@{}}\small < image > You are a helpful radiology assistant.\\
The following are multiple-choice questions about findings in chest X-ray in the frontal view. Identify if a specific type of abnormality is shown in the X-ray by responding with the corresponding answer choice letter(s).\\
Q: Given the X-ray image, which of the following abnormalities are indicated by the image?\\
(A) Atelectasis (B) Cardiomegaly (C) Consolidation (D) Edema (I) Pleural Effusion (O) No Abnormality (E) Enlarged Cardiomediastinum (F) Fracture (G) Lung Lesion (H) Lung Opacity (I) Pleural Effusion (J) Pleural Other (K) Pneumonia (L) Pneumothorax (M) Support Devices \end{tabular} \\ 
\Xhline{2.5\arrayrulewidth}
\end{tabular}
\end{table}

\begin{table}[h]
\centering
\small
\caption{\small{\textbf{Question arguments used for binary classification prompt templates.} We used the following arguments for the <~question~> placeholder in the binary classification prompts, Table~\ref{tab:cxr_classification_eval_prompts}, which were triggered depending on the corresponding CheXpert label to be in $\{0.0, 1.0\}$}.}
\renewcommand{\arraystretch}{1.5}
\label{tab:cxr_classifcation_question_arguments}
\begin{tabular}{l|l|p{7cm}}
\Xhline{2.5\arrayrulewidth}
\textbf{Condition} & \textbf{Argument for \ourmodel prompt} & \textbf{Argument for Gemini prompt} \\
\Xhline{2\arrayrulewidth}
Atelectasis     & Is there atelectasis?     & is atelectasis \\
Cardiomegaly    & Is there cardiomegaly?    & is cardiomegaly \\
Consolidation   & Is there consolidation?   & is consolidation \\
Edema           & Is there pulmonary edema? & is pulmonary edema \\
Pleural Effusion & Is there pleural effusion? & is pleural effusion  \\

\Xhline{2.5\arrayrulewidth}
\end{tabular}
\end{table}

\subsection{New balanced splits for VQA-Rad dataset}
\label{sub:new_rad_vqa_splits}

The official train/test split of the VQA-Rad~\cite{lau2018dataset} dataset comprises 1,797 QA pairs for training (i.e. dataset field QID\_para $\in$ \{'freeform', 'para'\}) for 313 different MedPIX\textregistered  images (per field IMAGEID), and 451 QA pairs for 203 different images in the test set (i.e. field QID\_para $\in$ \{'test\_freeform', 'test\_para'\}). Although images were sampled such that each is not only for a different case, but also a different patient, see~\cite{lau2018dataset}, 202 of the test IMAGEIDs and also match the train set IMAGEIDs. Hence most of the test images also appear in the train set, only the questions and answers differ. For some VQAs even the latter is not completely true, since VQA-Rad contains paraphrased questions which share the same answer.

To remove this train/test contamination issue, in~\citet{xu2023elixr} we proposed a different validation/test split, which is based on the IMAGEIDs in order to ensure disjoint images and thus patients. In this work we split this relatively large test set further into a new test and train set, which are roughly equal-sized, and assign a few remaining former test set IMAGEIDs and corresponding VQAs to the existing validation set, such that all three new splits not only are roughly equal-sized, but their ratio of open to closed questions (as determined by field A\_TYPE) are approximately equal within each of the three depicted anatomical regions (chest, head and abdomen), see Table~\ref{tab:vqarad-open-closed-ratio}. We chose to equalize this ratio since open-ended questions are more difficult for AI models, and the level of difficulty ought to be similar for each new split. Similarly, while swapping individual IMAGEIDs and associated question and answers, we approximately equalized the distribution of question types (field Q\_TYPE) within each split, in order to gain similar ones, see Table~\ref{tab:vqarad-question-type-distrib}. 

\begin{table}[h]
\centering
\small
\caption{\small{\textbf{Distribution of VQA-Rad answer types (closed- vs. open-ended) across the new balanced splits for each anatomical region.} The new train/validation/test splits not only guarantee the images and thus patients to be disjoint, but also provide a similar ratio of open to close-ended questions (dataset field A\_TYPE) per anatomical region, which typically affects VQA performance of AI models due to open-ended questions being more difficult to answer correctly.}}
\label{tab:vqarad-open-closed-ratio}
\renewcommand{\arraystretch}{1.2}
\begin{tabular}{c|c|c|c|c|c}
\Xhline{2.5\arrayrulewidth}
\multirow{2}{*}{\textbf{Anatomical Region}} &\multirow{2}{*}{\textbf{Split (balanced)}} & \multicolumn{3}{c|}{\textbf{Number of question-answers pairs}} & \textbf{Ratio of open to closed} \\ \cline{3-5}
\textbf{}            & \textbf{}                 & \textbf{Open}            & \textbf{Closed}  & \textbf{Total}  & \textbf{}                        \\ \hline
\multirow{3}{*}{Abdomen} 

                     & Train                     & 104                      & 153              & 257             & 68.0\%                           \\
                     & Validation                      & 97                       & 143              & 240             & 67.8\%                           \\ 
                     & Test                      & 99                       & 143              & 242             & 69.2\%                           \\ \hline
\multirow{3}{*}{Chest} 
                     & Train                     & 94                       & 185              & 279             & 50.8\%                           \\
                     & Validation                      & 107                      & 161              & 268             & 66.5\%                           \\ 
                     & Test                      & 82                       & 165              & 247             & 49.7\%                           \\\hline
\multirow{3}{*}{Head} 
                     & Train                     & 131                      & 118              & 249             & 111.0\%                          \\
                     & Validation                      & 113                      & 107              & 220             & 105.6\%                          \\   
                     & Test                      & 122                      & 144              & 246             & 84.7\%                           \\ \Xhline{2\arrayrulewidth}
\multicolumn{2}{c|}{All}                         & 949                      & 1,299            & 2,248           & 73.1\%  \\ \Xhline{2.5\arrayrulewidth}                         
\end{tabular}
\end{table}

\begin{table}[h]
\centering
\small
\caption{\small{\textbf{Distribution of VQA-Rad question types across the new balanced splits.} In addition to comparable open-to-closed-question ratios, the question types (dataset field Q\_TYPE) distribute similarly as well.}}
\label{tab:vqarad-question-type-distrib}
\renewcommand{\arraystretch}{1.2}
\begin{tabular}{l|ccc|ccc|ccc|c}
\Xhline{2.5\arrayrulewidth}
\multicolumn{1}{l|}{Anatomical Region} & \multicolumn{3}{c|}{Abdomen}             & \multicolumn{3}{c|}{Chest}                                    & \multicolumn{3}{c|}{Head}                & \multirow{2}{*}{All} \\ \cline{1-10}
\multicolumn{1}{l|}{Split (balanced)}  & Train & Val. & \multicolumn{1}{c|}{Test} & Train & \multicolumn{1}{c}{Val.} & \multicolumn{1}{c|}{Test} & Train & Val. & \multicolumn{1}{c|}{Test} & 
\\ \Xhline{2\arrayrulewidth}
Total             & 257      & 240     & 242    & 279     & 268    & 247    & 249     & 220    & 246   & 2248                 \\ \hline
Presence          & 105      & 95      & 100    & 101     & 92     & 93     & 79      & 67     & 68    & 800                  \\
Positional        & 26       & 27      & 21     & 37      & 36     & 42     & 35      & 42     & 50    & 316                  \\
Abnormality       & 19       & 18      & 27     & 15      & 40     & 21     & 22      & 25     & 15    & 202                  \\
Other             & 20       & 20      & 22     & 26      & 19     & 19     & 22      & 14     & 32    & 194                  \\
Modality          & 25       & 13      & 23     & 19      & 17     & 13     & 35      & 24     & 16    & 185                  \\
Size              & 17       & 20      & 11     & 41      & 26     & 27     & 10      & 5      & 14    & 171                  \\
Plane             & 12       & 9       & 13     & 14      & 12     & 13     & 17      & 16     & 14    & 120                  \\
Attribute Other   & 13       & 19      & 7      & 5       & 9      & 8      & 4       & 12     & 10    & 87                   \\
Organ             & 5        & 4       & 2      & 10      & 8      & 7      & 13      & 3      & 7     & 59                   \\
Color             & 6        & 7       & 8      & 2       & 0      & 0      & 9       & 7      & 13    & 52                   \\
Counting          & 3        & 8       & 3      & 5       & 1      & 2      & 0       & 1      & 1     & 24                   \\ \Xhline{2.5\arrayrulewidth}
   
\end{tabular}
\end{table}

\begin{table}[t]
\centering
\footnotesize
\caption{\small{Examples of some of our curated captions for histopathology patches used in our training set.}}
\label{tab:histopathology-generated-captions}

\begin{tabular}{m{3.4cm}|l}
\Xhline{2.5\arrayrulewidth}
\textbf{Task} & \textbf{Curated Captions} \\ \Xhline{2\arrayrulewidth}
Breast cancer detection in lymph nodes & \begin{tabular}{p{12cm}l@{}}
Microscopic view of a lymph node with infiltrating malignant epithelial cells consistent with metastatic breast carcinoma. \\ \\
Lymph node section on histopathology demonstrating features of metastatic carcinoma, likely originating from the breast. \\ \\
Region of a lymph node on histopathology, showing predominantly lymphocytes and immune cells.
\end{tabular} \\ \hline
Histologic subtyping for lung adenocarcinoma & \begin{tabular}{p{12cm}l@{}}
H\&E histopathology image demonstrating acinar subtype lung adenocarcinoma with glandular formation. \\ \\
Cribriform pattern in lung adenocarcinoma with closely packed, back-to-back glands lacking an organized pattern. \\ \\
Microscopic view of lung adenocarcinoma showing tumor cells spreading along the preexisting alveolar architecture, consistent with the lepidic subtype.
\end{tabular} \\ \hline
Prostate Gleason grading & \begin{tabular}{p{12cm}l@{}}
Microscopic view of a prostate biopsy with discrete, uniformly sized and shaped glands consistent with Gleason pattern 3 prostate cancer. \\ \\
Prostate biopsy image highlighting areas of fused glands or poorly formed glands, consistent with Gleason pattern 4 carcinoma. \\ \\
Microscopic view of a prostate tissue with discrete, uniformly sized and shaped glands consistent with Gleason pattern 3 prostate cancer. \\
\end{tabular} \\ \hline

Breast cancer nuclear pleomorphism grading & \begin{tabular}{p{12cm}l@{}}
Microscopic view of invasive breast carcinoma showing moderate nuclear atypia, including enlarged nuclei and prominent nucleoli (nucleopleomorphism score 2). \\ \\
Focus on infiltrating tumor cells within an H\&E stained image, demonstrating bland nuclear features suggestive of low-grade invasive breast carcinoma. \\ \\
H\&E histopathology image demonstrating invasive breast carcinoma with high-grade nuclear features (nucleopleomorphism score 3).
\end{tabular} \\ \hline
Breast cancer tubule formation grading & \begin{tabular}{p{12cm}l@{}}
This microscopic view demonstrates a predominance of well-defined glandular structures, indicating a tubule formation score of 1 in this breast carcinoma. \\ \\
The presence of some discernible tubules, alongside regions of less-defined glandular architecture, indicates a tubule formation score of 2. \\ \\
A tubule formation score of 3 is evident within this H\&E stained image, where the invasive carcinoma shows a scarcity of well-defined glandular structures.
\end{tabular} \\ \hline
Cervical dysplasia grading & \begin{tabular}{p{12cm}l@{}}
H\&E stained image of a cervical biopsy demonstrating features of cervical intrapethileal neoplasia grade 1 (CIN 1), including nuclear atypia in the lower third of the epithelium. \\ \\
Microscopic view of a cervical biopsy with CIN 1, showing enlarged nuclei and increased nuclear-to-cytoplasmic ratio in the basal layer of the epithelium. \\ \\
H\&E histopathology of a cervical biopsy focusing on high-grade dysplasia, showing prominent nuclear abnormalities and disruption of the normal epithelial architecture.
\end{tabular} \\

\Xhline{2.5\arrayrulewidth}
\end{tabular}
\end{table}

\begin{table}[t]
\centering
\footnotesize
\caption{\small{Overview of patch-level histopathology datasets used for fine-tuning and linear-probe evaluation. Tasks adopted from~\citep{lai2023domain}. The OOD column indicates whether the task was included in \ourmodel fine-tuning. Number of slides shows the counts split across train, validation, and test sets, respectively.}}
\label{tab:histopathology-tasks}
\renewcommand{\arraystretch}{1.25}
\begin{tabular}{l|l|c|l|c}
\Xhline{2.5\arrayrulewidth}
\textbf{Dataset} & \textbf{Task} & \begin{tabular}[c]{@{}l@{}}\textbf{Number of slides} \end{tabular}  & \textbf{Classes} & \textbf{OOD}  \\ \Xhline{2\arrayrulewidth}
\begin{tabular}[c]{@{}l@{}}CAMELYON16\\\cite{bejnordi2017diagnostic}\end{tabular}               & \begin{tabular}[c]{@{}l@{}}Breast cancer detection\\ in lymph nodes\end{tabular}            & 216/54/258                                    & Tumor, Non-Tumor                  & -       \\ \hline
\begin{tabular}[c]{@{}l@{}}Lung AD\\\cite{sadhwani2021comparative}\end{tabular}                 & 
\begin{tabular}[c]{@{}l@{}}Histologic subtyping \\ for lung adenocarcinoma\end{tabular}     & 73/25/50                                      & 9 classes$^*$         & -      \\ \hline
\begin{tabular}[c]{@{}l@{}}Gleason NCB\\\cite{nagpal2020development}\end{tabular}                       & \begin{tabular}[c]{@{}l@{}}Gleason grading on prostate \\ needle core biopsies\end{tabular} & 178/85/88                             & Benign, GP3, GP4, GP5             & -        \\ \hline
\begin{tabular}[c]{@{}l@{}}Gleason RP\\\cite{nagpal2019development}\end{tabular}                        & \begin{tabular}[c]{@{}l@{}}Gleason grading on radical \\ prostatectomies\end{tabular}       & 550/259/202                           & Benign, GP3, GP4, GP5            & -        \\ \hline
\begin{tabular}[c]{@{}l@{}}Breast IC\\\cite{jaroensri2022deep}\end{tabular}                     & \begin{tabular}[c]{@{}l@{}}Breast invasive \\ carcinoma detection\end{tabular}              & 573/288/669                                   & \begin{tabular}[c]{@{}l@{}}Benign, DCIS, \\ Invasive Carcinoma\end{tabular}   & -       \\ \hline
\begin{tabular}[c]{@{}l@{}}Breast NP\\\cite{jaroensri2022deep}\end{tabular}                     & \begin{tabular}[c]{@{}l@{}}Breast cancer nuclear \\ pleomorphism grading\end{tabular}       & 681/343/945                                   & NP1, NP2, NP3                     & -        \\ \hline
\begin{tabular}[c]{@{}l@{}}Breast TF\\\cite{jaroensri2022deep}\end{tabular}                     & \begin{tabular}[c]{@{}l@{}}Breast cancer tubule \\ formation grading\end{tabular}           & 681/343/945                                   & TF1, TF2, TF3                     & -       \\ \hline
\begin{tabular}[c]{@{}l@{}}CIN\\\cite{lai2023domain}\end{tabular}                               & \begin{tabular}[c]{@{}l@{}}Cervical dysplasia \\ grading\end{tabular}                       & 329/74/229                                    & Non-tumor, CIN 1, CIN 2+          & -      \\ \hline
\begin{tabular}[c]{@{}l@{}}CRC\\\cite{wulczyn2021interpretable} \end{tabular}                   & \begin{tabular}[c]{@{}l@{}}Colorectal carcinoma \\ detection\end{tabular}                   & 149/51/44                                     & Tumor, Non-Tumor                  & -       \\ \hline
\begin{tabular}[c]{@{}l@{}}Tissue type\\\cite{weng2019multimodal}\end{tabular}                       & \begin{tabular}[c]{@{}l@{}}Tissue type classification \\ (internal dataset)\end{tabular}    & 17319/6488/6719                          & 16 tissue types$^\dagger$               & \tick   \\ \hline  
\begin{tabular}[c]{@{}l@{}}TCGA study\\\cite{lai2023domain}\end{tabular}                   & \begin{tabular}[c]{@{}l@{}}TCGA study type \\ classification\end{tabular}                   & 2952/1466/1489                                & 10 TCGA study types$^\ddagger$           & \tick   \\
    
\Xhline{2.5\arrayrulewidth}
\end{tabular}

{\raggedright
\vspace{0.03in}
\scriptsize{
$^*$ Lung AD histologic subtypes and other classes: Acinar, Cribriform, Lepidic, Micropapillary, Papillary, Solid, Leukocyte, Necrosis, Non-tumor. \\
$^\dagger$ Tissue types: Appendix, Breast, Cervix, Colon and rectum, Fallopian Tube, Gallbladder, Liver, Lymph node, Ovary, Placenta, Prostate, Skin, Thyroid, Upper GI, Uterus, Vas deferens. \\ 
$^\ddagger$ TCGA study types: BLCA, BRCA, COAD, HNSC, KIRC, LIHC, LUAD, LUSC, OV, STAD. \\} 
}
\end{table}

\subsection{Polygenic risk prediction}

We crafted prompts for predicting the status of various health outcomes using both an individual's PRS image and their demographic information. An example prompt for predicting coronary artery disease is shown in Table \ref{tab-app:genomics_prompt_example}.

For linear probes of out-of-distribution outcomes, we used data of a related in-distribution outcome to train the linear probe and then evaluated the predictions on the out-of-distribution outcome. For example, in order to evaluate diabetic retinopathy, we train the linear probe to predict type 2 diabetes and evaluate the type 2 diabetes predictions on diabetic retinopathy data. In general, the most related in-distribution outcome is defined as the outcome with the highest Matthew's correlation coefficient with the out-of-distribution outcome across individuals in our training set (Table \ref{tab-app:genomics_ood_correlations}). To evaluate \ourmodelpolygenic, we directly prompted \ourmodelpolygenic to predict the out-of-distribution outcome without providing any information about correlations between out-of-distribution and in-distribution outcomes.

\begin{table}[]
    \centering
    \caption{\small{Examples of a prediction prompt for coronary artery disease using an individual's PRS image and demographic information. For privacy reasons, the PRS image shown is an average PRS image over 100 individuals.}}
    \label{tab-app:genomics_prompt_example}
    \renewcommand{\arraystretch}{1.5}
    \footnotesize
    \begin{tabular}{l|p{5cm}|c}
    \toprule
    \textbf{Image} & \textbf{Prompt} & \textbf{Target} \\
    \Xhline{2\arrayrulewidth}
    \raisebox{\dimexpr-\height + 1.5ex\relax}{\includegraphics[scale=0.18, trim=0 0 0 -5]{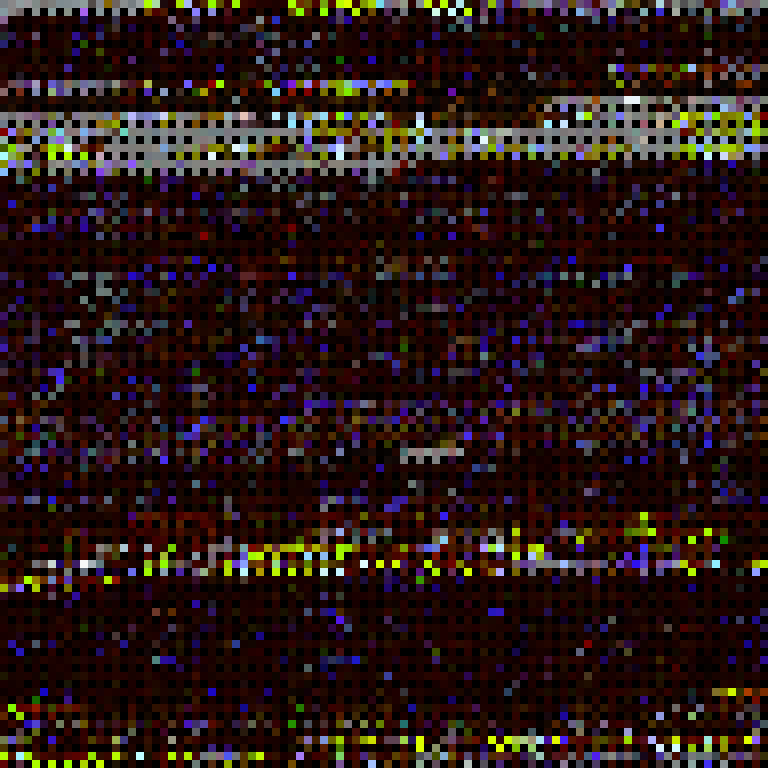}}
    &
    <\textbf{img}> Given this genomic data, and the following health information:
    
    age: 60; 
    
    sex: female; 
    
    body mass index: 26.1; 
    
    predict whether the individual has the following condition or not. Respond “Yes” or “No”. 
    
    Coronary artery disease: 
    &
    No \\ \bottomrule
    \end{tabular}
\end{table}

\begin{table}[]
    \centering
    \caption{\small{Overview of UK Biobank fields used to compile health outcomes for training and evaluating \ourmodelpolygenic}.}
    \label{tab-app:genomics_data_overview}
    \renewcommand{\arraystretch}{1.15}
    \footnotesize
    \begin{tabular}{l|p{4cm}|l}
    \Xhline{2.5\arrayrulewidth}
    \textbf{Health outcome}          & \textbf{UK Biobank fields involved}  &  \textbf{Conversion process} \\ \Xhline{2\arrayrulewidth}
    Coronary artery disease & 131307                   & \begin{tabular}[l]{@{}l@{}}If source of report of I25 (chronic ischaemic heart\\disease) available (field 131307).\end{tabular} \\ \hline
    Stroke                  & 6150, 131368, 131369, 41271, 41270 & \begin{tabular}[l]{@{}l@{}}Logical OR of:\\choice 3 of touchscreen questionnaire (field 6150),\\first reported stroke (field 131368),\\source of report stroke (field 131369),\\ICD-9 code 434.91 in field 41271,\\ICD-10 codes I63.* and I64.* in field 41270.\end{tabular}       \\ \hline
    Type 2 Diabetes         & 20002, 41270 &  \begin{tabular}[l]{@{}l@{}}Code 1223 in self-reported non-cancer illness code\\(field 20002), or ICD-10 codes E11.* in field 41270.\end{tabular}  \\ \hline
    Glaucoma                & 4689, 5326, 5327, 6148, 20003, 41202, 41203, 41205, 42104  & See \citet{khawaja2018genome} for details. \\ \hline
    COPD                    & 6152, 41270, 41271, 42040                    & See \citet{cosentino2023inference} for details. \\ \hline
    Rheumatoid arthritis    & 131851     & If source of report of M06 (field 131851). \\ \hline
    Major depression        & 20126, 41270     & \begin{tabular}[c]{@{}l@{}}Code 3, 4, 5 in bipolar and major depression status\\(field 20126), or ICD-10 codes F32.* and F33.* in field \\41270.\end{tabular}  \\ \hline
    All-cause mortality     & 40000     & If there is a value in date of death (field 40000).
 \\ \hline
    Hypertension            & 20002, 41271, 41270 & \begin{tabular}[c]{@{}l@{}}Logical OR of:\\Code 1065 in self-reported non-cancer illness code\\(field 20002),\\ICD-9 code 401.* and 405.* in field 41271,\\ICD-10 codes I10 and I15.* in field 41270.\end{tabular} \\ \hline
    Hypercholesterolemia    & 30780, 30760        & \begin{tabular}[c]{@{}l@{}}Labels are calculated using LDL (30780) and\\HDL(30760) by:\\LDL cholesterol >= 190 mg/dL OR\\(LDL >= 160 mg/dL AND\\((HDL < 40 mg/dL AND person is Male) OR\\(HDL < 55 mg/dL AND person is Female)))\\Data reported in units of mmol/L,\\to convert to mg/dL, multiply these by 38.67. \end{tabular} \\ \hline
    Atrial Fibrillation     & 20002, 41271, 41270 & \begin{tabular}[c]{@{}l@{}}Logical OR of:\\Code 1471 in self-reported non-cancer illness code\\(field 20002),\\ICD-9 code 4273 in field 41271,\\ICD-10 codes I48(|0|1|2|9) in field 41270.\end{tabular} \\ \hline
    Diabetic Retinopathy    & 6148, 20002, 5890, 5901, 41271, 41270  & \begin{tabular}[c]{@{}l@{}}Logical OR of:\\Code 1276 in self-reported non-cancer illness code\\(field 20002),\\Code 1 in eye problem/disorder (6148),\\Code 1, 2, 3 in Which eye(s) affected by\\diabetes-related eye disease(5890),\\There is a value in Age when diabetes-related eye\\disease diagnosed (5901),\\ICD-9 code 3620 in field 41271,\\ICD-10 codes H360 in field 41270.\end{tabular} \\ \hline
    Asthma                  & 20002, 41270                    & \begin{tabular}[c]{@{}l@{}}Code 1111 in self-reported non-cancer illness code\\(field 20002), or ICD-10 codes J45.*, J46.* in field 41270.\end{tabular} \\ \hline
    Pneumonia               & 41271, 41270     & \begin{tabular}[c]{@{}l@{}}ICD-9 code 401.* and 480, 481, 482, 483, 484,\\486 in field 41271,\\ICD-10 codes J12.*, J13, J14, J15.*, J16.*, J17.*, \\J18.* in field 41270.\end{tabular} \\ \Xhline{2.5\arrayrulewidth}
    \end{tabular}
\end{table}

\begin{table}[]
    \centering
    \caption{\small{\textbf{Polygenic risk prediction health outcomes correlation.} Most correlated In-distribution (ID) health outcomes for each out-of-distribution (OOD) health outcome. In general, the most related in-distribution outcome is defined as the outcome with the highest Matthew's correlation coefficient with the out-of-distribution outcome across individuals in our training set.}}
    \label{tab-app:genomics_ood_correlations}
    \renewcommand{\arraystretch}{1.5}
    \small
    \begin{tabular}{l|cc}
    \Xhline{2.5\arrayrulewidth}
    \textbf{OOD health outcome} & \begin{tabular}[c]{@{}l@{}}\textbf{ID health outcome}\end{tabular} & \begin{tabular}[c]{@{}l@{}}\textbf{Matthew's correlation coefficient}\end{tabular} \\ \toprule
    Hypertension         & Coronary Artery Disease & 0.266 \\ 
    Hypercholesterolemia & Major Depression        & 0.007 \\ 
    Atrial Fibrillation  & Coronary Artery Disease & 0.240 \\ 
    Diabetic Retinopathy & Type 2 Diabetes         & 0.325 \\ 
    Asthma               & COPD                    & 0.186 \\ 
    Pneumonia            & All-cause Mortality     & 0.352 \\ \bottomrule
    \end{tabular}
\end{table}

\clearpage
\section{Additional results}

\subsection{Data-efficient classification}\label{sec:data-efficient-classification}
We performed data-efficient classification for Chest X-ray classification task focusing on examples across 8 different findings (atelectasis, cardiomegaly, airspace opacity, consolidation, fracture, pneumothorax, pleural effusion, and pulmonary edema). We also deploy two out-of-distribution datasets including ChestX-ray14 and CheXpert for this purpose. Our data-efficient classification follows the protocol from~\cite{xu2023elixr} except that instead of training a Multilayer Perceptron (MLP) as a nonlinear classifier, we train a linear probe on top of the frozen image encoder. Following the ELEVATER\cite{li2022elevater} method, we initialize the weights of the final linear layer with the text embeddings for the class label. Training parameters includes a learning rate of 0.2, a batch size of 512, and 300 epochs utilizing the Layer-wise Adaptive Rate Scaling (LARS) optimizer. 

\begin{figure}[htp]
    \centering
    \includegraphics[width=0.55\linewidth]{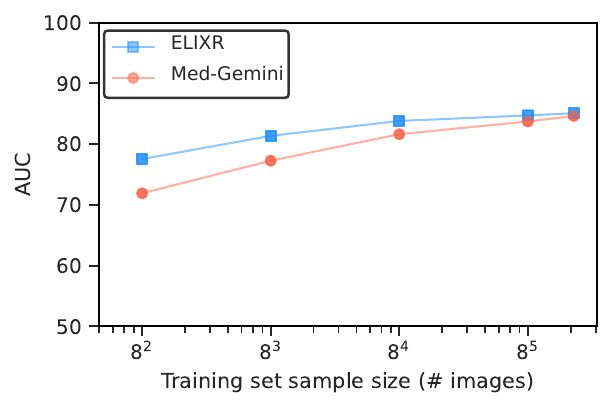}
    \caption{\small{Data-efficient classification results for \ourmodel~\vs ELIXR. We target classification across 8 different findings (atelectasis, cardiomegaly, airspace opacity, consolidation, fracture, pneumothorax, effusion, and pulmonary edema) and 2 out-of-distribution datasets (ChestX-ray14 and CheXpert). Linear probes trained on top of visual embeddings from \ourmodel show strong performance on classification although roughly one order of magnitude inferior to ELIXR at sample sizes as low as 64 examples.}}
    \label{fig-app:data_efficient_classification}
\end{figure}

In alignment with previous best-in-class method, ELIXR~\cite{xu2023elixr}, the linear classifiers were trained on 5 different varying sample sizes including 0.01\% to 100\% subsets of the training data to facilitate direct comparability of results to~\citet{xu2023elixr}. The smallest sample size includes 64 samples. Figure~\ref{fig-app:data_efficient_classification} shows aggregated results of \ourmodel~\vs ELIXR on ChestX-ray14 and CheXpert~\cite{xu2023elixr} for 5 and 6 various runs, respectively. Comparison between data-efficient classification results of \ourmodel~\vs ELIXR reveals that linear probes trained on top of visual embeddings from \ourmodel exhibit robust performance in data-efficient classification, although approximately one order of magnitude inferior than ELIXR at the sample size as low as 64 samples. 

\subsection{Polygenic risk prediction}
\label{sec-app:PRS}
\begin{figure}[t]
    \centering
    \includegraphics[width=1\linewidth]{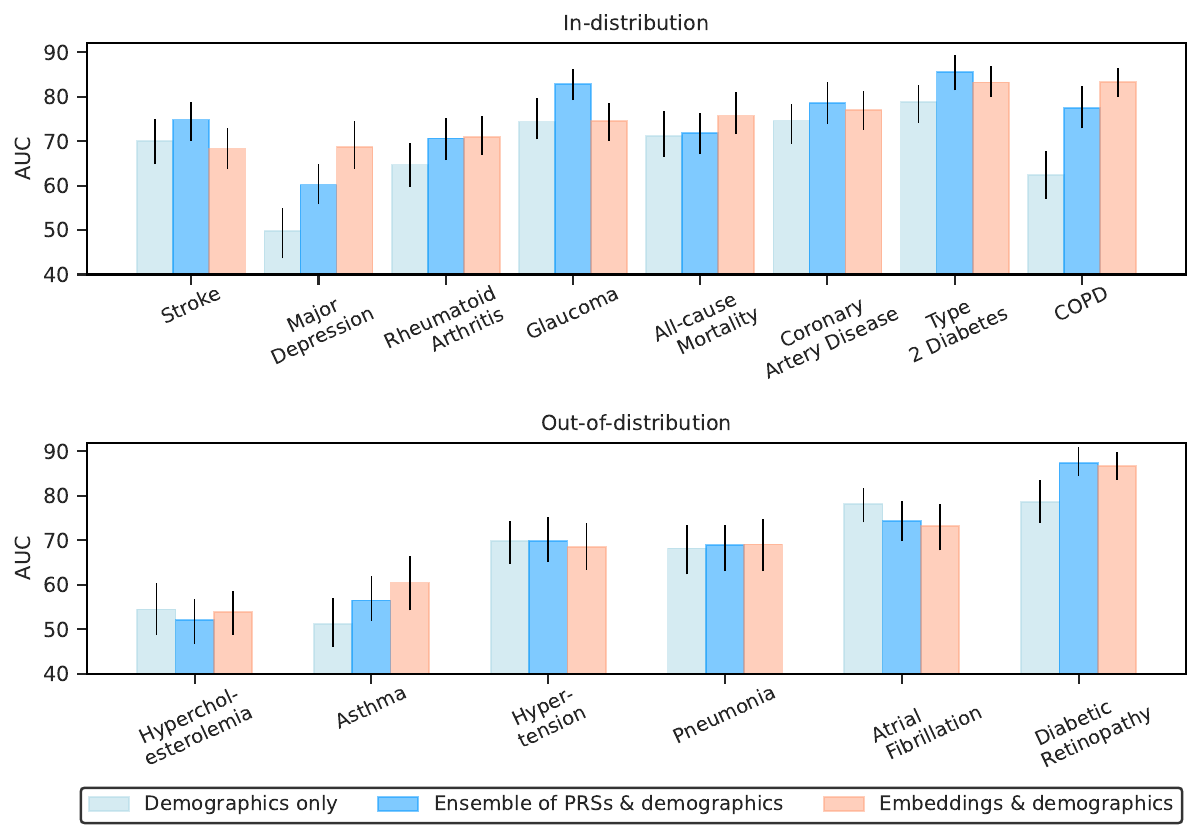}
    \caption{\small{Health outcome prediction using linear probes for both in-distribution (ID) and out-of-distribution (OOD) outcomes on balanced evaluation sets of 400 individuals. ``Demographics only'' used a linear probe of \texttt{age}, \texttt{sex}, and \texttt{BMI} to predict each health outcome, ``Ensemble of PRSs and demographics'' combined demographics with all 7,145 PRSs in a linear probe, and ``Embeddings and demographics'' combined demographics with genetic risk embeddings. For OOD health outcomes, the linear probes were trained to predict the most-correlated ID outcome (Table \ref{tab-app:genomics_ood_correlations}), and those predictions were then evaluated on the OOD outcome.}} 
    \label{fig-app:genomics_linear_gemini_set}
\end{figure}

\begin{figure}[htp]
    \centering
    \includegraphics[width=1\linewidth]{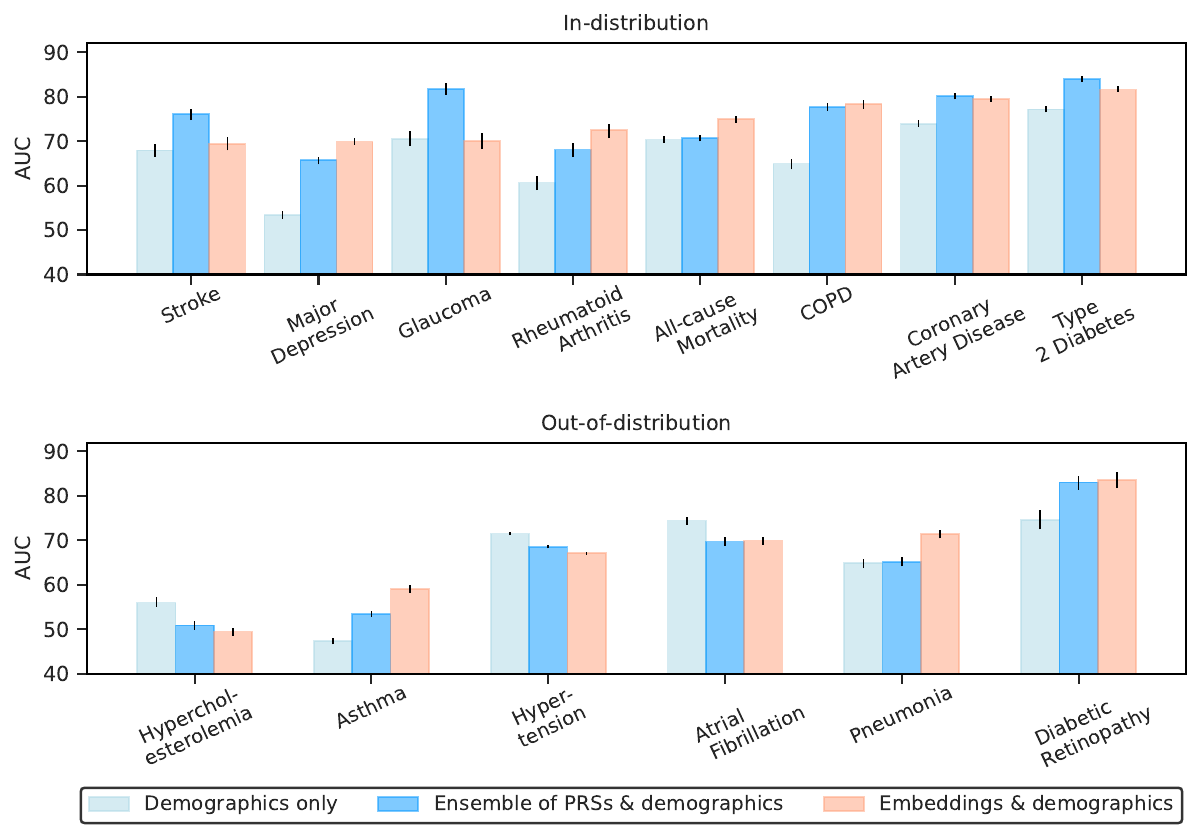}
    \caption{\small{Health outcome prediction using linear probes for both in-distribution (ID) and out-of-distribution (OOD) outcomes on larger balanced evaluation sets. For each outcome, the evaluation set includes all positive cases in our test set and the same number of negative controls. ``Demographics only'' used a linear probe of \texttt{age}, \texttt{sex}, and \texttt{BMI} to predict each health outcome, ``Ensemble of PRSs and demographics'' combined demographics with all 7,145 PRSs in a linear probe, and ``Embeddings and demographics'' combined demographics with genetic risk embeddings. For OOD health outcomes, the linear probes were trained to predict the most-correlated ID outcome (Table \ref{tab-app:genomics_ood_correlations}), and those predictions were then evaluated on the OOD outcome.}} 
    \label{fig-app:genomics_linear_larger_balanced_set}
\end{figure}

\begin{figure}[htp]
    \centering
    \includegraphics[width=0.8\linewidth]{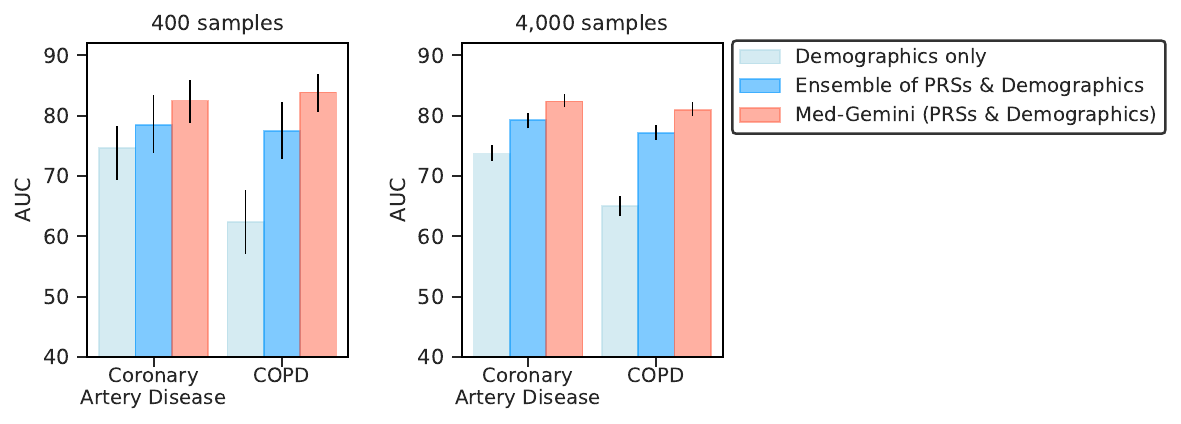}
    \caption{\small{Health outcome prediction performance on evaluation sets of 400 and 4,000 balanced samples for coronary artery disease and COPD. The left plot replicates Figure \ref{fig:genomics_mosaic_vs_baselines} results. The right plot is the analogous performance in larger samples of 4,000 balanced case-control datasets.}} 
    \label{fig-app:genomics_cad_copd_gemini_fourk}
\end{figure}

Beyond evaluating \ourmodelpolygenic, we also compared linear probes of the \ourmodelpolygenic embeddings to linear probes of the demographics only and the ensemble of PRSs and demographics. Figure \ref{fig-app:genomics_linear_gemini_set} uses the same balanced sets of 400 individuals as used in Figure \ref{fig:genomics_mosaic_vs_baselines}, and Figure \ref{fig-app:genomics_linear_larger_balanced_set} uses larger balanced sets containing all the positive cases per health outcome and an equal number of controls. The AUC metrics are relatively consistent between both evaluation sets. Furthermore, we computed  \ourmodelpolygenic performance on coronary artery disease and COPD in  4000-sample evaluations (Figure \ref{fig-app:genomics_cad_copd_gemini_fourk}), and observed stable results. Taken together, these results suggesting that our evaluation set of 400 individuals is representative of overall model performance.

\begin{figure}[htp]
    \centering
    \includegraphics[width=1\linewidth]{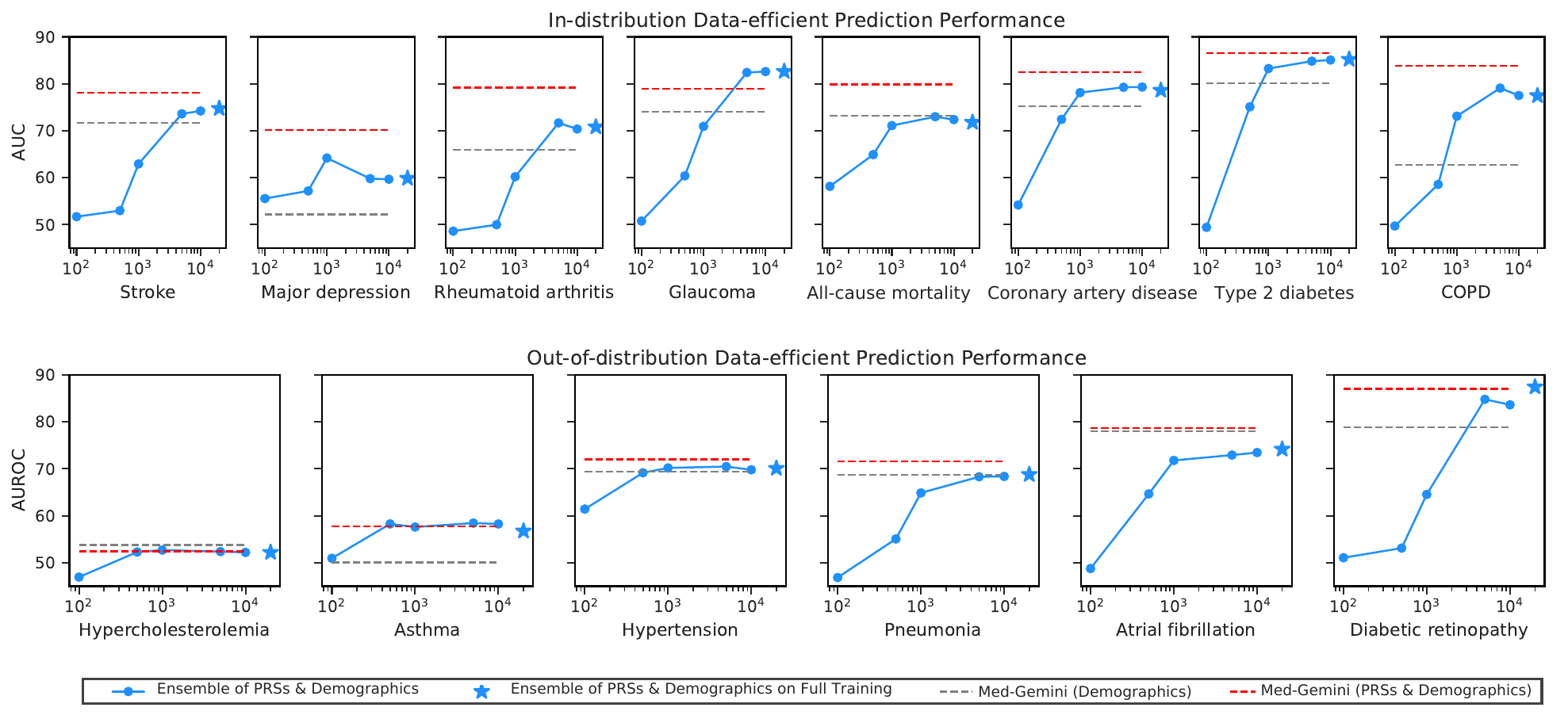}
    \caption{\small{Comparing health outcomes prediction performance of \ourmodelpolygenic (zero-shot for out-of-distribution outcomes) with PRS plus demographics linear probes trained with different sample sizes. ``Ensemble of PRSs \& Demographics'' are linear models trained against the given health outcome (or the most related ID health outcome for OOD outcomes) using all 7,145 PRSs and demographics. Dashed constant lines show the performance of \ourmodelpolygenic on predicting outcomes, given only demographics or genetic risk ``image'' plus demographics. All linear models in this experiment were trained on population-prevalence data splits.}}
    \label{fig-app:genomics_ablation_all}
\end{figure}

In addition, we demonstrated that using the \ourmodelpolygenic framework likely results in better predictive performance than linear models trained with all PRS featurizations plus demographics regardless of future sample sizes available by conducting sample size ablation tests on the PRS ensemble models. We observed performance plateaus for the linear model with at most $10^4$ samples (Figure \ref{fig-app:genomics_ablation_all}).

Finally, we investigated the relative contributions of the genomic embedding and modeling non-linear interactions between genomic representations and demographic information by comparing the performance of \ourmodelpolygenic to two other non-linear models: a gradient-boosted decision tree (GBDT) of the ``Embeddings and demographics'' (``Embeddings'') and a GBDT of the most correlated individual PRS at each of the three significance thresholds and demographics (``Best PRSs''). \ourmodelpolygenic and the GBDT of ``Embeddings and demographics'' yield comparable performance across all traits, and consistently outperform the GBDT of ``Best PRS'' for in-distribution outcomes, confirming the importance of both multi-PRS predictors and accurately modeling non-linear interactions between genetic contributors and demographic information (Table~\ref{tab-app:genomics_nonlinear}).

\begin{figure}[htp]
    \centering
    \includegraphics[width=1\linewidth]{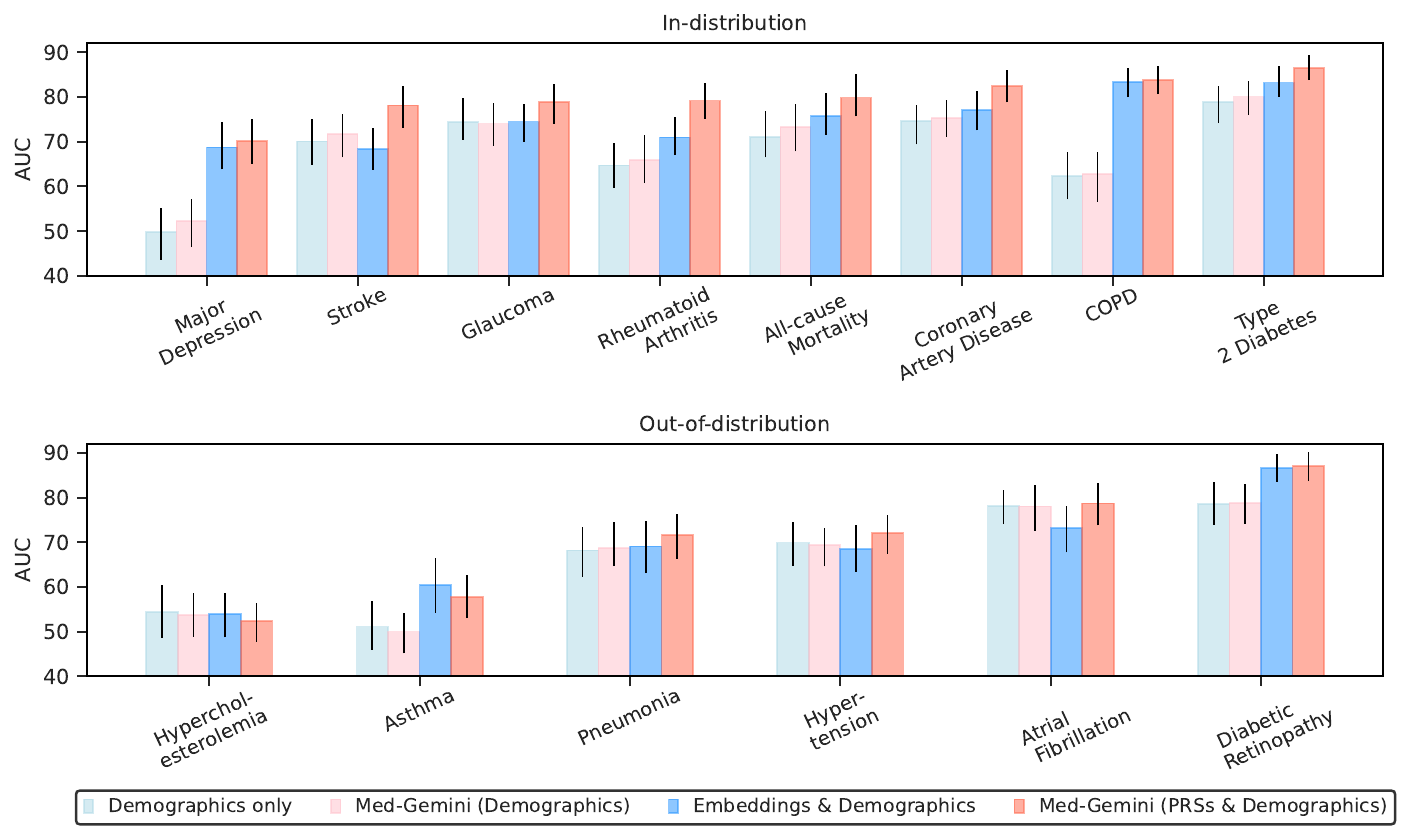}
    \caption{\small{Health outcome prediction using linear probes and \ourmodelpolygenic for both in-distribution (ID) and out-of-distribution (OOD) outcomes on balanced evaluation sets of 400 individuals. ``Demographics only'' used a linear probe of \texttt{age}, \texttt{sex}, and \texttt{BMI} to predict each health outcome, and ``Embeddings and demographics'' combined demographics with genetic risk embeddings. \ourmodelpolygenic was prompted with either demographic information only or the individual's PRS image and demographic information. For OOD health outcomes, the linear probes (``Demographics only'' and ``Embeddings and demographics'') were trained to predict the most-correlated ID outcome (Table \ref{tab-app:genomics_ood_correlations}), and those predictions were then evaluated on the OOD outcome.}} 
    \label{fig-app:genomics_coca_embeddings_vs_mosaic}
\end{figure}

\begin{table}[h]
\small
\centering
\caption{\small{\textbf{Prediction performance measured by AUC for non-linear models of genomics and demographics.}
The strong performance of \ourmodelpolygenic stems from both the inclusion of multiple PRS
in the genomic representation and capturing the non-linear interactions between genomics and demographics. ``Embeddings'', a GBDT of the \ourmodel embeddings and demographics. ``Best PRS'', a GBDT of the most correlated individual PRS with the outcome at each significance level and demographics.
}}
\label{tab-app:genomics_nonlinear}
\renewcommand{\arraystretch}{1.5}
\begin{tabular}{l|ccc}
\toprule
Outcome & \ourmodelpolygenic & Embeddings & Best PRS \\
\toprule
Major Depression & \textbf{70.2} & 68.3 & 57.1 \\
Stroke & \textbf{78.1} & 77.3 & 75.0 \\
Glaucoma & 78.9 & 78.1 & \textbf{80.4} \\
Rheumatoid Arthritis & \textbf{79.2} & 79.0 & 76.4 \\
All-cause Mortality & 79.9 & \textbf{80.3} & 75.3 \\
Coronary Artery Disease & \textbf{82.5} & 79.7 & 77.3 \\
COPD & \textbf{83.9} & \textbf{83.9} & 72.7 \\
Type 2 Diabetes & \textbf{86.5} & 86.1 & 85.3 \\
\midrule
Hypercholesterolemia & 52.4 & \textbf{54.1} & 53.5 \\
Asthma & 57.7 & 59.0 & \textbf{63.9} \\
Pneumonia & 71.6 & \textbf{73.3} & 70.3 \\
Hypertension & 72.0 & 70.2 & \textbf{72.1} \\
Atrial Fibrillation & \textbf{78.7} & 78.4 & 68.5 \\
Diabetic Retinopathy & 87.0 & \textbf{89.0} & 84.5 \\
\bottomrule
\end{tabular}
\end{table}

\clearpage
\subsection{MIMIC-CXR classification}
Table~\ref{tab-app:results-cxr-classification} shows the comparison between performance of \ourmodel and \geminixl measure by F1-score for the original label and revised label as explained in Section~\ref{sub:mimic_labels}. Revised MIMIC-CXR labels significantly improve chest X-ray classification performance measured by F1-score (\%). Our results demonstrate the impact of accurate ground truth on model evaluation. 

\begin{table}[h]
\small
\centering
\caption{\small{\textbf{MIMIC-CXR classification performance using the revised labels \vs the original.} Performance on chest X-ray MIMIC-CXR classification measured by F1-score (\%) when using the original \vs the revised labels. Our experiment shows the measured performance is improved when the revised version of labels are used as the ground-truth.}}
\label{tab-app:results-cxr-classification}
\renewcommand{\arraystretch}{1.5}
\begin{tabular}{l|l|cc|cc}
\Xhline{2.5\arrayrulewidth}
\multirow{2}{*}{\textbf{Datasets}}          &\multirow{2}{*}{\textbf{Condition}}& \multicolumn{2}{c|}{\textbf{Original Labels - F1(\%)}}   & \multicolumn{2}{c}{\textbf{Revised Labels - F1(\%)}}  \\ \cline{3-6}
                           &                  & \textbf{\ourmodel} & \textbf{\geminixl}         & \textbf{\ourmodel} & \textbf{\geminixl} \\
\Xhline{2\arrayrulewidth}
\multirow{6}{*}{MIMIC-CXR} 
                           & \multirow{1}{*}{Atelectasis}             & \textbf{99.3}     & 90.7          & \textbf{99.8}    & 88.1 \\ 
                           & \multirow{1}{*}{Cardiomegaly}            & 92.8              & \textbf{93.8} & 94.1             & \textbf{94.6}  \\
                           & \multirow{1}{*}{Consolidation}           & \textbf{79.5}     & 76.3          &\textbf{82.0}     & 77.0 \\
                           & \multirow{1}{*}{Edema}                   & \textbf{86.7}     & 86.1          &\textbf{86.8}     & 86.4 \\
                           & \multirow{1}{*}{Pleural Effusion}        &\textbf{91.0}      & 89.3          &\textbf{90.8}     & 88.1  \\
                           & \multirow{1}{*}{Class-weighted average}  &\textbf{92.1}      & 89.0          &\textbf{91.2}     & 88.6  \\
                           & \multirow{1}{*}{Normal/Abnormal}         &\textbf{40.9}      & 6.3           &\textbf{42.0}     & 29.7  \\ 

\Xhline{2.5\arrayrulewidth}
\end{tabular}
\end{table}

\subsection{Histopathology classification}
Table~\ref{tab:results-path-classification} details the linear probing results for the histopathology patch-classification task, reporting  1-vs-rest AUC (\%) with 95\% confidence intervals. The confidence intervals  were obtained using blocked bootstrap resampling over test set slides with 10,000 replicates. Our model's image embeddings match the performance of the histopathology-specialized model (PathSSL) on 6 out of 9 in-distribution tasks, with room for improvement on the remaining tasks. While \basemodel and \ourmodeltwod perform similarly overall, \ourmodel shows a trend towards higher mean AUC on most in-distribution tasks and both out-of-distribution tasks.

\begin{table}[]
\small
\centering
\caption{\small{\textbf{Histopathology patch-classification with linear-probe.} We measure the 1-$vs$-rest AUCs (\%) from linear-probing on our histopathology patch-classification tasks. The 95\% confidence intervals were obtained by a blocked bootstrap resampling (over test set slides) using 10000 replicates. The image embeddings from \ourmodel perform on par with a histopathology-specialized foundation model (PathSSL) on 6 of the 9 in-distribution tasks. The remaining 3 in-distribution and 2 out-of-distribution tasks leave room for improvement. \basemodel performs similarly to \ourmodeltwod across all tasks, although \ourmodel trends higher in terms of mean AUC(\%) on 7 of the 9 in-distribution tasks, and both out-of-distribution tasks.}}
\label{tab:results-path-classification}
\renewcommand{\arraystretch}{1.5}
\begin{tabular}{l|cccc}
\Xhline{2.5\arrayrulewidth}
\textbf{Dataset} & \textbf{AugReg-ImageNet} & \textbf{\geminixl} & \textbf{\ourmodel} & \textbf{PathSSL}~\citep{lai2023domain} \\ \Xhline{2\arrayrulewidth}
CAMELYON16	&	\begin{tabular}[c]{@{}c@{}}95.09 \\ \scriptsize{(93.38 - 96.58)}\end{tabular}	&	\begin{tabular}[c]{@{}c@{}}96.76 \\ \scriptsize{(95.13 - 98.13)}\end{tabular}	&	\begin{tabular}[c]{@{}c@{}}98.48 \\ \scriptsize{(97.63 - 99.17)}\end{tabular}	&	\begin{tabular}[c]{@{}c@{}}99.00 \\ \scriptsize{(98.26 - 99.53)}\end{tabular}	\\ \hline
Lung AD	&	\begin{tabular}[c]{@{}c@{}}86.04 \\ \scriptsize{(82.01 - 89.25)}\end{tabular}	&	\begin{tabular}[c]{@{}c@{}}87.90 \\ \scriptsize{(84.19 - 90.48)}\end{tabular}	&	\begin{tabular}[c]{@{}c@{}}89.57 \\ \scriptsize{(87.30 - 90.93)}\end{tabular}	&	\begin{tabular}[c]{@{}c@{}}94.41 \\ \scriptsize{(91.72 - 96.21)}\end{tabular}	\\ \hline
Gleason NCB	&	\begin{tabular}[c]{@{}c@{}}80.79 \\ \scriptsize{(78.04 - 83.62)}\end{tabular}	&	\begin{tabular}[c]{@{}c@{}}84.85 \\ \scriptsize{(82.42 - 87.37)}\end{tabular}	&	\begin{tabular}[c]{@{}c@{}}89.14 \\ \scriptsize{(87.29 - 91.04)}\end{tabular}	&	\begin{tabular}[c]{@{}c@{}}90.59 \\ \scriptsize{(88.73 - 92.45)}\end{tabular}	\\ \hline
Gleason RP	&	\begin{tabular}[c]{@{}c@{}}85.43 \\ \scriptsize{(83.12 - 87.72)}\end{tabular}	&	\begin{tabular}[c]{@{}c@{}}87.71 \\ \scriptsize{(85.30 - 89.93)}\end{tabular}	&	\begin{tabular}[c]{@{}c@{}}89.39 \\ \scriptsize{(87.22 - 91.39)}\end{tabular}	&	\begin{tabular}[c]{@{}c@{}}91.50 \\ \scriptsize{(89.77 - 93.10)}\end{tabular}	\\ \hline
Breast IC	&	\begin{tabular}[c]{@{}c@{}}89.47 \\ \scriptsize{(88.45 - 90.45)}\end{tabular}	&	\begin{tabular}[c]{@{}c@{}}91.66 \\ \scriptsize{(90.71 - 92.53)}\end{tabular}	&	\begin{tabular}[c]{@{}c@{}}92.18 \\ \scriptsize{(91.24 - 93.07)}\end{tabular}	&	\begin{tabular}[c]{@{}c@{}}94.32 \\ \scriptsize{(93.52 - 95.07)}\end{tabular}	\\ \hline
Breast NP	&	\begin{tabular}[c]{@{}c@{}}68.75 \\ \scriptsize{(66.66 - 70.73)}\end{tabular}	&	\begin{tabular}[c]{@{}c@{}}74.68 \\ \scriptsize{(72.70 - 76.59)}\end{tabular}	&	\begin{tabular}[c]{@{}c@{}}73.54 \\ \scriptsize{(71.32 - 75.71)}\end{tabular}	&	\begin{tabular}[c]{@{}c@{}}75.78 \\ \scriptsize{(73.50 - 77.90)}\end{tabular}	\\ \hline
Breast TF	&	\begin{tabular}[c]{@{}c@{}}74.49 \\ \scriptsize{(72.41 - 76.52)}\end{tabular}	&	\begin{tabular}[c]{@{}c@{}}78.25 \\ \scriptsize{(76.25 - 80.15)}\end{tabular}	&	\begin{tabular}[c]{@{}c@{}}75.66 \\ \scriptsize{(73.71 - 77.58)}\end{tabular}	&	\begin{tabular}[c]{@{}c@{}}83.34 \\ \scriptsize{(81.64 - 84.99)}\end{tabular}	\\ \hline
CIN	&	\begin{tabular}[c]{@{}c@{}}86.17 \\ \scriptsize{(84.49 - 87.77)}\end{tabular}	&	\begin{tabular}[c]{@{}c@{}}88.22 \\ \scriptsize{(86.58 - 89.79)}\end{tabular}	&	\begin{tabular}[c]{@{}c@{}}89.49 \\ \scriptsize{(87.93 - 91.03)}\end{tabular}	&	\begin{tabular}[c]{@{}c@{}}89.70 \\ \scriptsize{(87.96 - 91.30)}\end{tabular}	\\ \hline
CRC	&	\begin{tabular}[c]{@{}c@{}}97.29 \\ \scriptsize{(96.03 - 98.30)}\end{tabular}	&	\begin{tabular}[c]{@{}c@{}}98.45 \\ \scriptsize{(97.42 - 99.22)}\end{tabular}	&	\begin{tabular}[c]{@{}c@{}}98.54 \\ \scriptsize{(97.70 - 99.22)}\end{tabular}	&	\begin{tabular}[c]{@{}c@{}}98.95 \\ \scriptsize{(98.02 - 99.60)}\end{tabular}	\\ \hline  \hline
TCGA Study Type	&	\begin{tabular}[c]{@{}c@{}}86.65 \\ \scriptsize{(85.53 - 87.75)}\end{tabular}	&	\begin{tabular}[c]{@{}c@{}}91.85 \\ \scriptsize{(91.04 - 92.60)}\end{tabular}	&	\begin{tabular}[c]{@{}c@{}}92.30 \\ \scriptsize{(91.47 - 93.05)}\end{tabular}	&	\begin{tabular}[c]{@{}c@{}}96.38 \\ \scriptsize{(95.83 - 96.88)}\end{tabular}	\\ \hline
Tissue Type	&	\begin{tabular}[c]{@{}c@{}}89.37 \\ \scriptsize{(88.50 - 90.19)}\end{tabular}	&	\begin{tabular}[c]{@{}c@{}}92.25 \\ \scriptsize{(91.51 - 92.94)}\end{tabular}	&	\begin{tabular}[c]{@{}c@{}}92.70 \\ \scriptsize{(92.00 - 93.34)}\end{tabular}	&	\begin{tabular}[c]{@{}c@{}}94.76 \\ \scriptsize{(94.16 - 95.30)}\end{tabular}	\\
\Xhline{2.5\arrayrulewidth}
\end{tabular}
\end{table}

\clearpage
\section{Evaluation metrics}
\label{sec-app-evaluation-metrics}

Beyond human and expert evaluation, we leverage a range of automated metrics tailored to specific tasks. For classification tasks, this may include basic accuracy and AUC (Area Under the ROC Curve) metrics. For tasks like report generation, where the fidelity and informativeness of the generated text are crucial, we employ wide variety of metrics such as BLEU, Rouge-L or RadGraph F1-score to probe the quality of our models.

\paragraph{Accuracy} Used for image classification and close-ended VQA inference tasks. Measures the percentage of correct predictions \vs the ground truth. 

\paragraph{AUC (Area Under the ROC Curve)} AUC is a performance metric for classification models that indicates how well a model distinguishes between different classes. AUC is calculated by plotting the True Positive Rate (TPR) against the False Positive Rate (FPR) at various classification thresholds. The TPR measures the proportion of correctly identified positive instances, while the FPR measures the proportion of incorrectly identified negative instances. The area under this curve represents the model's overall ability to separate classes. An AUC of 1.0 indicates a perfect classifier, while an AUC of 0.5 implies the model has no better discriminative power than random guessing.

\paragraph{F1 Score} The F1 score is a valuable metric for evaluating classification models, especially when dealing with imbalanced datasets. F1 score is calculated as the harmonic mean of precision (the proportion of true positive out of all predicted positives) and recall (the proportion of true positives correctly identified). The Weighted F1 Score which is used for VQA, averaging F1 scores across classes based on their frequency. Macro-F1 score used for image classification averaging F1 scores across classes without considering imbalances.

\paragraph{Sensitivity} Used for image classification in ophthalmology related tasks. Measures the percentage of correctly identified positive cases out of all actual positive cases.  A model with high sensitivity minimizes false negatives.

\paragraph{Specificity} Used for image classification in ophthalmology related tasks. Measures the percentage of correctly identified negative cases out of all actual negative cases. A model with high specificity minimizes false positives.

\paragraph{Tokenized F1} Tokenized F1-score provides a granular evaluation of language models by calculating precision, recall, and F1-score at the individual token level. This means it rewards partial matches, recognizing the model's ability to identify elements within a sequence even if they're not perfectly aligned. For this purpose True positives and false positives are determined as the number of correctly generated tokens and tokens generated but not present in the ground truth, respectively. False negatives are tokens present in the ground truth but missed by the model.

\paragraph{Rouge-L} Rouge-L measures evaluates the quality of generated text and text summarization by comparing the longest common subsequence (LCS) between generated and reference text~\cite{lin2004rouge}. Higher scores indicate better content and better salient point capturing. Rouge-L assesses the similarity between generated and reference text by measuring the overlap of their LCS and calculating recall based on the LCS length relative to the reference text  This metric takes into account the order of words in the text, which makes it particularly suitable for evaluating summaries or text generation tasks where the order of words matters. The higher the Rouge-L score, the better the quality of the generated text compared to the reference text. ROUGE-L relies heavily on LCS and exact matches limiting the contextual understating of the generated text and increasing the sensitivity to sentence length. A high ROUGE-L score doesn't necessarily ensure that the generated text is grammatically correct, well-structured, or reads naturally.

\paragraph{CIDEr} CIDEr (Consensus-based Image Description Evaluation) is a metric specifically designed to assess the quality of captions generated for images and short text passages. It goes beyond simple word overlap by considering both the n-gram matches (sequences of consecutive words) and the importance of those n-grams~\cite{vedantam2015cider}. In the preprocessing, both generated and reference texts are converted to lowercase and common stop words (``the'', ``a'', ``an'') are removed. Words are also stemmed, reducing them to their root form (e.g., ``running'' becomes ``run''). Then every generated text is broken down into a series of n-grams which are sequences of `n' consecutive words. A weight is assigned to each n-gram based on its Term Frequency-Inverse Document Frequency (TF-IDF). This means common n-grams across all texts receive lower weights, while those that are more informative and distinctive get higher weights. The cosine similarity is calculated between the TF-IDF weighted n-gram vectors of the generated text and each reference.  The individual similarity scores are averaged to produce the final CIDEr score. CIDEr can struggle to recognize texts that are semantically similar but use different synonyms and suffer from limited contextual understanding. 

\paragraph{BLEU score} The BLEU (Bilingual Evaluation Understudy) score is a widely used metric for evaluating the quality of AI generated text. It essentially compares a generated text to a set of human-written reference, providing a score that indicates how similar they are~\cite{papineni2002bleu}.  BLEU focuses on n-gram precision, meaning it checks how often sequences of n consecutive words in the generated text appear in any of the reference. It also considers a brevity penalty to discourage generations that are significantly shorter than the reference text. Higher BLEU scores indicate better translation quality, with a perfect score of 1.0 signifying a perfect match between the generated text and the reference. BLEU score has limitations including lack of penalization for grammatical correctness, fluency, or semantic equivalence. Additionally, the quality of the reference and ground truth can impact the BLEU score.

\paragraph{RadGraph F1-score}
RadGraph F1-score~\cite{jain2021radgraph} is a performance metric specifically designed to evaluate the accuracy of models that extract structured medical information from radiology reports. Unlike standard F1-scores, RadGraph F1-score considers not only whether a finding is correctly identified but also the accuracy of its relationships with other findings within the report. This is crucial because radiology reports often describe complex relationships between abnormalities, locations, and other attributes. Although RadGraph F1-score has its shortcomings, in comparison to other automated NLG metrics provides a more holistic assessment of a model's ability to understand the nuanced information present in free-text radiology reports.

While RadGraph F1-score offers a more nuanced evaluation than standard F1-scores for radiology report analysis, it has potential limitations. First, it relies on accurate RadGraph creation from the original text. Errors in entity extraction or relation identification during this pre-processing stage could cascade into the RadGraph F1-score calculation. Secondly, it might be overly strict for partial matches and slight discrepancies in relationships or minor variations in wording could significantly penalize the score. Finally, it may not fully account for the clinical relevance of certain errors, treating all mismatches equally despite the potential for varying real-world impact.

To compute the RadGraph F1-score, the model's predictions on chest X-ray images are compared against ground-truth report made by radiologists or other experts. To increase robustness of our calculation to slight format changes, before passing the ground-truth and the generated report through the RadGraph F1-score package~\cite{yu2023evaluating}, we normalize both free-form text to lowercase. The F1-score takes into account both false positives (cases where the model incorrectly identifies an abnormality) and false negatives (cases where the model fails to detect a true abnormality). By considering both precision (the ratio of true positives to the total number of predicted positives) and recall (the ratio of true positives to the total number of actual positives), the F1-score provides a balanced assessment of the model's performance. A higher RadGraph F1-score indicates better performance in accurately identifying abnormalities in medical images, which is crucial for assisting radiologists in diagnosis and treatment planning.

\clearpage
\section{Supplementary Table for Performance Summary}

Table \ref{tab-appendix:performance-all} presents the aggregate performance of \ourmodel compared to the previous state-of-the-art (SoTA), or a strong baseline where available. Figure \ref{fig:teaser} illustrates the relative improvement gained by using one of our \ourmodel models over the SoTA or strong baseline, using \basemodel as a reference point when no SoTA is available.  For pathology classification, we averaged AUC performance across all sub-datasets. For report generation, we calculated the micro average performance across normal and abnormal cases, expert identified ``AI generated report is superior or similar to original report" (see Table~\ref{tab:report-eval-2d})



\begin{table}[hp]
\centering
\footnotesize
\caption{\small{\textbf{Overall Performance Summary of \ourmodel} This table represent the aggregated results comparing \ourmodel to the previous state-of-the-art (SoTA), \basemodel or strong baseline where available.}}
\label{tab-appendix:performance-all}
\renewcommand{\arraystretch}{1.2}
\begin{tabular}{@{\hspace{0.1em}}l@{\hspace{0.1em}}cccccc@{\hspace{0.1em}}}
\toprule
Capabilities                     & Datasets                & Metric   & \ourmodel        &\basemodel & \begin{tabular}[c]{@{}c@{}}Baseline or \\ SoTA\end{tabular}            & Reference                  \\ \toprule
\multirow{3}{*}{\begin{tabular}[l]{@{}c@{}}Report\\ Generation\end{tabular}}
                                 & MIMIC-CXR         & RadGraph & 24.4  & N/A       & 20.5                        & \citet{tanno2024consensus}  \\
                                 & MIMIC-CXR         & \begin{tabular}[c]{@{}c@{}}Expert\\ (AI superior)\end{tabular}   & 47.6  & N/A       & 43.0                        & \citet{tanno2024consensus}  \\
                                 & IND1              & \begin{tabular}[c]{@{}c@{}}Expert\\ (AI superior)\end{tabular}   & 75.4  & N/A       & 63.7                        & \citet{tanno2024consensus}  \\ \midrule
                                 & MIMIC-CXR VQA           & Accuracy & 78.6  & 70.9     & 68.1                        & \citet{xu2023elixr}          \\
                                 & Slake-VQA               & Accuracy & 84.8  & 70.4     & 91.1                        & \citet{li2023self}           \\
                                 & VQA-Rad CXR             & Expert   & 71.9  & N/A      & 57.9                        & \citet{xu2023elixr}          \\
                                 & VQA-Rad CXR             & Accuracy & 78.8  & 62.4     & N/A                         & N/A                         \\ 
\multirow{-5}{*}{VQA}            & PathVQA                 & Accuracy & 83.3  & 62.8     & 90.9                        & \citet{sun2024pathasst}      \\ \midrule
\multirow{14}{*}{\begin{tabular}[l]{@{}c@{}}Genomic\\Disease\\  Prediction\end{tabular}}
                                 & Coronary artery disease & AUC      & 82.5  & N/A       & 78.5                        &  \multirow{14}{*}{\begin{tabular}[c]{@{}c@{}}Ensemble of PRSs \\ and Demographics\end{tabular}}\\
                                 & Stroke                  & AUC      & 78.1  & N/A       & 74.8                        &  \\
                                 & Type 2 diabetes         & AUC      & 86.5  & N/A       & 85.5                        &  \\
                                 & Glaucoma                & AUC      & 78.9  & N/A       & 82.8                        &  \\
                                 & COPD                    & AUC      & 83.9  & N/A       & 77.4                        &  \\
                                 & Rheumatoid arthritis    & AUC      & 79.2  & N/A       & 70.6                        &  \\
                                 & Major depression        & AUC      & 70.2  & N/A       & 60.2                        &  \\
                                 & Allcause mortality      & AUC      & 79.9  & N/A       & 71.8                        &  \\
                                 & Hypertension            & AUC      & 72.0  & N/A       & 69.9                        &  \\
                                 & Hypercholesterolemia    & AUC      & 52.4  & N/A       & 52.1                        &  \\
                                 & Atrial fibrillation     & AUC      & 78.7  & N/A       & 74.4                        &  \\
                                 & Diabetic retinopathy    & AUC      & 87.0  & N/A       & 87.5                        &  \\
                                 & Pneumonia               & AUC      & 71.6  & N/A       & 68.9                        &  \\
                                 & Asthma                  & AUC      & 57.7  & N/A       & 56.5                        &  \\ \midrule
                                 & Hard Exudates           & F1       & 87.3  & 61.5     & N/A                          & N/A                       \\
                                 & Hemorrhage              & F1       & 85.3  & 57.8     & N/A                          & N/A                      \\
                                 & PRP Scars               & F1       & 82.3  & 57.0     & N/A                          & N/A                      \\
                                 & DR Lesion               & F1       & 86.4  & 63.9     & 92.0                         & \citet{krause2018grader}  \\
                                 & PAD-UFES-20             & F1       & 71.4  & 60.3     & 70.0                         & \citet{derm2024google}     \\
                                 & MIMIC-CXR               & F1       & 90.7  & 86.8     & N/A                          & N/A                       \\
                                 & CheXpert                & F1       & 48.3  & 42.6     & 60.6                         & \citet{tiu2022expert}      \\
                                 & ChestX-ray14            & F1       & 46.7  & 34.2     & 58.3                         & \citet{majkowska2020chest}  \\
\multirow{-9}{*}{Classification} & Pathology Patch         & AUC      & 89.2  & 88.4     & 91.7                         & \citet{sun2024pathasst}     \\ \bottomrule           
\end{tabular}
\end{table}

\end{document}